\documentclass{book}

\usepackage[latin1]{inputenc}
\usepackage[]{amsmath}
\usepackage[]{amssymb}
\usepackage[mathcal]{euscript}
\usepackage{color}
\usepackage{listings}
\usepackage{amsmath,amssymb,amsfonts}
\usepackage{graphics}
\usepackage{graphicx}

\usepackage{mathrsfs}
\usepackage{algorithm}
\usepackage{algorithmic}
\usepackage{bm}
\usepackage{bbm}

\usepackage{verbatim}

\def\TReg{\textsuperscript{\textregistered}}

\def\Deg{$^\circ\,$}

\def\R{\mathrm{I\kern-0.4ex R}}
\def\N{\mathrm{I\kern-0.4ex N}}
\newcommand{\Set}[1]{\mathcal{#1}}
\newcommand{\PM}[1]{\mathscr{P}(#1)}

\newcommand{\mat}[1]{\mathbf{#1}}
\newcommand{\ve}[1]{\bm{#1}}
\newcommand{\grad}[0]{\nabla}
\newcommand{\dom}[0]{\mathbf{dom}\,}
\newcommand{\trace}[0]{\mathbf{tr}}
\newcommand{\shallbe}[0]{\stackrel{!}{=}}
\newcommand{\statetrans}[1]{\stackrel{#1}{\to}}
\newcommand{\frobenius}[0]{\text{F}}
\newcommand{\normdist}[0]{\mathcal{N}}

\def\E{\mathbb{E}}
\def\V{\mathbb{V}}
\def\C{\mathbb{C}}
\def\notexists{\mathrm{| \kern-1.0ex \exists}}
\DeclareMathOperator*{\argmax}{arg\,max}
\DeclareMathOperator*{\argmin}{arg\,min}
\DeclareMathOperator*{\setunit}{\cup}

\newtheorem{assumption}{Assumption}[chapter]
\newtheorem{definition}[assumption]{Definition}
\newtheorem{theorem}{Theorem}[chapter]
\newtheorem{proposition}[theorem]{Proposition}
\newtheorem{lemma}[theorem]{Lemma}
\newtheorem{remark}{Remark}[chapter]

\newcommand{\gap}[0]{\vspace{0.25cm}}
\newcommand{\antigap}[0]{\vspace{-0.25cm}}
\newcommand{\closetoequation}[0]{
  \vspace{-0.6cm} 

  \hspace{-0.6cm}
}
\newcommand{\proof}[1]{\antigap \textbf{Proof:} #1}

\begin{document}

\pagenumbering{Roman}

\newfont{\TUFontA}{cmssbx20}
\newfont{\TUFontB}{cmssi14}
\thispagestyle{empty}
\parbox{10cm}{{\TUFontA Technische Universit\"at Berlin}\\
$\text{\hspace{0.1cm}\TUFontB Fakult\"at f\"ur Elektrotechnik und Informatik}$}
\parbox{3cm}{\includegraphics[scale=0.04]{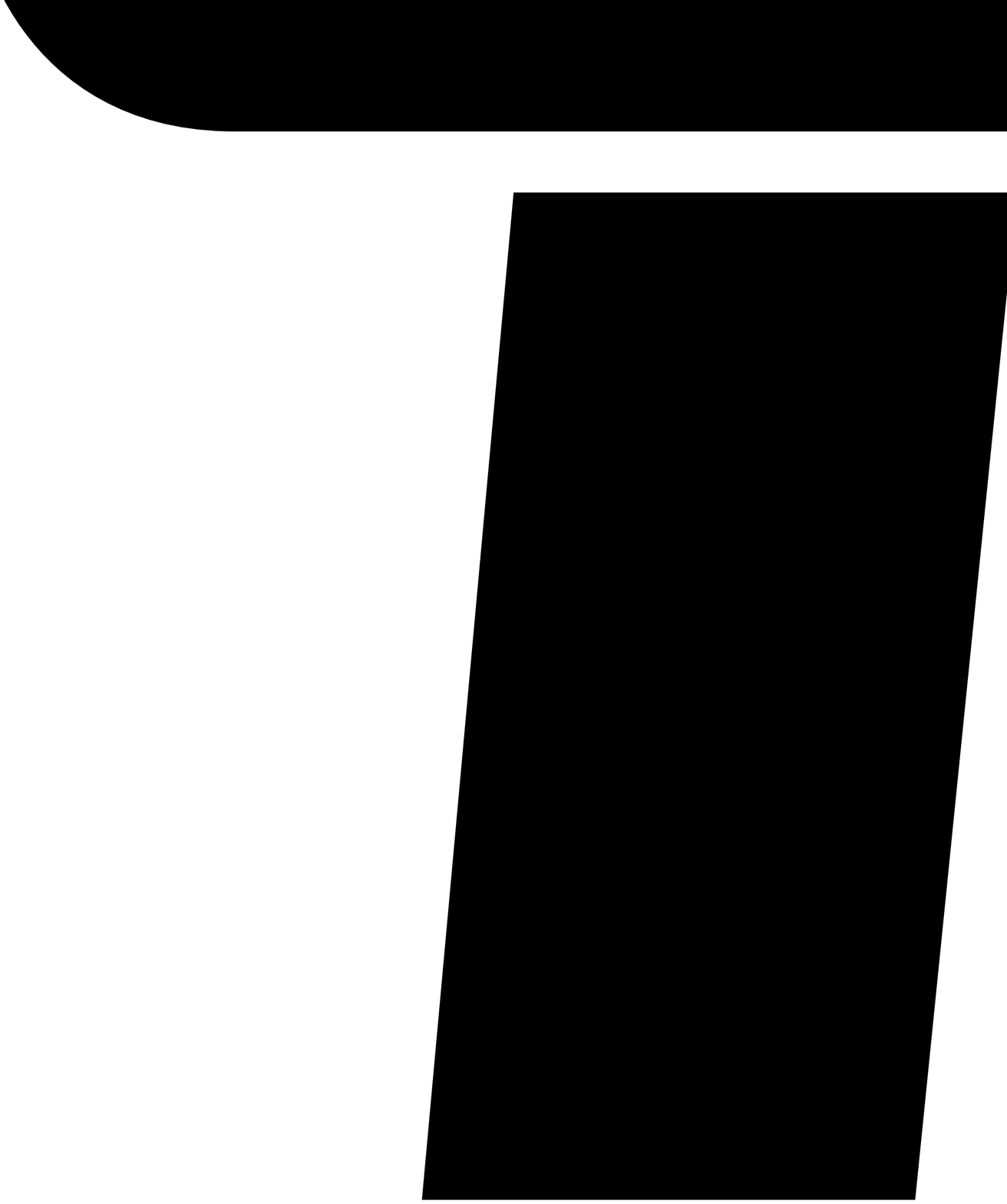}}

\vspace{2.5cm}
\begin{center}
  \Huge \textbf{ Robot Navigation using Reinforcement Learning 
                and Slow Feature Analysis}
\end{center}
\vspace{2cm}

\begin{center}
  \large
  Diplomarbeit im Studiengang Informatik \\
  \vspace{1cm}
  Vorgelegt von Wendelin B\"ohmer \\
  Matrikel 197608
\end{center}

\vfill

{\large
\textbf{Aufgebensteller:}

Prof. Dr. Klaus Obermayer

\textbf{Mitberichter:}

Dr. Robert Martin 

\textbf{Betreuer:}

Dipl.-Inf. Steffen Gr\"unew\"alder
}


\newpage
$ $\vspace{3cm}

\textbf{\LARGE Eidesstattliche Erkl\"arung}
\vspace{0.5cm}

Die selbst\"andige und eigenh\"andige 
Ausfertigung versichert an Eides statt.
\vspace{2cm}

Berlin den \today
\vspace{0.6cm}

...............................................................................................

(Wendelin B\"ohmer)


\newpage
$ $\vspace{3.75cm}

\textbf{\LARGE Zusammenfassung}
\vspace{0.5cm}

Bei der Anwendung von \textit{Reinforcement Learning}
Algorithmen auf die physische Welt 
muss man einen Weg finden,
aus komplexen Sensordaten Umgebungszust\"ande 
herauszufiltern.
Obwohl die meisten bisherigen Ans\"atze
hierf\"ur Heuristiken verwenden, legt die Biologie
nahe dass es eine Methode 
geben muss, die solche Filter
\textit{selbstst\"andig} konstruieren kann.

Neben der Extraktion von Umgebungszust\"anden
m\"ussen diese auch in einer f\"ur
moderne Reinforcement Algorithmen brauchbaren
Form pr\"asentiert werden.
Viele dieser Algorithmen arbeiten mit
linearen Funktionen, so dass die Filter 
gute lineare Approximationseigenschaften
aufweisen sollten.

Diese Diplomarbeit m\"ochte eine
un\"uberwachte Lernmethode namens
\textit{Slow Feature Analysis} (SFA)
vorschlagen, um diese Filter zu generieren.
Trainiert mit einer Zufallsfolge von
Sensordaten kann SFA eine Serie von
Filtern erlernen.
Theoretische Betrachtungen zeigen dass 
diese mit steigender Modellklasse
und Trainingbeispielen zu trigonometrischen
Funktionen konvergieren, welche f\"ur ihre guten linearen
Approximationseigenschaften bekannt sind.

Wir haben diese Theorie mit Hilfe eines 
Roboters evaluiert.
Als Aufabe soll der Roboter das Navigieren in einer
einfachen Umgebung erlernen,
wobei der
\textit{Least Squares Policy Iteration} (LSPI)
Algorithmus zum Einsatz kommt.
Als einzigen Sensor verf\"ugt der Roboter \"uber eine
auf seinem Kopf montierte Kamera,
deren Bilder sich aber aufgrund ihrer Komplexit\"at
nicht direkt als Umgebungszust\"ande eignen.
Wir konnten zeigen dass LSPI dank der
durch SFA generierten Filter eine 
Erfolgsrate von ca. $80\%$  erreichen kann.


\newpage
$ $\vspace{3.75cm}

\textbf{\LARGE Abstract}
\vspace{0.5cm}

The application of \textit{reinforcement learning}
algorithms onto real life problems
always bears the challenge of filtering 
the environmental state out of raw sensor readings.
While most approaches use heuristics,
biology suggests that there must 
exist an unsupervised method to
construct such filters \textit{automatically}.

Besides the extraction of environmental
states, the filters have to represent them in
a fashion that support modern reinforcement algorithms.
Many popular algorithms 
use a linear architecture, so one should aim
at filters that have good approximation
properties in combination with linear functions.

This thesis wants to propose the unsupervised 
method \textit{slow feature analysis} (SFA)
for this task.
Presented with a random sequence of sensor readings, 
SFA learns a set of filters.
With growing model complexity and
training examples, the filters
converge against trigonometric
polynomial functions. 
These are known to possess excellent
approximation capabilities and should therfore
support the reinforcement algorithms well.

We evaluate this claim on a robot.
The task is to learn a navigational control
in a simple environment using the
\textit{least square policy iteration} (LSPI) algorithm.
The only accessible sensor is a head mounted
video camera, but
without meaningful filtering, video images
are not suited as LSPI input.
We will show that filters learned by
SFA, based on a random walk video of the robot,
allow the learned control to navigate successfully
in ca. $80\%$ of the test trials.


\newpage
$ $\vspace{2cm}

\textbf{\LARGE Acknowledgments}
\vspace{0.5cm}

This work is based on my experience in
the \textsc{NeuRoBot} project
and on the work and help of
many people.
I want to thank all participants of
the project. 
\gap

In particular, I want thank
Steffen Gr\"unew\"alder, who
led the machine learning aspect
of the project,
and Yun Shen, who developed the presented
version of kernel SFA 
with me and Steffen.
\gap

Mathias Franzius also deserves
acknowledgment here,
since his work on place cells
and slow feature analysis has 
motivated this thesis.
\gap

At last I want to thank professor
Klaus Obermayer, whose lectures and
research group gave me the opportunity
to enter the world of academics.
\vspace{2cm}

\textbf{\LARGE Notational conventions}
\vspace{0.5cm}

\textbf{\large Algebra notation}

\begin{tabular}{ll}
$x$           & Variable \textit{x} \\
$\ve x$       & Column vector \textit{x} \\
$[x_1,\ldots,x_n]$ & An $n$ dimensional row vector with entries $x_i$ \\
$\ve 0$       & A vector with all entries being 0 \\
$\ve 1$       & A vector with all entries being 1 \\
$\mat M$      & Matrix \textit{M} \\
$\mat I$      & The identity matrix \\
$\mat M^\top$ & The \textit{transpose} of matrix $\mat M \in \R^{n \times m}$: 
                $\forall ij:
                M_{ij}^\top = M_{ji}$ \\
$\trace(\mat M)$ & The \textit{trace} 
                   of matrix $\mat M \in \R^{n \times n}: \trace(\mat M) 
                   = \sum\limits_{i=1}^n M_{ii}$\\
$||\ve x||$   & Arbitrary \textit{norm} of vector $\ve x$ \\
$||\ve x||_2$ & $L_2$-norm of vector $\ve x\in \R^n$: $||\ve x||_2 = \sqrt{\sum\limits_{i=1}^n x_i^2}$ \\
$||\mat M||_\frobenius$& \textit{Frobenius} norm of matrix $\mat M \in \R^{n \times m}$: \\
              & $||\mat M||_\frobenius = \sqrt{\sum\limits_{i=1}^n \sum\limits_{j=1}^m M_{ij}^2}
                = \sqrt{\trace(\mat M^\top \mat M)} = \sqrt{\trace(\mat M \mat M^\top)}$
\end{tabular}

\newpage
$ $\vspace{2cm}

\textbf{\large Analysis notation}

\begin{tabular}{ll} 
$\Set A$      & Set \textit{A} \\
$\emptyset$   & The empty set \\
$[a,b]$       & The real interval between $a$ and $b$: $[a,b] = \{x|a \leq x \leq b\} \subset \R$\\
$\Set A \times \Set B$ & The set of all tuples $(a,b)$ with
                $a \in \Set A$ and $b \in \Set B$ \\
$|\Set A|$    & The \textit{cardinality} of set $\Set A$ \\
$\PM{\Set A}$ & The \textit{power set} of set $\Set A$ \\
$f(x)$        & Function $f: \Set A \to \Set B$ \\
$\ve f(x)$    & Multivariate function $\ve f: \Set A \to \Set B^n$ \\
$\dom f(x)$   & The \textit{domain} $\Set A$ of function $f: \Set A \to \Set B$ \\
$\grad f(x)|_{x_0}$  & The \textit{gradient} of function $f: \Set A \to \Set B$ with respect to \\
              & its argument $x \in \Set A$ at position $x_0$: 
                $\grad f(x)|_{x_0} = \frac{\partial f(x)}{\partial x}|_{x_0}$
\end{tabular}

\vspace{2cm}

\textbf{\large Statistics notation}

\begin{tabular}{ll}
$\E[x]$        & The \textit{expectation} of the random variable $x$ with \\
               & distribution function $p(x)$: $\E[x] = \int x \, p(x) \, dx$ \\
$\E_t[f(t)]$   & The expectation over all $n$ realizations of variable $t$, \\
               & normally the empirical mean over index $t$: 
                 $\E_t[f(t)] = \frac{1}{n} \sum_{t=1}^n f(t)$ \\
$\C[x,y]$      & The \textit{covariance} between random variables $x$ and $y$ with \\
               & joint distribution function $p(x,y)$: \\
               & $\C[x,y] = cov(x,y) = \int\int (x-\E[x])\,(y-\E[y])\,p(x,y)\,dx\,dy$ \\
$\C[\ve x]$    & The \textit{covariance matrix} of the random vector $\ve x$, \\
               & i.e. the covariance between all entries: $(\C[\ve x])_{ij} = \C[x_i,x_j]$ \\
$\V[x]$        & The \textit{variance} of the random variable $x$: $\V[x] = \C[x,x]$ \\
$x \sim p(y,\cdot)$ & The random variable $x \in \Set X$ is drawn according to the distribution \\
               & function $p: \Set Y \times \Set X \to [0,1]$, where 
                 $\forall y \in \Set Y: \sum\limits_{x \in \Set X} p(y,x) = 1$
\end{tabular}

\newpage
$ $\vspace{2cm}

\textbf{\LARGE List of figures}
\vspace{0.5cm}

\begin{tabular}{ll}
Figure \ref{fig:method}: Proposed methodology. &
  Page \pageref{fig:method} \\
Figure \ref{fig:policy_and_value}: Example value function. &
  Page \pageref{fig:policy_and_value} \\
Figure \ref{fig:qvalues}: Example Q-value function. &
  Page \pageref{fig:qvalues} \\
Figure \ref{fig:Berkes05a}: Results from Berkes and Wiskott \cite{Berkes05a}. &
  Page \pageref{fig:Berkes05a} \\
Figure \ref{fig:Berkes05b}: Handwritten digits from the MNIST database. &
  Page \pageref{fig:Berkes05b} \\
Figure \ref{fig:FranziusHierarchy}: Experimental setup of  Franzius et al. 
          \cite{Franzius07} &
  Page \pageref{fig:FranziusHierarchy} \\
Figure \ref{fig:Franzius07}: SFA responses of Franzius et al. \cite{Franzius07}. &
  Page \pageref{fig:Franzius07} \\
Figure \ref{fig:Wyss06}: Results from Wyss et al. \cite{Wyss06}. &
  Page \pageref{fig:Wyss06} \\
Figure \ref{fig:FranziusPC}: Artificial and measured place cells. &
  Page \pageref{fig:FranziusPC} \\
Figure \ref{fig:experiment}: Experiment overview. &
  Page \pageref{fig:experiment} \\
Figure \ref{fig:problem_description}: The \textsc{Pioneer 3DX} robot. &
  Page \pageref{fig:problem_description} \\
Figure \ref{fig:illumination}: Illumination dependency of video images. &
  Page \pageref{fig:illumination} \\
Figure \ref{fig:textures}: Wall textures in the experiment. &
  Page \pageref{fig:textures} \\
Figure \ref{fig:setup_overview}: Simulated experiment overview. &
  Page \pageref{fig:setup_overview} \\
Figure \ref{fig:egypt_overview}: Alternative environment overview. &
  Page \pageref{fig:egypt_overview} \\
Figure \ref{fig:preprocessed_images}: Prepared video examples. &
  Page \pageref{fig:preprocessed_images} \\
Figure \ref{fig:preprocessing}: Image preparation overview. &
  Page \pageref{fig:preprocessing} \\
Figure \ref{fig:SvC}: Slowness vs. model complexity. &
  Page \pageref{fig:SvC} \\
Figure \ref{fig:slowness}: Slowness for kernel SFA. &
  Page \pageref{fig:slowness} \\
Figure \ref{fig:sfa_compare}: SFA responses on test sets. &
  Page \pageref{fig:sfa_compare} \\
Figure \ref{fig:nomix}: Unmixed first SFA response. &
  Page \pageref{fig:nomix} \\
Figure \ref{fig:mix}: Mixed second and third SFA response. &
  Page \pageref{fig:mix} \\
Figure \ref{fig:egyptfeats}: Alternative environment SFA responses. &
  Page \pageref{fig:egyptfeats} \\
Figure \ref{fig:pi_overview}: Policy iteration training set and reward. &
  Page \pageref{fig:pi_overview} \\
Figure \ref{fig:egyptreward}: Alternative environment training set and reward. &
  Page \pageref{fig:egyptreward} \\
Figure \ref{fig:pi_overfitting}: Convergence quality for artificial state representation. &
  Page \pageref{fig:pi_overfitting} \\
Figure \ref{fig:pi_nf_sim}: Convergence quality for SFA state representation. &
  Page \pageref{fig:pi_nf_sim} \\
Figure \ref{fig:robotest_sfa}: Test trajectories of the robot. &
  Page \pageref{fig:robotest_sfa} \\
Figure \ref{fig:pi_nf_egypt}: Convergence quality of the alternative environment. &
  Page \pageref{fig:pi_nf_egypt}
\end{tabular}

\newpage
$ $\vspace{2cm}

\textbf{\LARGE List of algorithms}
\vspace{0.5cm}

\begin{tabular}{ll}
Algorithm \ref{alg_least_squares}: Linear least squares regression & 
  Page \pageref{alg_least_squares} \\
Algorithm \ref{alg_ridge_regression}: Linear ridge regression &
  Page \pageref{alg_ridge_regression} \\
Algorithm \ref{alg_kernel_regression}: Kernelized ridge regression &
  Page \pageref{alg_kernel_regression} \\
Algorithm \ref{alg_select_sv}: Greedy support vector selection algorithm &
  Page \pageref{alg_select_sv} \\
Algorithm \ref{alg_lstd}: LSTD &
  Page \pageref{alg_lstd} \\
Algorithm \ref{alg_lsq}: LSQ &
  Page \pageref{alg_lsq} \\
Algorithm \ref{alg_lspi}: LSPI &
  Page \pageref{alg_lspi} \\
Algorithm \ref{alg_control}: Control &
  Page \pageref{alg_control} \\
Algorithm \ref{alg_linearSFA}: Linear SFA &
  Page \pageref{alg_linearSFA} \\
Algorithm \ref{alg_kernelSFA}: Kernel SFA with kernel matrix approximation &
  Page \pageref{alg_kernelSFA} \\
Algorithm \ref{alg_reference_policy}: Reference policy &
  Page \pageref{alg_reference_policy}
\end{tabular}

\tableofcontents

\newpage
\pagenumbering{arabic}
\setcounter{page}{1}

\chapter{Introduction}
Modern robotics faces a major drawback.

Over the last decades, hard- and software has matured
into commercially applicable products.
Nowadays, robotic control is able to cross deserts,
navigate independently on other planets
and even climb stairs on two legs.
That is, if the robot is fine tuned to the course,
operators stand by to correct inevitable jams
and stairs have the right height and form.
Modern robotic control is not very \textit{adaptive}.

The scientific discipline of adaptive control,
on the other hand, is developed and tested
mainly on discrete, small sized toy examples.
Complex sensory input, uncertain and noise afflicted,
is not compatible with these standard methods.
A natural way to merge these two technologies
is to find a \textit{preprocessing} - a process that
strips the essential information from
raw sensor data.

But what \textit{is} essential information?
Surely we can only answer this question in
context of the individual control domain at hand.
But whatever choice we make, we can be certain that
we will forget \textit{something}.
An ideal preprocessing would also \textit{adapt}
to the control domain on its own.

In this thesis we want to investigate the properties
of an unsupervised technique, called 
\textit{slow feature analysis}, as an 
\textit{auto-adaptive preprocessing}
in the domain of environment specific navigational control.

\section{Motivation}
Roughly 35 years ago, biologists discovered a cluster
of cells in the hippocampal area of rodents,
which encode the spatial position of the animal.
Within minutes of seemingly random movement 
in an unknown environment, the cells specialize 
to fire only around one position each.
Moreover, the cell population covers the 
whole accessible area, which lead many scientists
to believe that the so called \textit{place cells}
are a preliminary stage of the rodents navigational control
\cite{Franzius08}.

How these cells adapt, on the other hand,
is still a question open to debate.
Until recently, the only explanation was the
so called \textit{path integration},
a technique in which the rodent integrates
its movement up to estimate the current position.
Since the sense of movement can be flawed,
small mistakes will sum up over time
and have to be corrected by external stimuli,
which have to be identified in the environment, first.
The computational approach to this problem,
used in robotics, is called 
\textit{simultaneous localization and mapping} (SLAM)
and will be discussed as an alternative to our 
approach in section \ref{sec:slam}.

Last year, Franzius et al. \cite{Franzius07} were
able to show that a memoryless feed forward
network is able to learn place cell behaviour.
The network adapts to videos of a head mounted camera
(substitutional for the rodents eyes)
by an unsupervised learning technique
called \textit{slow feature analysis} (SFA).

\gap

\textit{Reinforcement learning} (or \textit{neuro-dynamic programming})
is a method to learn a control based on reward and punishment.
A set of rewarded/punished example movements is
generalized to estimate the expected sum of future rewards
(\textit{value}) at every position and
for every possible action.
Given a current position, the so called \textit{state} of the system,
the control chooses the action that promises the highest value.
Obviously, the efficiency of this approach depends on
how well the value can be estimated,
which in return depends on the coding of
the state.

Place cells provide an intuitive coding for linear
architectures. Weighting every cells output with the
mean value of its active region can summed up approximate
any value function up to a quality depending on the
number of cells.
Therefore, the place cells of Franzius et al. should
be a natural (and biological plausible) preprocessing for
linear value estimators.

As it turns out, properly trained SFA produces a mapping
of video images into corresponding trigonometric polynomials 
in the space of robot positions.
Franzius' place cells were products of an additional
step \textit{independent component analysis} (ICA)
which does not influence the linear approximation quality
and can therefore be omitted.

The goal of this thesis is to formulate this basic idea
into a working procedure and to demonstrate its soundness 
in a real world robot navigation experiment.

\section{Method}
We want to learn a preprocessing using 
\textit{slow feature analysis} (SFA) out
of the video images of a robots head mounted camera.
This preprocessing should represent the robots
position and orientation (its \textit{state})
in a fashion suitable for linear function approximation.
With this state at hand, we want to
use the reinforcement learning method
\textit{policy iteration} (PI) to learn a
control for the robot.

Note that the preprocessing is adapted to one
environment, e.g. one room only, and will
not work anywhere else.
However, the same is true for any
control learned by reinforcement learning,
so SFA fits well within this framework.

Both SFA and PI need an initial random walk, 
crossing the whole environment.
Therefore the robot has to drive around by choosing random actions, first.
The only needed sensor information of this phase 
is the video of a head mounted camera.
However, for the reinforcement method, we also need to record
the \textit{reward} and \textit{punishment} at every step.
As simple task, the robot receives reward 
for entering a \textit{goal area}
and punishment for getting too close to walls.
The resulting control should drive into the
goal area as quickly as possible, 
while keeping its distance to walls.

Secondly, SFA will learn
a series of filters based on the recorded video.
The output of these filters, applied on the initial
video, is used as input of policy iteration.
PI estimates the expected sum of future rewards (value)
for every action and position.

\begin{figure}[t]
\hspace{-0.3cm}
\includegraphics[scale=.4]{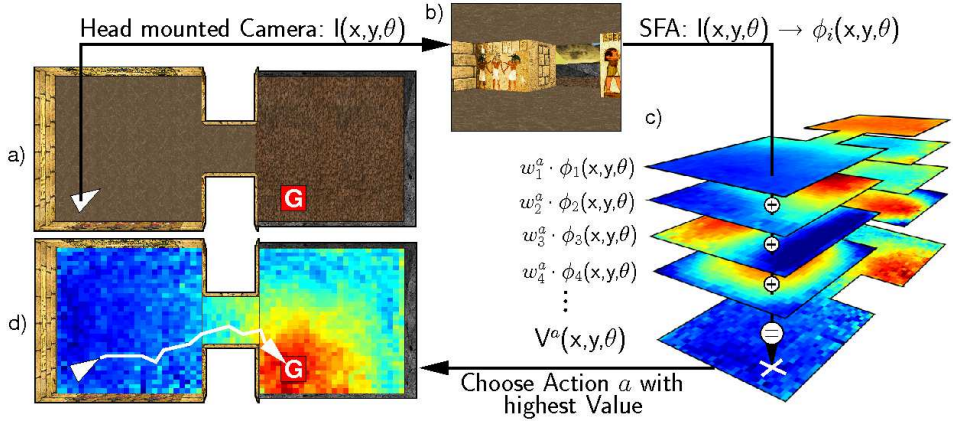}
\begin{center}
\vspace{-.5cm}
\caption{Proposed methodology. See text for a description.}
\label{fig:method}
\end{center}
\vspace{-.5cm}
\end{figure}

\paragraph{Control}
With the value estimator at hand, 
or to be precise its parameter vectors $\ve w^{(1)}, \ldots, \ve w^{(a)}$
(one for every action),
the control works as depicted in figure \ref{fig:method}:
\begin{enumerate}
\item[(a)]
The robot starts in an unknown position and wishes to
navigate into a goal area, marked with a \textbf{G} for 
demonstration purposes.
\item[(b)]
A head mounted video camera shoots a picture $I(x,y,\theta$) 
out of the current perspective.
\item[(c)]
The series of filters $\phi_1 \ldots \phi_n$, 
learned by \textit{slow feature analysis}, 
produce one real valued output $\phi_i(x,y,\theta$), each.
For every possible action,
the output is multiplied by a weight vector $\ve w$,
found by \textit{reinforcement learning}.
The sum of the weighted outputs is the
\textit{value estimation} $V^a(x,y,\theta)$ of this action.
\item[(d)]
The robot executes the action with the highest value
and repeats the procedure until it reaches the goal.
\end{enumerate}

\paragraph{Multiple tasks}
The reinforcement method employed in this thesis, 
\textit{least squares policy iteration} (LSPI),
chooses the most promising in a finite 
number of actions $a$, based on the current state
$\ve x \in \R^d$.
Unfortunately, its complexity is $O(d^3a^3)$
in time and $O(d^2a^2)$ in space.
An efficient kernel SFA algorithm itself has a complexity
of at least $O(m^3)$ in time and $O(m^2)$ in space,
where $m$ is the number of \textit{support vectors}
(comparable to $d$).
So why not use a kernel version of LSPI when
there is no significant advantage to 
the proposed method in computational complexity?

For once, LSPIs complexity also depends on the
number of actions $a$.
An SFA preprocessing can use the available memory up
to $O(m^2)$ to produce a number of filters $d << m$.
This way one can consider a much larger number of actions.
More importantly, the state space extracted by SFA
can be used in more than one reinforcement problem.
For example, when the robot should learn two tasks,
one can learn both given the same initial video
but different rewards.
Therefore, determining the preprocessing once allows to learn
multiple tasks quickly afterwards.

\paragraph{Theory}
In chapter \ref{chap:rl} the theoretical background
of reinforcement learning is presented as well
as the complete derivation of LSPI with all 
necessary algorithms.

Chapter \ref{chapter:sfa} describes slow feature analysis
theoretically. It also contains an overview of recent
applications and algorithms, including a novel
derivation of a kernelized algorithm.

Both chapters are based on common concepts of
machine learning and kernel techniques, which
are introduced for the sake of completeness
in chapter \ref{sec:pre}.

\section{Experiments}
Within the \textsc{NeuRoBot} project, the author was able
to perform experiments with a real robot.
To check the theoretical predictions, we constructed
a rectangular area with tilted tables, in which the
robot should navigate towards some virtual goal area.
We tested two goal areas, which were not marked or
discriminable otherwise besides the reward
given in training.

To evaluate the proposed method less time consuming,
the author also implemented a simulated version
of the above experiment.
At last, theoretical predictions exist only for
rectangular environments.
To test the behaviour outside this limitation,
we used the simulator to create an environment
consisting of two connected rooms.

\gap
A detailed description of these experiments,
as well as a thorough analysis of both SFA
and LSPI under optimal conditions,
can be found in chapter \ref{chapter:empiric}.
The authors conclusions and
an outlook of possible future works 
are presented in chapter \ref{chapter:outlook}.

\chapter{Preliminaries}
\label{sec:pre}
In this chapter we introduce the reader to the
general concepts of machine learning,
especially to \textit{linear models}
(at the example of least squares regression)
in section \ref{sec:regression}
and \textit{kernel techniques} (with some
less common procedures we will
need in this thesis) in
section \ref{sec:kernel}.

We will start with a general introduction
of regression (sec.~\ref{sec:regression} and 
\ref{sec:reg_optimization_problem}),
followed by a discussion of convexity
(sec.~\ref{sec:reg_convexity}),
model choices (sec.~\ref{sec:reg_model_classes})
and validation techniques (sec.~\ref{sec:reg_validation}).
On this basis, we will finally define two 
commonly used linear regression algorithms: 
\textit{least squares regression} and 
\textit{ridge regression}
(sec.~\ref{sec:reg_least_squares}).

In the second part, we will give an overview on
kernel techniques (sec.~\ref{sec:kernel}
and \ref{sec:kernel_trick}), demonstrate them
on a kernel regression algorithm (sec.~\ref{sec:kernel_regression}),
discuss a kernelized version
of the \textit{covariance eigenvalue problem}
(sec.~\ref{sec:kernel_cov}) and end with kernel matrix
approximation methods (sec.~\ref{sec:kernel_approximation})
and a corresponding support vector selection algorithm
(sec.~\ref{sec:kernel_sv_selection}).

\section{Regression}
\label{sec:regression}
We live in a world where things follow \textit{causal relationships},
which should be expressible as 
functions of \textit{observations}
(i.e. products of the real ''causes'').
The exact nature and design of these relationships
are unknown but might be inferred from past observations.
To complicate things even further,
these observations can also be afflicted by noise,
i.e. random distortions independent of the
real ''cause''.
\textit{Regression} deals with inference of functional 
relationships from past observations, based on
some simplifying assumptions.

\begin{assumption}
\label{ass:regression}
The random variable $\ve t \in \Set T$ depends on the
observation $\ve x \in \Set X$ by $\ve t = \ve f(\ve x) + \ve \eta$
with $\ve \eta \in \Set T$ being another zero mean random variable.
\end{assumption}
We assume that some kind of observation
yields an \textit{input sample} $\ve x$, which 
can be assigned a target value by $\ve t$ by 
an expert or process of nature.
Due to errors in this process, $\ve t$ is distorted by
a random variable $\ve \eta$, independent of $\ve x$.

Estimating the volume inside a balloon based on its diameter
would be an example. 
Clearly there is a relationship, but since real balloons do not
follow ideal mathematical shapes, it is hard to find.
Physicists would take measurements of balloon diameters 
(the input samples) and air volumes
(the target values) to find a function which explains
most measurements with a minimum amount of noise.
Since measurements are prone to be inexact and
the rubber of individual balloons differs slightly,
we have to expect some noise in the target values.
An exact reproduction of the observed target values will
therefore reproduce measure errors and mispredict
unseen future inputs.
The physicists approach to this problem is formulated as
\textit{Ockhams razor}, which in short states that
theories have to be simple as well as explanatory.

In the following, we will denote
$\Set S = \{(\ve x^{(1)},\ve t^{(1)}), \ldots, 
(\ve x^{(n)},\ve t^{(n)})\} \in \PM{\Set X \times \Set T}$
as the \textit{training set}.
Regression aims to estimate the function 
$\ve f(\ve x)$ based only on this data.
The balance of Ockhams razor is difficult to maintain
and led to a variety of regression algorithms.


\subsection{Optimization problem}
\label{sec:reg_optimization_problem}
We wish to find a function $\ve y: \Set X \to \Set T$ which
estimates the real (but unknown) relationship $\ve f$.
The only knowledge available is the training set
$\Set S$, so we construct a \textit{cost function} $C$
which defines the optimization goal indirectly
by comparing the target values with the predictions
made by some given function $\ve y$.

\begin{definition}[Cost function]
$ $ \\
A cost function has the form 
$C: (\Set X \to \Set T) \times \PM{\Set X \times \Set T} \to \R$.
\end{definition}
The intuition is that the cost function is small if
$\ve f$ is applied, and hopefully also small for
functions similar to $\ve f$.
Minimizing $C$ with respect to $\ve y$
should therefore lead to a similar function.

The approach suggesting itself is to evaluate 
$\ve y$ at every sample $\ve x^{(i)}$
and consider the distance under some norm $|| \cdot ||$ 
to target $\ve t^{(i)}$ as the \textit{empirical cost} for
this sample.
Because we have no reason to favor any sample,
we sum up the individual costs and the result is called 
the \textit{empirical cost function} \cite{Bishop06}:
\begin{equation}
\label{eq:regression_cost_function}
C^{emp}(\ve y, \Set S )
= \sum\limits_{j=1}^n ||\ve y(\ve x^{(j)}) - \ve t^{(j)} ||
\end{equation}
Note that, due to the norm, the noise term $\eta$ in assumption
\ref{ass:regression} will not cancel out.
The only way for $C$ to reach zero is to reproduce
the noise of $\ve t$ in $\ve y$,
which is not desirable.

If we wish to avoid this problem, we can invoke
Ockhams razor and penalize complex functions 
with a \textit{regularization} term $R(\ve y) \in \R$.
A less complex function will be less likely to
reproduce the observed noise and 
therefore predict unseen data better.
Examples for $R$ would be 
the length of the parameter vector (linear functions),
the number of support vectors (kernel machines)
or the number of hidden neurons (neuronal networks).
Together with the empirical cost we obtain the 
\textit{regularized cost function} \cite{Bishop06}:
\begin{equation}
\label{eq:regression_cost_function_reg}
C^{reg}(\ve y, \Set S)
= \sum\limits_{j=1}^n ||\ve y(\ve x^{(j)}) - \ve t^{(j)} || + R(\ve y)
\end{equation}
Note that the regularization term $R(\ve y)$ is independent of 
the sample size $n$ and with growing $n$ will loose relevance.
Section \ref{sec:reg_validation} will discuss the effects
of sample size in more detail.

\begin{definition}[Optimal function]
\label{def:opt_function}
The global minimum of a cost function $C(\ve y, \Set S)$ 
with respect to $\ve y$ is called optimal function $\ve y^*$ of $C$:
\begin{equation}
\ve y^* = \argmin\limits_{y} C(\ve y, \Set S)
\end{equation}
\end{definition}
The optimal function $\ve y^*$ of $C$ 
is the function we were looking for
and should resemble $\ve f$ at least on the training set.
Arbitrary functions can not be represented in a
computer, so a common approach chooses a parameterized function
class $\Set F \subset (\Set X \to \Set T)$ to pick 
$\ve y = \Set F(\ve w)$ from.

One approach to find the optimal function is to
calculate the gradient of $C$ with respect to the 
function class parameters $\ve w$ 
at an arbitrary start position $\ve w_0$.
One can follow the negative gradient $-\grad C|_{w_0}$ 
in small steps, leading to smaller costs 
if the step size is well chosen.
This method is called \textit{gradient descent} \cite{Haykin98}.

The disadvantage is that the method will
stop at every $\ve w$ that fulfills
$\grad C|_{w} = 0$, called an 
\textit{extrema}\footnote{
If also $\grad\grad C|_{w} > 0$
we speak of a minimum, but not necessarily
a global minimum.} of $C$.
Depending on the considered function class $\Set F$
the cost function can have multiple extrema
with respect to the parameters $\ve w$, and
not all have to be global minima after definition
\ref{def:opt_function}.
In other words, gradient descent will
converge to \textit{whatever} extrema the 
gradient lead to, starting at $\ve w_0$.
This means we can never be sure that we reached a
global minimum, 
if no closed solution of $\grad C$ is possible.

Keeping this in mind, it is wise to choose
$\Set F$ such that every extrema of
the cost function is a global minimum,
ideally the only one.
In this thesis we will focus on convex functions.


\subsection{Convexity}
\label{sec:reg_convexity}
There are other functions without local minima,
but convex functions provide other desirable
mathematic properties (see \cite{Boyd04} for details).
In some special cases it is even possible
to derive a closed analytical solution
(section \ref{sec:reg_least_squares}).

\begin{definition}[Convex function \cite{Boyd04}]
A function $f: \R^n \to \R$ is convex if
$\dom f$ is a convex set and if for all
$x,y \in \dom f$, and $\theta$ with
$0 \leq \theta \leq 1$, we have
\begin{equation}
\label{eq:convex_function}
f(\theta x + (1-\theta)y) \leq \theta f(x) + (1-\theta) f(y).
\end{equation}
\end{definition}
We can tighten the definition further.
When the strict inequality in equation
\ref{eq:convex_function} holds, we speak
of a \textit{strictly convex function}.
These have the additional advantage that the
minimum is \textit{unique},
i.e. no other extrema exists.

Anyway, the main property we are interested 
in both cases is that every extrema is a global minimum:
\begin{proposition}[First order convexity condition \cite{Boyd04}]
\label{the:first_order_convex}
Suppose $f$ is differentiable (i.e. its gradient
$\grad f$ exists at each point in $\dom f$,
which is open). Then $f$ is convex if and only if
$\dom f$ is convex and
\begin{equation}
f(y) \geq f(x) + \grad f(x)^\top(y-x)
\end{equation}
holds for all $x,y \in \dom f$.
\end{proposition}
\proof{See \cite{Boyd04}.}\gap

Every extrema $x$ of $f$ satisfies $\grad f(x) = 0$.
According to proposition \ref{the:first_order_convex},
if $f$ is convex then $\forall y \in \dom f: f(y) \geq f(x)$,
which identifies $x$ as a global minimum of $f$.
\gap

However, we do not necessarily wish the functions
$\ve y \in \Set F$ but the discussed cost functions 
to be convex:
\begin{proposition}[Convex cost function]
Let $\Set S \in \PM{\Set X \times \Set T}$
be arbitrary but given.
If $R(\ve y)$ and all $\ve y \in \Set F$ 
are convex functions then the cost functions 
(eq.~\ref{eq:regression_cost_function}) and
(eq.~\ref{eq:regression_cost_function_reg})
are convex in the domain $\Set F$.
\end{proposition}
\proof{$\ve y(\ve x^{(j)})$ and $-\ve t^{(j)}$
are convex, and how lemma \ref{lem:convex_sum} shows,
the sum of them is convex too.
Lemma \ref{lem:convex_norms} proves that norms
are convex, so their sum has to be, too.
At last, $C^{emp}$ and $R(\ve y)$ are convex,
and so is their sum $C^{reg}$. $\Box$
}

\begin{lemma}[Pointwise sum \cite{Boyd04}]
\label{lem:convex_sum}
The pointwise sum of two convex functions $f_1$ and $f_2$,
$f = f_1 + f_2$ with $\dom f = \dom f_1 \cap \dom f_2$,
is a convex function.
\end{lemma}
\proof{ Insert $(\theta x + (1-\theta)y)$ into
$f_1(\cdot) + f_2(\cdot)$ and apply (eq.~\ref{eq:convex_function}) two times.
$\Box$}

\begin{lemma}[Norms \cite{Boyd04}]
\label{lem:convex_norms}
Every norm on $\R^n$ is convex.
\end{lemma}
\proof{If $f: \R^n \to \R$ is a norm, 
and $0 \leq \theta \leq 1$, then
\begin{equation*}
f(\theta \ve x + (1-\theta) \ve y)
\leq f(\theta \ve x) + f((1-\theta) \ve y)
= \theta f(\ve x) + (1-\theta) f(\ve y).
\end{equation*} 
The inequality follows from the triangle inequality,
and the equality follows from homogeneity of a norm.
$\Box$}


\subsection{Function classes}
\label{sec:reg_model_classes}
Convexity of the cost function depends on its formulation
and the considered function class 
$\Set F \subset (\Set X \to \Set T)$.
The family of convex functions is large
(see \cite{Boyd04} for examples).
However, we want to introduce two classes
that will play a mayor role in this thesis
because they allow a closed analytical
solution under the squared $L_2$ norm.

\begin{definition}[Linear function \cite{Boyd04}]
A function $\ve f: \R^p \to \R^q$ is linear
if for all $\ve x, \ve y \in \R^p$ and $\alpha,\beta \in \R$
it satisfies the condition
\begin{equation}
\label{eq:lin_function}
\ve f(\alpha \ve x + \beta \ve y) = \alpha \ve f(\ve x) + \beta \ve f(\ve y).
\end{equation}
\end{definition}
Linear functions have nice analytical properties, 
e.g. they can be uniquely determined by a matrix
$\mat W \in \R^{p \times q}$: $\ve f(\ve x) = \mat W^\top \ve x$.
Convexity be seen by comparison of (eq.~\ref{eq:lin_function}) 
and (eq.~\ref{eq:convex_function}).

\begin{definition}[Affine function \cite{Boyd04}]
An affine function $f:\R^p \to \R^q$ is the sum of a linear function and a constant:
$\ve f(\ve x) = \mat W^\top \ve x + \ve b$.
\end{definition}
The \textit{bias} $\ve b$ violates the linear condition,
but affine functions share many properties with
linear functions.
They can express more relationships, however.
For example, a line that does not cross the 
origin can be expressed as an affine, but
not as a linear function.

By extending the input vector $\ve x$ by a constant
(i.e. $x_0=1$) one can express an affine function
as a linear function.
This trick can be applied if the algorithm at hand
requires linearity but the data can only
be properly predicted by an affine function.

\begin{lemma}
Affine functions are convex.
\end{lemma}
\proof{
We show convexity for every component 
$f_i(\ve x) = \ve w_i^\top \ve x + b_i$.
\antigap
\begin{eqnarray*}
f_i(\theta \ve x + (1-\theta)\ve y)
&=& \ve w_i^\top(\theta \ve x + (1-\theta)\ve y) + b_i\\
&=& \theta \, \ve w_i^\top \ve x + (1-\theta) \, \ve w_i^\top \ve y 
    + (1 - \theta +\theta) \, b_i\\
&=& \theta \, f_i(\ve x) + (1-\theta) \, f_i(\ve y)
\end{eqnarray*} 

\antigap\antigap\antigap
\begin{flushright}$\Box$\end{flushright}
}


\subsection{Validation}
\label{sec:reg_validation}
In regression, we do not know the real target function 
$\ve f$ of assumption \ref{ass:regression},
but aim to find a ''similar'' function 
$\ve y \in \Set F$.
Since we can not construct a similarity
measure between $\ve f$ and $\ve y$ directly,
we define a cost function instead.
Though one can obtain an $\ve y^*$ optimal 
to the cost function,
the approach is susceptible to many 
sources of error:
\begin{itemize}
\item
The cost function can be chosen differently,
leading to different optimal functions.
Here the choice of regularized vs. empirical
arises as well as the choice of the regularization term.
Anyway, the two presented equations are not the
only imaginable cost functions 
(see \cite{Haykin98} for more).
\item
The considered function class $\Set F$ might not be
suited for the data at hand. 
On the one hand, $\Set F$ can be too \textit{weak}, 
e.g. a parabola can not be represented
by a linear function.
If, on the other hand, the \textit{sample size} $n$
is too small, $\ve y^*$ can reduce the individual
costs of all training samples to zero, 
but take unreasonable values between them.
This effect is called \textit{over-fitting} \cite{Bishop06}.
\item
The training sets \textit{sampling} might introduce errors.
If a region of $\Set X$ is left out, $\ve f$ can not
be estimated there.
More general, \textit{unbalanced sampling} leads
to a good estimation of $\ve f$ where many samples are available,
but a poor estimation where this is not the case.
The sum of the individual costs
tolerates big errors in rarely sampled regions,
if it means to shrink the error 
in highly sampled regions even slightly.
This effect is also influenced by the choice of $\Set F$.
Especially linear and affine functions are known to react
badly to unbalanced sampling \cite{Bishop06}.
\end{itemize}

The last point can be circumnavigated if one can ensure
\textit{identical and independent distributed} (iid)
sampling. This way, given enough samples, the training set
will be sampled homogeneously from $\Set X$.
However, in most practical cases one can not give
such a guaranty.

The common procedure to validate $\ve y^*$ is to
withhold a \textit{test set} 
$\Set S' \in \PM{\Set X \times \Set T}, \Set S' \cap \Set S = \emptyset$ 
from the training.
A comparison of the normalized empirical costs
of training and test set
(called \textit{training} and \textit{test error})
demonstrates how well $\ve y^*$ \textit{generalizes}.
If the errors are approximately the same, the optimal
function predicts unseen inputs obviously as well 
as the training samples.
However, if the test error is 
significantly higher than the training error,
$\ve y^*$ is probably over-fitting.
Counter measures would include more training samples,
a less complex function class or stronger
regularization.

Another useful test is the computation of $\ve y^*$
for the same cost function and training data,
but a different function class $\Set F$.
The comparison of the test and training errors
can tell us which function class is better suited.
Especially if one can choose the complexity in
a family of function classes (e.g. number
of support vectors in kernel machines),
it is interesting at which complexity the
error saturates.
Finding this point (i.e. the simplest model
which estimates $\ve f$ well) is sometimes referred
to as \textit{model selection} or
\textit{model comparison} \cite{Bishop06}.

\begin{algorithm}[b]
\caption{Linear Least Squares Regression}
\label{alg_least_squares}
\begin{algorithmic}
\REQUIRE $\ve x^{(1)}, \ldots, \ve x^{(n)} \in \R^p; 
          \ve t^{(1)}, \ldots, \ve t^{(n)} \in \R^q$ 
\STATE $\mat C_0$ = zeros($p$,$p$)
\STATE $\mat B_0$ = zeros($p$,$q$)
\FOR{$i=1,\ldots,n$}
  \STATE $\mat C_i$ = $\mat C_{i-1} + \ve x^{(i)} \ve x^{(i)\top} $
  \STATE $\mat B_i$ = $\mat B_{i-1} + \ve x^{(i)} \ve t^{(i)\top}$
\ENDFOR
\STATE $\mat W$ = inv$(\mat C_n) \mat B_n$ 
\STATE \textbf{return} $\mat W$
\end{algorithmic}
\end{algorithm}


\subsection{Linear regression algorithms}
\label{sec:reg_least_squares}

The \textit{least squares regression problem} applies the
squared $L_2$ norm and the class of linear functions
on the empirical cost function 
(eq.\ref{eq:regression_cost_function}):
\begin{equation}
\label{eq:least_squares_cost_function}
C^{emp}(\ve y,\Set S) 
= \sum\limits_{j=1}^n ||\ve y(\ve x^{(j)}) - \ve t^{(j)} ||_2^2
= || \mat W^\top \mat X - \mat T ||_\frobenius^2
\end{equation}
where $|| \cdot ||_\frobenius^2$ is the squared Frobenius norm
and the matrices $\mat X_{ij} = x_i^{(j)}$ and $\mat T_{ij} = t_i^{(j)}$
are introduced for simplicity.

Extending the input samples by a constant (i.e. $x_0 = 1$)
allows us to use the same formulation for 
the class of affine functions.

To find the optimal function $\ve y^*(\ve x) = \mat W^{*\top} \ve x$
we set the derivation of equation 
\ref{eq:least_squares_cost_function}
with respect to the parameter 
matrix $\mat W$ to zero:
\begin{eqnarray}
\label{eq:least_squares_solution}
\frac{\partial C^{emp}(\mat W, \Set S)}{\partial \mat W} 
&=& 2  \mat X \mat X^\top \mat W - 2 \mat X \mat T^\top 
\shallbe 0 \nonumber \\
\Rightarrow \mat W^* &=& (\mat X \mat X^\top)^{-1} \mat X \mat T^\top
\end{eqnarray}
which holds, providing the rows of $\mat X$ are linearly independent
and the covariance matrix $\mat X \mat X^\top$ therefore of full rank.

\gap
We can also use the regularized cost function
(eq.~\ref{eq:regression_cost_function_reg}) with
the regularization term 
$R(\ve y) = R(\mat W) = \lambda\,\trace(\mat W^\top \mat W)$
which penalizes the squared $L_2$ norm of the column vectors of $\mat W$.
$\lambda \in \R$ regulates how strong the penalty for
complex functions is.
\begin{eqnarray*}
\frac{\partial C^{reg}(\mat W, \Set S)}{\partial \mat W} 
&=& 2 \mat X \mat X^\top \mat W - 2 \mat X \mat T^\top + 2 \lambda \mat W
\shallbe 0 \nonumber \\
\Rightarrow \mat W^* &=& (\mat X \mat X^\top + \lambda \mat I)^{-1} \mat X \mat T^\top
\end{eqnarray*}
Regression with this regularization term is known as
\textit{ridge regression} \cite{Bishop06} or
in the context of neural networks as \textit{weight decay} \cite{Haykin98}.

\begin{algorithm}[b]
\caption{Linear Ridge Regression}
\label{alg_ridge_regression}
\begin{algorithmic}
\REQUIRE $\ve x^{(1)}, \ldots, \ve x^{(n)} \in \R^p; 
          \ve t^{(1)}, \ldots, \ve t^{(n)} \in \R^q; \lambda \in \R$
\STATE $\mat C_0$ = zeros($p$,$p$)
\STATE $\mat B_0$ = zeros($p$,$q$)
\FOR{$i=1,\ldots,n$}
  \STATE $\mat C_i$ = $\mat C_{i-1} + \ve x^{(i)} \ve x^{(i)\top} $
  \STATE $\mat B_i$ = $\mat B_{i-1} + \ve x^{(i)} \ve t^{(i)\top}$
\ENDFOR
\STATE $\mat W$ = inv$(\mat C_n + \lambda \mat I) \mat B_n$ 
\STATE \textbf{return} $\mat W$
\end{algorithmic}
\end{algorithm}

Both algorithms have a complexity of 
$O(p^2)$ in space and max$(O(np^2),O(p^3)$) in time.
The $O(p^3)$ term originates in the matrix inversion.


\section{Kernel techniques}
\label{sec:kernel}
Realistic processes can seldom be approximated sufficiently
with linear function classes.
Classes of nonlinear functions (e.g. neural networks) 
could provide satisfactory results,
but their cost functions are rarely convex
and the optimization therefore complicated.

Another approach is the nonlinear expansion of the
input samples $\ve x \in \Set X$.
This way, projected into a high dimensional \textit{feature space},
the problem can be treated as linear, while the solution
in original space $\Set X$ is nonlinear.
The classic example to demonstrate this is the XOR problem.

In the XOR problem, the input $\ve x \in \{0,1\}^2$ and the 
target $t \in \{0,1\}$ are connected by a simple rule:
If both entries of $\ve x$ are the same then $t = 0$,
otherwise $t = 1$.
It is easy to see that there is no linear function
$y(\ve x) = \ve w^\top \ve x$ that
solves this problem,
but if we \textit{expand} the input vector 
$\ve x' = [x_1, x_2, x_1 x_2]^\top$,
the parameter vector $\ve w = [1, 1, -2]^\top$ 
explains the relationship perfectly.
Thus, with the suggested expansion, the XOR problem
is \textit{linearly solvable}.

Expansion almost always increases the
dimensionality of the input.
The fact that one normally does not know the
perfect expansion beforehand
rarely lead to feasible expansions
that solve the problem.
In the case that a suitable expansion size
would outnumber the number of training samples,
the kernel trick can be employed.


\subsection{The kernel trick}
\label{sec:kernel_trick}

The kernel trick defines the nonlinear expansion
indirectly. One exploits the \textit{representer theorem}, 
which states that the optimal function of a cost function
can be represented as scalar products of the training set.

Kernels are a broad class of functions, which
are equivalent to a scalar product in a corresponding
vector space.
The intuition of the kernel trick is to exchange
the euclidian scalar product by one in a high dimensional
nonlinear feature space, i.e. another kernel.

Choosing a nonlinear kernel is equivalent
to a nonlinear expansion of the input vector.
This way one can handle huge feature spaces.
The drawback is that the optimal function 
depends on scalar products to all training samples,
which can be infeasible for large sample sizes.

In the following we introduce the elements 
of kernel methods used in this thesis.
To follow the derivation path in full length,
the reader is referred to \cite{Schoelkopf02}.

\begin{definition}[(Positive definite) kernel \cite{Schoelkopf02}]
Let $\Set X$ be an nonempty set. A function $k$ on $\Set X \times \Set X$
which for all $n \in \N$ and all $x^{(1)}, \ldots, x^{(n)} \in \Set X$
gives rise to a positive definite Gram matrix is called a positive
definite kernel.
\end{definition} 
\antigap
For a definition of \textit{Gram matrix} and 
\textit{positive definite matrix}, see \cite{Schoelkopf02}.
\gap

One can prove that \textit{every} positive definite kernel 
function uniquely implies
a scalar product $\langle \psi(x), \psi(x') \rangle = k(x,x')$
with $\psi(x)$ being the projection of $x \in\Set X$
into another space.

To be more exact, one can show that a kernel function
implies a \textit{reproducing kernel Hilbert space} (RKHS) 
of functions $f: \Set X \to \R$ with a scalar product.
With a detour over \textit{Mercer kernels} 
(which are equivalent to positive definite kernels)
one can show that it is possible to construct
a mapping $\psi(\cdot)$ for which $k(\cdot,\cdot)$ acts as
a dot product.

In the following we will refer to an arbitrary kernel function 
$k(\cdot, \cdot) = \langle \psi(\cdot), \psi(\cdot) \rangle_{\Set H}$ 
in the corresponding RKHS $\Set H$.

\begin{theorem}[Representer Theorem \cite{Schoelkopf02}]
\label{representer_theorem}
Denote by $R: [0,\infty) \to \R$ a strictly monotonic
increasing function, by $\Set X$ a set and by
$C: (\Set X, \R, \R)^n \to \R \cup \{\infty\}$ an
arbitrary loss function. Then each minimizer $y: \Set X \to \R$
of the regularized risk
\begin{equation}
C((\ve x^{(1)}, t^{(1)},  y(\ve x^{(1)})),
\ldots, (\ve x^{(n)}, t^{(n)}, y(\ve x^{(n)})))
+ R(||y||_{\Set H})
\end{equation}
admits a representation of the form
\begin{equation}
y(\ve x) = \sum\limits_{i=1}^n \alpha_i k(\ve x^{(i)}, \ve x).
\end{equation}
\end{theorem} \gap
\proof{See \cite{Schoelkopf02}.
Note that a \textit{loss function} is a slightly
different defined cost function and
\textit{regularized risk} a cost function
with a regularization term.} \gap 

Multivariate functions $\ve y: \R^p \to \R^q$
can be represented component wise
$y_j(\ve x) = \sum_{i=1}^n A_{ij} k(\ve x^{(i)}, \ve x)$,
giving rise to a new parameter matrix $\mat A \in R^{n \times q}$:
\begin{equation}
\label{eq:kernel_representation}
\ve y(\ve x) = \mat A^\top \ve k(\ve x)
\end{equation}
where $\ve k(\ve x) = [k(\ve x^{(1)},\ve x), \ldots, 
k(\ve x^{(n)}, \ve x)]^\top$ is called a 
\textit{kernel expansion} of $\ve x$.

\begin{remark}[''Kernel trick'' \cite{Schoelkopf02}]
Given an algorithm which is formulated in terms
of a positive definite kernel $k$, one can construct an
alternative algorithm by replacing $k$ by another
positive definite kernel.
\end{remark}
The kernel trick allows the replacement of
scalar products (which are positive definite
kernels) by complex, nonlinear
kernels.
One can take this replacement as
a projection $\psi(\cdot)$ into a high dimensional
feature space. 
In this space, we can
solve our optimization problem linear.
The only restriction is that the original algorithm 
must be formulated entirely with scalar products.

\newpage

\begin{algorithm}[t]
\caption{Kernelized Ridge Regression}
\label{alg_kernel_regression}
\begin{algorithmic}
\REQUIRE $\ve x^{(1)}, \ldots, \ve x^{(n)} \in \R^p; 
          \ve t^{(1)}, \ldots, \ve t^{(n)} \in \R^q; 
          \lambda \in \R; k: \R^p \times \R^p \to \R$
\STATE $\mat K$ = zeros($n$,$n$)
\FOR{$i=1,\ldots,n$}
  \FOR{$j=1,\ldots,n$}
    \STATE $\mat K_{ij}$ = $k(\ve x^{(i)},\ve x^{(j)}) $
  \ENDFOR
\ENDFOR
\STATE $\mat A$ = inv$(\mat K + \lambda \mat I) \mat T^\top$ 
\end{algorithmic}
\end{algorithm}

\subsection{Kernelized regression}
\label{sec:kernel_regression}
To demonstrate the kernel trick, we will derive a
kernelized version of the linear ridge regression
algorithm.

First we have to notice that there are no 
scalar products in
algorithm \ref{alg_ridge_regression}.
Therefore, it is necessary  to reformulate the
algorithm \cite{Bishop06}. 
Again, we start by setting 
the derivation of $C^{reg}$ with respect
to $\mat W$ to zero:
\begin{eqnarray}
\frac{\partial C^{reg}(\mat W, \Set S)}{\partial \mat W} 
&=& 2 \mat X \mat X^\top \mat W - 2 \mat X \mat T^\top + 2 \lambda \mat W
\shallbe 0 \nonumber \\
\label{eq:krr_XA}
\Rightarrow \mat W^* &=& - \frac{1}{\lambda} (\mat X \mat X^\top \mat W^* 
                - \mat X \mat T^\top) = \mat X \mat A
\end{eqnarray}
with $\mat A = -\frac{1}{\lambda} (\mat X^\top \mat W^* - \mat T^\top)$.
Next we will clear $\mat A$ of its dependency on $\mat W^*$.
Substituting (eq.~\ref{eq:krr_XA}) in $C^{reg}$ and 
the derivation with respect to $\mat A$ yields:
\begin{eqnarray*}
C^{reg}(\mat A, \Set S) 
&=& \trace((\mat A^\top \mat X^\top \mat X - \mat T)^2)
= \trace((\mat A^\top \mat K - \mat T)^2) \\
\frac{\partial C^{reg}(\mat A, \Set S)}{\partial \mat A}
&=& 2 \mat K \mat K^\top \mat A - 2 \mat K \mat T^\top 
    + 2 \lambda \mat K \mat A
  \shallbe 0\\
\Rightarrow \mat A^* &=& (\mat K + \lambda \mat I)^{-1} \mat T^\top
\end{eqnarray*}
where the Gram matrix of scalar products 
$\mat K_{ij} = \ve x^{(i)\top} \ve x^{(j)} \in \R^{n \times n}$
can be replaced by any other \textit{kernel matrix}
$\mat K_{ij}' = k(\ve x^{(i)}, \ve x^{(j)})$.
Prediction follows (eq.~\ref{eq:kernel_representation}):
\begin{equation}
\ve y(\ve x) = \mat W^\top \ve x
= \mat {A^*}^\top \mat X^\top \ve x = \mat {A^*}^\top \ve k(\ve x)
\end{equation}

The complexity of this algorithm is $O(n^2)$ in space
and $O(n^3)$ in time.

\newpage

\subsection{Kernels}
\label{sec:kernels}
Since the kernel implies the feature space, 
choosing a nonlinear expansion is equivalent
to choosing a kernel.
Therefore it is necessary to know common
kernel functions and their properties.

\paragraph{Polynomial kernels}
As we have seen in the XOR example, 
a polynomial expansion can be useful. 
The direct approach 
would be to collect all multivariate monomials up
to \textit{degree} $d$ in a vector.
For $p$ dimensional input vectors, the expanded 
vector would have dimensionality
$\left( {d + p-1}\atop{d} \right) = \frac{(d+p-1)!}{d!(p-1)!}$.

The corresponding kernel function
\begin{equation}
k(\ve{x},\ve{x'}) = (\ve{x}^\top \ve{x'} + 1)^d
\end{equation}
projects into the same space of
polynomials with degree $\leq d$.
To demonstrate this let us consider the 
following example \cite{Bishop06}:
The input vectors shall be $\ve x, \ve x' \in \R^2$.
The polynomial kernel function of degree 2 for
those two is:
{\small \begin{eqnarray*}
k(\ve x, \ve x') 
&=& (\ve x^\top \ve x' + 1)^2 \\
&=& (x_1 x'_1 + x_2 x'_2 + 1)^2 \\
&=& (x_1 x'_1)^2 + (x_2^2 x'_2)^2 + 2 x_1 x'_1 x_2 x'_2 
  + 2 x_1 x'_1 + 2 x_2 x'_2 + 1 \\
&=& [x_1^2, x_2^2, \sqrt{2} \, x_1 x_2,\sqrt{2} 
      \, x_1, \sqrt{2}\, x_2, 1] \;
    [(x'_1)^2, (x'_2)^2, \sqrt{2} \, x'_1 x'_2, 
      \sqrt{2} \, x'_1, \sqrt{2} \, x'_2, 1]^\top \\
&:=& \ve \psi(\ve x)^\top \ve \psi(\ve x')
\end{eqnarray*} }
The projection $\psi(\ve x)$ defined this way spans
the space of polynomials of degree 2 and has
therefore dimensionality 6.

Polynomial kernels are all about the euclidian \textit{angle}
between two inputs.
Additional, the euclidian \textit{length} of both vectors play a role.
If one likes to get rid of the last effect, 
a \textit{normalized polynomial kernel} 
returns only the cosine between $\ve x$ and 
$\ve x'$ to the power of $d$:
$k(\ve x, \ve x') = (\frac{\ve x^\top \ve x'}{||x|| \, ||x'||})^d$.
The induced feature space is of lower dimensionality, but 
the kernel reacts less extreme to large input vectors.

\paragraph{RBF kernels}
A \textit{radial basis functions} (RBF) kernel
has the general form of
\begin{equation}
k(\ve{x},\ve{x'}) = exp\left(- ({||\ve{x}-\ve{x'}|| \over \sqrt{2} \sigma})^d\right).
\end{equation}
Different norms $|| \; ||$ and parameters $d$
and $\sigma$ generate different kernels,
which all imply an infinite dimensional RKHS
\cite{Schoelkopf02}.
The two most popular both use the $L_2$ norm and
are called Laplacian ($d=1$) and Gaussian ($d=2$).
The latter is by far the most common kernel
and has become synonymous with the name RBF kernel.

RBF kernels are based on the distance of two input vectors.
The kernel parameter $\sigma$ determines the radius 
of influence for training samples.
In well sampled regions (and well adjusted $\sigma$) 
the kernel shows good approximation properties but
where no training data is available the kernels
will produce only small output and therefore
perform poorly.

\subsection{Kernelized covariance eigenvalue problem}
\label{sec:kernel_cov}
Many unsupervised learning algorithms 
(e.g. PCA \cite{Shawe04}) demand an
eigenvalue decomposition of the covariance matrix:
\begin{equation}
\label{eq:original_covariance}
cov(\mat X) 
= \E[\ve x \ve x^\top] - \E[\ve x] \E[\ve x]^\top
= \frac{1}{n} \mat X \mat X^\top 
  - \frac{1}{n^2}\mat X \ve 1 \ve 1^\top \mat X^\top
= \mat U_c \mat \Lambda_c \mat U_c^\top
\end{equation}
In most cases one aims to project samples $\ve x$ onto
eigenvectors $\mat U_c$: $\ve x' = \mat U_c^\top \ve x$.
To catch more complex, nonlinear relationships,
one can expand $\ve x$ into a nonlinear
feature space $\ve \psi(\ve x)$.
The kernel trick can help to describe the
expansion more efficient 
(e.g. Kernel PCA \cite{Schoelkopf97}).

\paragraph{Original approach}
One starts with ensuring zero mean by
subtracting the sample mean from all columns
$\mat X_c = \mat X - \frac{1}{n} \mat X \ve 1 \ve 1^\top$.
Subsequently, one can perform the eigenvalue decomposition
on the centered covariance matrix 
$\frac{1}{n}\mat X_c \mat X_c^\top = \mat U_c \mat \Lambda_c \mat U_c^\top$.
Standard implementations of eigenvalue decompositions have a
complexity of $O(p^3)$ with $p$ being the dimensionality 
of samples $\ve x$.

\paragraph{Inner and outer product \cite{Shawe04}}
If we want to reformulate the covariance
matrix eigenvalue decomposition,
we can exploit a relationship between the 
\textit{inner product} $\mat X^\top \mat X$ and
\textit{outer product} $\mat X \mat X^\top$
of $\mat X$.

The \textit{singular value decomposition}
$\mat X = \mat U \mat \Sigma \mat V^\top$
with $\mat \Sigma = [\mat \Lambda^{1/2} \, \mat 0]$
refers to eigenvalues and eigenvectors of both 
outer and inner product.
Therefore, an eigenvalue decomposition of the inner product
$\mat X^\top \mat X = \mat V \mat \Lambda \mat V^\top$
produces the same non zero eigenvalues 
$\mat \Lambda_r$ as the outer product 
and the corresponding eigenvectors $\mat U_r$
and $\mat V_r$
are related by:
\begin{equation}
\label{eq:eigenvector_relationship}
\mat U_r = \mat X \mat V_r \mat \Lambda_r^{-1/2}
\end{equation}

\paragraph{Kernelized approach}
We proceed as before.
With the Gram matrix of scalar products
$\mat K = \mat X^\top \mat X$ at hand,
we first have to center the data represented by it.
Subtracting the sample mean 
$\mat X_c = \mat X - \frac{1}{n} \mat X \ve 1 \ve 1^\top$,
it is easy to proof that the centered 
kernel matrix is:
\begin{equation}
\mat K_c = \mat X_c^\top \mat X_c 
= (\mat I - \frac{1}{n} \ve 1 \ve 1^\top) \mat K 
  (\mat I - \frac{1}{n} \ve 1 \ve 1^\top)
\end{equation}

Secondly, we perform the eigenvalue decomposition
$\mat K_c = \mat V_c \mat \Lambda_c \mat V_c^\top$. 
Using (eq.~\ref{eq:eigenvector_relationship}) we
can now express the term $\mat U_c^\top \ve x$
with the nonzero eigenvalues $\mat \Lambda_r$ 
and corresponding eigenvectors $\mat V_r$ of $\mat K_c$:
\begin{equation}
\label{eq:kernel_eigenvector_projection}
\mat U_c^\top \ve x 
= \sqrt{n}\, \mat \Lambda_r^{-1/2}\, \mat V_r^\top \mat X^\top \ve x
\end{equation}
Replacing $\mat K$ by another kernel matrix $\mat{\tilde K}$
and the term $\mat X^\top \ve x$ in 
(eq.~\ref{eq:kernel_eigenvector_projection}) by $\tilde{\ve{k}}(\ve x)$,
we implicitly project the data into a
feature space $\psi(\ve x)$ defined by the kernel:
\begin{equation}
\mat{\tilde U}_c^\top \psi(\ve x)
= \sqrt{n}\, \mat{\tilde \Lambda}_r^{-1/2}\, 
  \mat{\tilde V}_r^\top \tilde{\ve k}(\ve x)
\end{equation}


\subsection{Kernel matrix approximation}
\label{sec:kernel_approximation}
The application of the kernel trick is restricted to 
a small amount of samples ($n \approx 5000 \ldots 10000$),
because the kernel matrix stores $n \times n$ entries.
If this number is exceeded, one would like
to sacrifice approximation quality for 
manageable kernel matrix sizes.
Different approaches have been made to overcome this problem 
by means of an approximation $\hat{\mat{K}} \in \R^{m \times m}$ 
of the kernel matrix $\mat{K} \in \R^{n \times n}$, where $m << n$
(see for example \cite{Rasmussen06}).
They all have in common that they assume 
a given subset of the training samples 
(\textit{support vectors}) to be
is representative for the whole set.

An algorithm to select support vectors is
presented in section \ref{sec:kernel_sv_selection}.

\paragraph{Subset of Data \cite{Rasmussen06}}
The simplest approach to kernel
matrix approximation is called 
\textit{subset of data} (SD).
In this approach only the subset of $m$
support vectors is used and 
the approximation 
therefore has size $m \times m$. 
Because, only the support vectors influence 
$\tilde{\mat{K}}$ all information of 
the remaining $n-m$ samples is lost.

If the support vectors are chosen randomly
out of the training set,
subset of data is also known as  
the \textit{Nystr\"om method} \cite{Williams01}.

\paragraph{Projected Process \cite{Rasmussen06}}
A better suited approach is called \textit{projected process} (PP). 
Here,  the non support vector rows are removed but 
all columns are kept and the resulting matrix 
$\hat{\mat{K}}$ has size  $\R^{m \times n}$. 
The matrix $\hat{\mat{K}} \hat{\mat{K}}^\top \in \R^{m \times m}$ 
is used to approximate $\mat{K}^2$.  
While preserving much more information, 
the method can only be applied to algorithms which use $\mat{K}^2$.

This restriction can be circumnavigated, 
if an eigenvalue decomposition 
of $\mat K$ has to be performed anyway.
Because $\mat K^2 = \mat V \mat \Lambda^2 \mat V^\top$
for $\mat K = \mat V \mat \Lambda \mat V^\top$,
it is sufficient to perform the eigenvalue
decomposition on $\hat{\mat K} \hat{\mat K}^\top$.
Taking the square root of the eigenvalues $\hat{\mat \Lambda}$,
together with the unchanged eigenvectors $\hat{\mat V}$,
yields the projected process approximation of 
$\mat K \approx \hat{\mat V} \hat{\mat \Lambda}^{1/2} \hat{\mat V}^\top$.

\subsection{Support vector selection}
\label{sec:kernel_sv_selection}
Whatever approximation method one chooses,
the choice of support vectors is crucial
to preserve as much information as possible.
Selecting an optimal set of support vectors 
means to minimize a specific cost function
with respect to the set of support vectors,
which is a hard combinatorial problem \cite{Boyd04}.

A number of heuristics have been proposed
to find a suitable set.
Beside the purely random \textit{Nystr\"om method} \cite{Williams01},
the \textit{sequential sparse Bayesian learning algorithm} \cite{Tipping03}
(related to the \textit{relevance vector machine} \cite{Tipping01}) 
estimates the contribution of training samples
to the optimization problem.
The procedure works iteratively, but not greedy.
For big training sets, the convergence is very slow.
However, the method is specified for a Bayesian
setting (like Gaussian processes \cite{Bishop06}, for example)
and therefore not suitable for our purposes.

\newpage
\begin{algorithm}[h]
\caption{Greedy support vector selection algorithm}
\label{alg_select_sv}
\begin{algorithmic}
\REQUIRE $\nu, \ve x^{(1)}, \ldots, \ve x^{(n)}$ 

\STATE $SV_1$ = $\{x^{(1)}\}$
\STATE $\mat K_1^{-1}$ = $1/k(\ve x^{(1)},\ve x^{(1)})$
\FOR{$i$ = ${2, \ldots, n}$}
  \STATE $\ve a_i$ = $\mat K_{i-1}^{-1} \ve k(\ve x^{(i)})$ 
  \STATE $\epsilon_i$ = $k(\ve x^{(i)},\ve x^{(i)}) 
                        -\ve k(\ve x^{(i)})^\top \ve a_i$ 
  \IF{$\epsilon_i < \nu$}
    \STATE $SV_i$ = $SV_{i-1} \cup \{ \ve x^{(i)} \}$
    \STATE $\mat K_i^{-1}$ = $\frac{1}{\epsilon_i} 
          \left[ \begin{matrix} \epsilon_i K_{i-1}^{-1} + \ve a_i \ve a_i^\top 
              & -\ve a_i \\ -\ve a_i^\top & 1 \end{matrix} \right]$ 
  \ELSE
    \STATE $SV_i$ = $SV_{i-1}$
    \STATE $\mat K_i^{-1}$ = $\mat K_{i-1}^{-1}$
  \ENDIF
\ENDFOR
\STATE \textbf{return} $SV_n$
\end{algorithmic}
\end{algorithm}
\vspace{1.5cm}

In this thesis, we want to investigate another
heuristic, which was proposed by 
Csat\'o and Opper \cite{Csato01} 
and applied to a kernelized version of
least square regression by Engel \cite{Engel03b}.

We assume the data $\ve x$ to be mapped into a
feature space $\psi(\ve x)$, defined implicitly
by a kernel $k(\ve x,\ve x') = \psi(\ve x)^\top\psi(\ve x')$.
Theorem \ref{representer_theorem}
(representer theorem) assures that the optimal
function is a linear combination of 
scalar products with training samples.
If two expanded training samples would be linear
dependent, obmitting one would not change 
the optimal function.
Likewise, those samples which can be approximated
badly using linear combinations of the remaining set
are likely to be important.

The idea of algorithm \ref{alg_select_sv} is
that for a given set of support vectors 
$SV = \{\tilde{\ve x}^{(1)}, \ldots, \tilde{\ve x}^{(m)}\}$,
we define the \textit{approximated linear
dependence} (ALD) condition \cite{Engel03b}:
\begin{equation}
\label{eq:ald_condition}
\epsilon(\ve x) = \min\limits_{\ve a} 
  \left\Arrowvert \sum\limits_{j=1}^m 
    a_j \psi(\tilde{\ve x}) 
    - \psi(\ve x) \right\Arrowvert^2 \leq \nu
\end{equation}
where $\nu$ is an accuracy parameter determining
the level of sparsity.
The term $\epsilon(\ve x)$ describes the minimal
distance to the affine hull of set $SV$.
It serves as an importance measure,
which is independent of the problem
we want to optimize afterwards.

\newpage
When we denote the kernel matrix 
$\mat K_{ij} = k(\tilde{\ve x}^{(i)}, \tilde{\ve x}^{(j)})$
and the kernel expansion 
$ \ve k(\ve x) = [k(\tilde{\ve x}^{(1)}, \ve x), \ldots, 
k(\tilde{\ve x}^{(m)}, \ve x)]^\top$, we can express the
left side of (eq.~\ref{eq:ald_condition}) as
$\epsilon(\ve x) = \min_{\ve a} \ve a^\top \mat K \ve a
  - 2 \ve a^\top \ve k(\ve x) + k(\ve x, \ve x) := \min_{a} L(\ve x, \ve a)$.
The derivation with respect to $\ve a$ yields:
\begin{eqnarray*}
\frac{\partial L(\ve x, \ve a)}{\partial \ve a} &=& 2 \mat K \ve a - 2 \ve k(\ve x) 
\shallbe 0 \\
\Rightarrow \ve a^* &=& \mat K^{-1} \ve k(\ve x)
\end{eqnarray*}
which allows us to reformulate the ALD condition (eq.~\ref{eq:ald_condition}):
\begin{equation}
\label{eq:ald_condition2}
\epsilon(\ve x) = k(\ve x, \ve x) 
- \ve k(\ve x)^\top \mat K^{-1} \ve k(\ve x) \leq \nu
\end{equation}

All training samples $\ve x$ for which 
(eq.~\ref{eq:ald_condition2})
holds are considered ''well approximated'' and 
therefore omitted \cite{Engel03b}.

More precise, algorithm \ref{alg_select_sv} runs sequentially through
the training samples and tests if the current sample $\ve x^{(i)}$
fulfills the ALD condition for the current set of
support vectors $SV_{i-1}$.
If the condition fails, the sample becomes a 
support vector and
the inverse kernel matrix (which is guaranteed to
be invertible for $\nu > 0$)
is updated using   the \textit{Woodbury matrix identity}
\cite{Bishop06}.

The overall complexity of the algorithm 
is $O(m^2)$ in space and $O(nm^2)$ in time.

The support vector set obtained by this method
should be well-conditioned for most optimization problems.
It also has the nice property
that for every $\nu$ there is a maximum size for $SV_n$,
which is independent of $n$.

\begin{proposition}
Assume that (i) k is a Lipschitz continuous Mercer kernel on
$\Set X$ and (ii) $\Set X$ is a compact subset of $\R^d$.
Then there exists a constant $C$ that depends on $\Set X$
and on the kernel function such that for any training
sequence $\{\ve x^{(i)}\}_{i=1}^\infty$, and for any
$\nu > 0$, the number of selected support vectors N,
satisfies $N \leq C \nu^{-d}$.
\end{proposition}
\proof{See \cite{Engel03b}.}\gap

As a downside, the number of 
selected support vectors $m$ depends largely 
on input set $\Set X$ 
and kernel $k(\cdot,\cdot)$,
There is no way to estimate $m$ from the parameter $\nu$,
which is responsible for the sparsity of $SV_n$.
In practice one is interested in a support vector
set of an appropriate (large but not too large) size.
The experiments conducted in chapter \ref{chapter:empiric}
used RBF kernels and kept a fixed $\nu = 0.1$
while varying the kernel width $\sigma$.
This procedure was repeated until a set
of suitable size was found.

\chapter{Reinforcement Learning}
\label{chap:rl}
In this chapter we will give an outline how
to solve control problems (like robot
navigation) with reinforcement learning.
First, we will define the necessary terms
of \textit{Markov decision processes} 
(sec.~\ref{sec:markov}), followed by a
discussion of state space representation
(sec.~\ref{sec:state_rep}), especially
trigonometric polynomials (sec.~\ref{sec:trig_poly}).
Secondly, we will discuss linear algorithms
for \textit{value} (LSTD, sec.~\ref{sec:lstd})
and \textit{Q-value} estimation (LSQ, sec.~\ref{sec:lsq}).
Finally, we introduce the principles of 
\textit{policy iteration} and present
the \textit{least squares policy iteration}
(sec.~\ref{sec:policy_iteration})
followed by a summarizing conclusion of the 
complete process (sec.~\ref{sec:rl_conclusion}).

\section{Introduction}

\label{sec:rl_introduction}
Learning behaviour for an \textit{autonomous agent} 
means learning to choose the right actions at the right time.
In machine learning, this is called a \textit{policy}: A function that 
chooses an action based on some input representing
the \textit{state} of the world.
In human terms, to define a policy we have to define two things
first: The perception and the goal of the agent.
The first defines what we see as a ''state'', the second what we
see as ''right''.

While perception of the world is an open field which is mostly
circumnavigated by defining some ''essential variables'' like
position and orientation,
behavioural experiments (\textit{classical conditioning},
for example \textit{Pavlov's dog bell})
lead to a simple approach to the second question:
Reward and punishment.
In short, an agent should try to maximize his future reward while
minimizing future punishment.
Thus, training the agent becomes a simple choice \textit{what} it shall
and shall not do, eliminating the need to show it \textit{how}.

Technically, we assume to have access to some function that
filters meaningful (discrete or continuous) states out of the sensor data 
available to the agent.
The exact nature of this function is of little importance
to this chapter, besides that it provides the \textit{full}
state of the world.
In this thesis, we want to show that \textit{Slow Feature Analysis}
(SFA, chapter \ref{chapter:sfa}) can learn such a 
function under the right conditions.

The evolution of state and reward over time is modelled as a stochastic 
\textit{Markov decision process} (MDP).
Since a critical choice leading to reward or punishment
might be necessary long before its fruits will be received,
we can not use standard regression approaches.
Instead we try to estimate future reward (punishment is simply 
negative reward) for all actions in a state and take the action
which is most promising.
The sum of the expected future reward is called \textit{value}
(sec.~\ref{sec:value_estimation}).

Unfortunately, the future (and with it the value) we are trying to 
predict for the policy, depends on the policy itself.
Since this is a ''hen or egg'' problem, \textit{policy iteration} methods
repeat the steps \textit{value estimation} and \textit{policy determination}
until they converge to the \textit{optimal policy} which maximizes the value
for all states (section \ref{sec:policy_iteration}).


\subsection{Markov processes}
\label{sec:markov}
After learning,  we wish for the agent to move towards the reward.
To do this, it has to learn which transition exist 
between a given set of states and where the reward is given.
Once this knowledge is somehow established,
it needs to find a decision function,
telling it where to choose which action.

As we will see, the central element is the estimation
of the sum of expected future rewards, called \textit{value}.
Value estimation is easiest in
\textit{Markov reward processes} (MRP), so we
will start by introducing them to the reader.
The MRP itself is unable to model control tasks,
but there exists an extension named 
\textit{markov decision processes} (MDP).
MDP models the element of choice by 
introducing a set of \textit{actions}
between which a policy can choose.
We will finish this section by defining
the \textit{optimal policy}, a function
which chooses the action that will maximize
the future reward.

\begin{definition}[Markov reward process MRP]
  $\quad$ The discounted MRP 
  $M = (S, P, R, \gamma)$ 
  consists of a set of $n$ states
  $S = \{s_1, \ldots, s_n \}$, the transition probabilities
  $P: S \times S \to [0,1]$ with 
  $\forall s \in S: \sum_{i=1}^n P(s,s_i) = 1$,
  a function that assigns a real valued reward to
  every transition $R: S \times S \to \R$,
  and a discount factor $\gamma \in (0,1]$.
\end{definition}

MRPs are used to describe
the statistical properties of the \textit{random walk} 
of a variable $x \in S$ in \textit{discrete time steps}
and the reward it collects in the meantime.
In opposition to iid drawing, 
the random walk is subject to the \textit{Markov property}
\begin{equation}
P(x_t | x_{t-1}, \ldots, x_0) = P(x_t | x_{t-1}),
\end{equation}
in other words the probability of being in state $x_t \in S$ 
at time $t$ depends
exclusively on the predecessor state $x_{t-1} \in S$.
In most situations reward depends solely on the outcome, so we
simplify in the following $R: S \to \R$ 
as only dependent on the \textit{target state} of a transition.

\begin{figure}[t]
\begin{center}
\includegraphics[scale=.25]{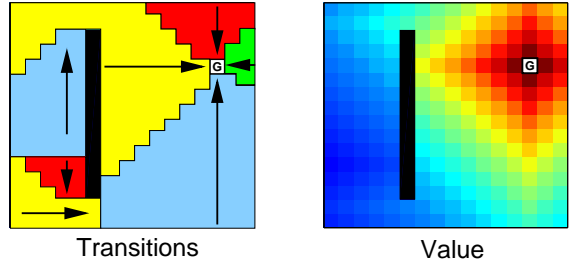}
\caption{A deterministic $16 \times 16$ states MRP (left) and the resulting 
        value for $\gamma=0.9$ (right). Arrows indicate the transition direction,
        entering \textbf{G} is rewarded.}
\label{fig:policy_and_value}
\end{center}
\end{figure}

\begin{definition}[Value \cite{Bratdke96}]
  The value $V: S \to \R$ 
  is the expected 
  sum of future rewards, discounted by the
  factor $\gamma \in (0,1]$:
  \begin{equation}
    V(x) = \E \left[{\sum\limits_{t=0}^\infty \gamma^{t} R(x_{t+1})} \mid {x_0 = x} \right]
  \end{equation}
\end{definition}
Due to the infinite sum,
one commonly uses a recursive definition which
is known as the \textit{Bellman function} \cite{Bratdke96}:
\begin{equation}
\label{eq:mrp_bellman}
V(x_t) = \E\left[R(x_{t+1}) + \gamma V(x_{t+1})\right]
= \sum\limits_{x' \in S} P(x_t, x') 
(R(x') + \gamma V(x'))
\end{equation}
Intuitively, the value function $V(x)$ tells us which 
states are promising (figure \ref{fig:policy_and_value}).
However, it does not say how to get there.
For control purposes, we would need a function 
that returns the value
of all possible actions in the current state.
Within the framework of
Markov decision processes, we can define exactly such
a function, called \textit{Q-value}.

\begin{definition}[Markov decision process MDP \cite{Lagoudakis03}]
  $\quad$    The discounted MDP
  $M = (S, A, P, R, \gamma)$
  is an extension of the MRP to control problems.
  It defines an additional set of $m$ actions 
  $A = \{a_1, \ldots, a_m\}$
  and transition probabilities 
  $P: S \times A \times S \to [0,1]$ and reward function 
  $R: S \times A \times S \to \R$ 
  are extended to depend on the executed action as well.
\end{definition}

As in the MRP case, 
we simplify the reward function as exclusively dependent
on the target state: $\forall s \in S: R(s) = R(\cdot,\cdot,s)$.

\begin{definition}[(Stationary) Policy \cite{Lagoudakis03}]
$\quad$ A function $\pi: S \times A \to [0,1]$ with 
$\forall s \in S: \sum\limits_{a \in A} \pi(s,a) = 1$
is called a stationary policy.
\end{definition}

A policy represents a control or decision automat.
$\pi(s,a)$ is the probability that the automat
chooses action $a$ in state $s$.
Since the value in a MDP depends on future decisions,
it is necessary that the policy is stationary
(i.e. does not change) throughout value evaluation.

\begin{figure}[t]
\hspace{-2.0cm}
\includegraphics[scale=.47]{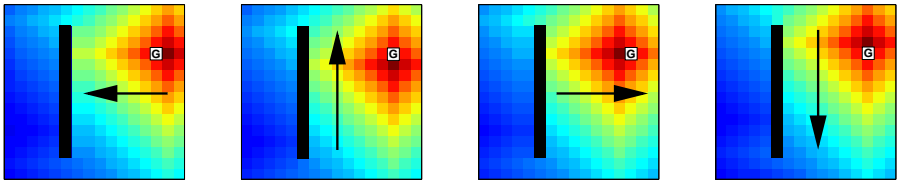}
\vspace{-1cm}
\begin{center}
\caption{The Q-values of a $16 \times 16$ MDP with 4 actions 
         and the left of figure \ref{fig:policy_and_value} 
         interpreted as a deterministic policy.
         Arrows indicate to which of the actions
         (movement directions) the Q-values belong.
         Note that in front of the goal, the Q-value of every
         action is highest.}
\label{fig:qvalues}
\end{center}
\end{figure}

\begin{definition}[Q-value \cite{Lagoudakis03}]
\label{def:qvalue}
  The Q-value $Q^\pi: S \times A \to \R$ 
  of an MDP
  is the value of an action executed in a state,
  if all following actions are selected with respect to 
  the stationary policy $\pi$:
  {\small \begin{eqnarray}
  \label{eq:mdp_qvalue}
  Q^\pi(x_t,a_t) &=& \E[R(x_{t+1}) + \gamma Q^\pi(x_{t+1},a_{t+1}) 
    \mid x_{t+1} \sim P(x_t, a_t, \cdot), a_{t+1} \sim \pi(x_{t+1}, \cdot)] \nonumber \\
  &=& \sum\limits_{x' \in S} P(x_t,a_t,x') 
    \Big( R(x') + \gamma \sum\limits_{a'\in A} \pi(x',a') \, Q^\pi(x', a') \Big)
  \end{eqnarray} }
\end{definition}
The Q-value function defines implicitly a value function for every action
(see figure \ref{fig:qvalues}).
This way one can navigate by always choosing the action with the 
highest Q-value.

Of course, we can also reformulate the 
classical value function (eq.~\ref{eq:mrp_bellman}), 
which is now dependent on policy $\pi$,
in terms of Q-values:
\begin{equation}
\label{eq:mdp_bellman}
V^\pi(x_t) = \sum\limits_{a \in A} \pi(x_t, a)\, Q^\pi(x_t, a)
\end{equation}

\begin{lemma}
\label{lem:value_equal}
For a given policy $\pi$, the value of a
MDP $M = (S, A, P, R, \gamma)$ 
is equal to the the value of the
MRP $M' = (S, P', R, \gamma)$ with 
$\forall s,s' \in S: P(s,s') = \sum_{k=1}^m \pi(s, a_k) P(s,a_k,s')$.
\end{lemma}
\proof{
Insert (eq.~\ref{eq:mdp_qvalue}) in (eq.~\ref{eq:mdp_bellman}).
Together with the definition of $P'(s,s')$ one can derive
(eq.~\ref{eq:mrp_bellman}). $\Box$ }

\begin{definition} [Optimal policy] 
  The optimal stationary policy $\pi^*$ maximizes the value $V^\pi(s)$
  for every state $s \in S$:
  \begin{equation}
    \forall \pi, \forall s \in S: V^{\pi^*}(s) \geq V^\pi(s) 
  \end{equation}
\end{definition}
This is obviously the control function we were looking for.
A control algorithm simply has to draw the actions
according to $\pi^*$ and is guaranteed to find the best
way to the reward.
Lemma \ref{lemma_opt_action} shows how to define such a policy 
with the help of Q-values.

\begin{lemma}
\label{lemma_opt_action}
(i) There can exist several $\pi^*$ but (ii) at least
one is equivalent to a deterministic selection function 
$\alpha^*: S \to A$ with $\forall s \in S: \pi^*(s, \alpha^*(s)) = 1$:
\begin{equation}
\label{eq:mdp_opt_action}
\alpha^*(s) = \argmax\limits_{a \in A} Q^{\pi^*}(s, a)
\end{equation}
\end{lemma}
\proof{
(i) 
Consider $a_1, a_2 \in A$ such that
$\max\limits_a Q^{\pi}(s,a) 
= Q^{\pi}(s,a_1) = Q^{\pi}(s,a_2)$.
For every $\lambda \in [0,1]$, 
the policy $\pi(s,a_1) = \lambda$, $\pi(s,a_2) = (1-\lambda)$
is optimal.\\
(ii) 
Due to
(eq.~\ref{eq:mdp_bellman}) and (eq.~\ref{eq:mdp_opt_action}), 
$\forall s \in S: V^{\pi^*}(s) = \max\limits_a Q^\pi(s,a)$.
Therefore, $\pi^*$ is optimal.
$\Box$}\gap

Of course, the Q-values them self depend on the policy
and the only way to solve this dilemma is to improve
both quantities in an alternating fashion
(see \textit{policy iteration}, 
section \ref{sec:policy_iteration}).


\subsection{State representation}
\label{sec:state_rep}
The MDP formalism provides us with a (possibly infinite) 
set of states $S$.
At this point we are interested in value estimation, 
that means we want to find a function
$v: S \to \R$ that at least approximates the value
for all states.
If we restrict ourself to linear functions,
i.e. $v(\ve x) = \ve w^\top \ve x$,
we need a \textit{linear representation}
$\phi: S \to \R^m$ to project the states into a
subset of $\R^m$.

\begin{definition}[State representation]
\label{def:representation}
Let $\Set X$ be an arbitrary set.
An injective function $\phi: S \to \Set X$ will be called a
representation of the set of states $S$.
If $\Set X \subseteq \R^m$ for $m \in \N^+$,
$\phi$ will be called a linear representation.
\end{definition}

We aim here to find a representation that works
with linear functions, i.e.
$V^\pi(s) \approx v^\pi(\phi(s)) = \ve w^{\top} \phi(s)$,
in other words a linear representation.
The injectivity of $\phi$ guarantees that no two 
states have the same representation.
Obviously the representation need additional
properties to approximate a value function well.
However, if $S$ is a finite state space,
there exists a representation
that allows the \textit{exact} approximation, i.e. 
$V^\pi(s) = v^\pi(\phi(s))$.

\begin{definition}[Tabular representation \cite{Bishop06}]
\label{def:tabular_rep}
$\quad\quad$ The linear representation 
$\phi: S \to \{0,1\}^n \subset \R^n$
is called a tabular representation of $S = \{s_1, \ldots, s_n\}$, 
if $\forall i \in \{1,\ldots,n\}: \phi_i(s_i) = 1$ 
and $\forall j \neq i: \phi_j(s_i) = 0$.
\end{definition}

\begin{lemma}
\label{lem:lin_val}
The value function of every MRP or MDP
can be exactly approximated as a linear function
of a tabular representation (def.~\ref{def:tabular_rep}).
\end{lemma}
\proof{
The linear function $v^\pi(s) = \ve w^\top \phi(s)$
with $\forall i: w_i = V^\pi(s_i)$
represents the value $V^\pi(s)$ exactly,
since $\forall i: v(s_i) = 
\sum_{j=1}^n w_j \phi_j(s_i) = w_i = V^\pi(s_i)$. $\Box$}
\gap

The size $n$ of this representation is equal to the
number of states.
Since the value estimation algorithms we will discuss later
have a complexity of $O(n^3)$, this is not feasible for
larger problems.
Especially if we have a continuous, i.e. infinite  state space, we are not
able to approximate the value function exactly anymore.

\paragraph{Continuous state space}
Due to the infinite number of states, 
we speak of the 
\textit{transition probability kernel} $P(s,a,\cdot)$
and a \textit{reward function} $R(\cdot)$ \cite{Antos08}.

Instead of giving a formal derivation of
continuous state spaces, we will restrict us
to subsets of the euclidian space.
Other spaces are imaginable,
but euclidian subsets should cover most practical
cases, from robot positions to pressure intensity.

\begin{assumption}[Compact subset \cite{Antos08}]
\label{ass:cont_state}
The continuous state space $S$ is a compact subset of the
$d$ dimensional euclidian space.
\end{assumption}
For example, the closed set of tuples that 
form the unit square $[0,1]^2$
is a compact subset of the 2-dimensional
euclidian space.

The approximation quality of a linear function
will depend entirely on the state representation $\phi$
and the value function at hand.
The value will probably be continuous on
most parts of a continuous state space, 
so it makes sense to
choose a set of basis functions
that are known to approximate continuous
functions well.
Examples are \textit{polynomials}, \textit{splines} \cite{Wahba90} and
\textit{trigonometric polynomials} \cite{Rudin87}.


\subsection{Trigonometric polynomials}
\label{sec:trig_poly}
In this thesis, \textit{trigonometric polynomials}
will play a major role, so
we shortly introduce them here and
discuss their approximative capabilities.

\begin{definition}[Trigonometric polynomial \cite{Bronstein99}]
A trigonometric polynomial $T: \R \to \R$ of degree $d$ with the
coefficients $a_0, \ldots, a_d$ and $b_1, \ldots, b_d$
is defined as:
\antigap\antigap\antigap

\begin{equation}
T(x) = a_0 + \sum\limits_{j=1}^d a_j cos(2 \pi j x)
+ \sum\limits_{j=1}^d b_j sin(2 \pi j x)
\end{equation}
\end{definition}
Maybe the most famous trigonometric polynomial
(of degree $d \to \infty$) is the \textit{Fourier series} \cite{Bronstein99}.
Trig. polynomials of degree $d$ are a linear combination of
the first $2d+1$ \textit{trigonometric basis functions} $\psi(x)$:
\begin{equation}
\label{eq:trig_basis_fun}
\forall j \in \N: \psi_j(x) = \Big\{ 
{\footnotesize \begin{array}{c}
  cos(j \pi x) \quad\quad\quad\; j \text{ even}\\
  sin((j+1) \pi x) \quad j \text{ odd} 
\end{array}}
\end{equation}
So we can express every trig. polynomial
of degree $d$ as $T(x) = \sum_{i=0}^{2d} w_i \psi_i(x)$.
However, most continuous state spaces will be multidimensional.

\newpage
\begin{definition}[Multivariate trigonometric polynomial]
A multivariate\\ trigonometric polynomial $T: \R^p \to \R$
of degrees $d_1,\ldots,d_p$ in the respective input
dimensions, with the coefficients $\ve w \in \R^m$ and 
$m = {\prod_{n=1}^p (2d_p+1)}$ is defined as:
\end{definition} \antigap\antigap\antigap
\begin{eqnarray}
T(\ve x) &=& \sum\limits_{n_1=0}^{2d_1} \ldots \sum\limits_{n_p=0}^{2d_p} 
            w_{\mu(n_1, \ldots, n_p)} \prod_{k=1}^p \psi_{n_k}(x_k)\\
\mu(n_1, \ldots, n_p) &=& \sum\limits_{i=1}^{p} 
    \left(n_i \prod\limits_{j=i+1}^{p} (2d_j+1) \right),
    \quad \text{(index function)}
\end{eqnarray}
Intuitively spoken, the multivariate trig. polynomial is
the weighted sum of all combinations of trigonometric basis functions
in all input dimensions.

\begin{definition}[Trigonometric representation]
\label{def:trig_rep}
The linear representation $\phi: S \to \R^m$ of the continuous state space
$S = [-1,1]^p \subset \R^p$ called trigonometric representation
with degrees $d_1,\ldots,d_p$ and $m = {\prod_{n=1}^p (2d_p+1)}$  is defined as:
\antigap
\begin{equation}
\forall i \in \{1,\ldots,p\}: \forall n_i \in \{1,\ldots,d_i\}: \quad
\phi_{\mu(n_1,\ldots,n_p)}(\ve x) = \prod\limits_{k=1}^p \psi_{n_k}(x_k)
\end{equation}
\end{definition}
Thus, we can express every
multivariate trig. polynomial as a linear function with 
a trig. representation of the same degree:
$T(\ve x) = \sum_{i=1}^m w_i \phi_i(\ve x)$.

\begin{theorem}[\cite{Rudin87}, Thm 4.25]
For every continuous function
$f: \R^p \to \R$ and every $\epsilon > 0$, there is 
a trigonometric polynomial $T(x)$, such that
$\forall x \in [-1,1]^p: |f(x) -T(x)| < \epsilon$.
\end{theorem}
\proof{See \cite{Rudin87}.}\gap

This theorem states that we can approximate \textit{any}
continuous function point wise
arbitrary well as a multivariate trigonometric polynomial.

However, since we are initially unaware of the
the value function, we have to pick 
the degrees \textit{before} the estimation.
This leaves us with the question how
big the estimation error will be for 
given degrees $d_1,\ldots,d_p$.
One can expect that it
is related to the \textit{smoothness}
of the value function at hand,
which strictly does not even have to be continuous
(e.g. at walls).
See Lorentz \cite{Lorentz66} for details.

\newpage

\section{Value Estimation}
\label{sec:value_estimation}
In this section we want to present algorithms to
estimate value and Q-value functions.
Instead of giving a complete overview of the topic, 
we want to restrict us to
\textit{model free} approaches, because for
control purposes we are only interested
in the Q-value, anyway.

We understand under \textit{model free value estimation}
the absence of an explicit model of the MDP, i.e.
$P(\cdot,\cdot,\cdot)$ and $\pi(\cdot,\cdot)$. 
Also, the state space $S$ and reward
function $R(\cdot)$ does not need to be known.
However, we explicitely need to know action space
$A$ and discount factor $\gamma$.
Of course, there also exist \textit{explicit model} approaches
for value estimation. For example in \cite{Parr08} 
independent models of the transition matrix and 
the reward function were learned.
It is also shown that for linear models
their approach yields \textit{exactly} the 
same solution as a model free approach.

Because of these restrictions, we can only
minimize the \textit{Bellman error} \cite{Parr08}
\begin{equation}
BE(\hat V) 
= \E_t[ R_t + \gamma \hat V(x'_{t}) - \hat V(x_t)]
\end{equation}
over a set of transitions $x_t \to x'_t$ with
reward $R_t$.

In this thesis, we only consider the case of linear
function classes as estimation model.
The standard approach is called 
\textit{temporal difference learning}.
Classic algorithms like \textit{TD($\lambda$)}
and \textit{Monte Carlo} \cite{Bertsekas96} 
are left out for the benefit of
\textit{least squares temporal difference} algorithms.
These circumnavigate a number of problems of
classical algorithms we can not go into
detail here.


\begin{algorithm}[b]
\caption{LSTD}
\label{alg_lstd}
\begin{algorithmic}
\REQUIRE $\gamma; \; \{(\phi(x_i), R_i, \phi(x'_i))\}_{i=1}^n 
                      \subset (\R^m \times \R \times \R^m)$
\STATE $\mat A_0$ = zeros($m$,$m$)
\STATE $\ve b_0$ = zeros($m$,$1$)
\FOR{$i=1,\ldots,n$}
  \STATE $\mat A_i$ = $\mat A_{i-1} + \phi(x_i) (\phi(x_i) - \gamma \phi(x'_i) )^\top $
  \STATE $\ve b_i$ = $\ve b_{i-1} + \phi(x_i) R_i$
\ENDFOR
\STATE $\ve w$ = pinv$(\mat A_n) \ve b_n$ 
\STATE \textbf{return} $\ve w$
\end{algorithmic}
\end{algorithm}

\subsection{LSTD}
\label{sec:lstd}
The \textit{least squares temporal difference} 
(LSTD) algorithm estimates value functions of MRP
and was first developed by Bradtke and Barto
\cite{Bratdke96}. Later, in line with the extension
of \textit{TD} to \textit{TD($\lambda$)}, 
the algorithm was extended to LSTD($\lambda$) 
by Boyan \cite{Boyan99}.
We will discuss the original LSTD algorithm, but in
a formalism that resembles Boyans extension.

LSTD is defined on MRP,
so we assume a training set
$\Set T = \{(\phi(x_t), r_t, \phi(x'_t))\}_{i=1}^n$
$\subset (S \times \R \times S)$
of $n$ transitions $x_t \to x'_t$ with reward $r_t$.
The sampling of $x_t$ affects the solution
and should be as uniform as possible.
The states can be represented by an arbitrary 
injective function $\phi: S \to \R^m$.


For readability, we define the matrices
$\mat \Phi_{ij} = \phi_i(x_j)$ and $\mat \Phi'_{ij} = \phi_i(x'_j)$.

\paragraph{Optimization problem}
LSTD minimizes a cost function related to the
Bellman error \cite{Bratdke96}:
\begin{eqnarray}
C^{lstd}\left(\ve w, (\mat \Phi, \mat \Phi', \ve R)\right) 
&=& \left\Arrowvert \E_t \left[\phi(x_t) \left(\hat V^\pi(x_t) - 
    (R_t + \gamma \hat V^\pi(x'_{t})) \right) \right] \right\Arrowvert_\text{F}^2 \nonumber \\
&=& \frac{1}{n} \left\Arrowvert 
      \mat\Phi (\mat \Phi - \gamma \mat \Phi')^\top \ve w 
      - \mat \Phi \ve R 
    \right\Arrowvert_\text{F}^2 \label{eq:lstd_cost}
\end{eqnarray}
with the linear function $\hat V^\pi(x) = \ve w^\top \phi(x)$
and $\E_t[\cdot]$ being the empirical mean.

Deriving (eq.~\ref{eq:lstd_cost}) with respect to $\ve w$
yields the minimum $\ve w^*$:
\begin{eqnarray*}
\frac{\partial C^{lstd}\left(\ve w, 
  (\mat \Phi, \mat \Phi', \ve R)\right)}{\partial \ve w}
&=& \frac{2}{n} \mat{\dot \Phi}_\gamma \mat \Phi^\top 
      \mat \Phi \mat{\dot \Phi}_\gamma^\top \ve w
  - \frac{2}{n} \mat{\dot \Phi}_\gamma \mat \Phi^\top 
      \mat \Phi \ve R \shallbe 0 \\
\Rightarrow \ve w^* &=& (\mat{\dot \Phi}_\gamma \mat \Phi^\top 
      \mat \Phi \mat{\dot \Phi}_\gamma^\top )^{-1}
      \mat{\dot \Phi}_\gamma \mat \Phi^\top 
      \mat \Phi \ve R \\
&=& (\mat\Phi (\mat \Phi - \gamma \mat \Phi')^\top)^{\dagger} \mat \Phi \ve R
\end{eqnarray*}
where $\mat{\dot \Phi}_\gamma = \mat \Phi - \gamma \mat \Phi'$
and $\dagger$ denotes the pseudo inverse.
The last line holds if the rows of 
$\mat \Phi \mat{\dot \Phi}_\gamma^\top$ are 
linearly independent \cite{Boyan99}.

LSTD (algorithm \ref{alg_lstd}) has a complexity 
of $O(m^2)$ in space and $O(m^3)$
in time with $m$ being the length of the state
representation.


\subsection{LSQ}
\label{sec:lsq}

\begin{algorithm}[b]
\caption{LSQ}
\label{alg_lsq}
\begin{algorithmic}
\REQUIRE $\phi: S \times A \to \R^m; \; \gamma; \{ x_i, a_i, R_i, x'_i, a'_i \}_{i=1}^n 
          \subset (S \times A \times \R \times S \times A)$
\FOR{$i=1,\ldots,n$}
  \STATE $\phi_i = \phi(x_i,a_i)$
  \STATE $\phi'_i = \phi(x'_i, a'_i)$
\ENDFOR
\STATE $\ve w$ =  LSTD($\gamma; \{(\phi_i, R_i, \phi'_i)\}_{i=1}^n$)
\STATE \textbf{return} $\ve w$
\end{algorithmic}
\end{algorithm}

The \textit{least squares Q-value} (LSQ) algorithm
was introduced by Lagoudakis et al. \cite{Lagoudakis01}
to extend LSTD to Q-values.
The basic idea is an extension of the state representation
$\phi: S \to \R^m$ to state action representation 
$\phi: S \times A \to \R^m$.

\begin{definition}[State action representation \cite{Lagoudakis01}]
\label{def:sa_pairs}
An arbitrary linear representation $\phi: S \to \R^m$
(e.g. of a continuous state space) can be extended
by a discrete set of actions $A=\{a_1,\ldots,a_k\}$ 
to a combined representation $\phi: S \times A \to \R^{mk}$
with $\forall s \in S, \forall a_i \in A: \phi(s,a_i) 
= [\ve z_a^\top, \phi(\ve s)^\top, \ve z_b^\top]^\top$,
$\ve z_a \in \{0\}^{(i-1)n}$ and $\ve z_b \in \{0\}^{(m-i)n}$.
\end{definition}

Formally, we assume a training set
$\Set T = \{(x_i,a_i,R_i,x'_i,a'_i)\}_{i=1}^n$
of $n$ observed transitions
$x_i \statetrans{a_i} x'_i$,
the collected reward $R_i$
and the next action $a'_i$ chosen according to $\pi$.
The states might be available in an additional representation, 
but they serve only as input for $\phi(\cdot, \cdot)$.

\begin{proposition}
\label{pro:action_state_pair}
Given a stationary policy $\pi$ and a
state action representation $\phi: S \times A \to S'$, 
the Q-value  $Q^\pi(s,a)$ of the MDP
$M = (S,A,P,R,\gamma)$
is equal to the value $V'(\phi(s,a))$ of the MRP
$M' = (S', P', R', \gamma)$, where \\
$\forall s,s' \in S, \forall a,a' \in A:$
\antigap
\begin{eqnarray*}
&(i)&
S' = \setunit\limits_{s \in S} \setunit\limits_{a \in A} \{ \phi(s,a) \} \\
&(ii)&
P'\big(\phi(s,a), \phi(s',a')\big) = P(s,a,s') \pi(s',a') \\
&(iii)&
R(\phi(s,a),\phi(s',a')) = R(s,a,s')
\end{eqnarray*}
\end{proposition}
\proof{ $\forall s \in S, \forall a \in A$:
\antigap
{\small \begin{eqnarray*}
Q^\pi(s,a) &=& \sum\limits_{s'\in S} P(s,a,s') \Big( R(s,a,s') 
  + \gamma \sum\limits_{a' \in A} \pi(s',a') Q^\pi(s',a') \Big)\\
&=& \sum\limits_{s'\in S} \sum\limits_{a'\in A} \Big( 
      \pi(s',a') P(s,a,s') R(s,a,s')
  + \gamma P(s,a,s') \pi(s',a') Q^\pi(s',a') \Big) \\
&=& \sum\limits_{s'\in S} \sum\limits_{a' \in A} P'(\phi(s,a),\phi(s',a')) \Big(
      R(\phi(s,a),\phi(s',a'))
  + \gamma V'(\phi(s',a')) \Big) \\
&=& V'(\phi(s,a))
\end{eqnarray*} } 
\closetoequation
The last equality holds only if 
$\phi(\cdot,\cdot)$ is injective. $\Box$
}\gap

LSQ combines the state and action
using a injective function $\phi(\cdot,\cdot)$ 
(e.g. definition \ref{def:sa_pairs}),
and then utilizes LSTD to estimate $V'(\phi(\cdot,\cdot))$.
Since no explicit model of the MDP is needed,
estimating $V'(\phi(\cdot,\cdot))$ is an
easy way to estimate $Q^\pi(\cdot,\cdot)$.
The policy affects the solution through $a'_i \sim \pi(x'_i, \cdot)$,
which satisfies definition \ref{def:qvalue} (Q-values).

\section{Policy Iteration}
\label{sec:policy_iteration}
With lemma \ref{lemma_opt_action} we have found
an intuitive way to realize the optimal policy
by means of Q-values.
These can be estimated model free by the LSQ
algorithm (sec.~\ref{sec:lsq}).
However, the training set is sampled using a 
(not necessary known) policy and 
LSQ needs a policy to choose $a'$.

A solution to this problem is called
\textit{policy iteration} and consists of
a sequence of monotonically improving policies
$\pi_0, \ldots, \pi_m$ of the form \cite{Bertsekas96}:
\begin{eqnarray}
\alpha_i(s) & =& \argmax\limits_a Q^{\pi_i}(s,a) \nonumber \\
\forall s \in S&:&  \pi_{i+1}(s,\alpha_i(s)) = 1 \label{eq:pol_iter}
\end{eqnarray}
The evaluation of $\pi_m$ obviously requires an iterative
alternation of \textit{Q-value estimation} of policy
$\pi_i$ and \textit{policy improvement} (eq.~\ref{eq:pol_iter})
to achieve a so called \textit{greedy policy} $\pi_{i+1}$.
For the \textit{initial policy} $\pi_0$,
one often chooses a random policy with equal probabilities for
all actions.

In the case of tabular state representation it is possible
to prove the monotonic improving property of 
policy iteration.
Approximations of $Q^\pi$, however, complicate
convergence proofs and depend on
approximation quality and Q-value
estimation algorithm \cite{Lagoudakis03}.

\subsection{Sampling}
For estimating Q-values, we need a training set
$\Set T = \{(x_i, a_i, r_i, x'_i, a'_i)\}_{i=1}^n
\subset S \times A \times \R \times S \times A$.
All $x'_i \sim P(x_i,a_i,\cdot)$ and $r_i \sim R(x_i,a_i,\cdot)$ 
follow the MDP and all $a'_i \sim \pi_j(x'_i,\cdot)$
are drawn according to the current policy.
The start states $x_i$ and actions $a_i$, however,
have to be sampled by another process.

Most experiments are performed in \textit{trajectories}.
For every trajectory of length $l \in \N^+$ we draw a start state
according to some distribution $x'_0 \sim \hat s(\cdot)$
and follow a stationary policy $\hat \pi$,
i.e. $\forall i \in \{1,\ldots,l\}: 
x_i = x'_{i-1}, a_i \sim \hat \pi(x_i,\cdot)$.

However, this approach can lead to some
obstacles that can ruin the convergence of policy iteration.
We will demonstrate the problem on a simple version 
of the \textit{Sarsa} algorithm \cite{Sutton98}
and the solution with the \textit{LSPI} algorithm \cite{Lagoudakis01}.

\paragraph{Sarsa} 
The easiest approach is to record a number of trajectories
with a uniform start distribution $\hat s$
and to follow the current policy, i.e. 
$\hat \pi = \pi_i$. 

Since equation \ref{eq:pol_iter} is based on the
maximum of $Q^\pi(s,\cdot)$ in a state $s \in S$,
it is necessary that $Q^\pi(s,a)$ is approximated
well for all $a\in A$.
This is not the case here, 
since every policy but the first is greedy.
Sarsa will therefore choose the same action every time it 
visits the same state\footnote{
To be precise, in \cite{Sutton98} Sarsa chooses
actions $\epsilon$-greedy, i.e. with a (iteration dependent) 
probability of $\epsilon$ greedy and with 
$(1-\epsilon)$ random.}.
This will lead to a poor approximation of all but
the currently preferred action and therefore to
poor policy convergence.

\begin{algorithm}[t]
\caption{LSPI}
\label{alg_lspi}
\begin{algorithmic}
\REQUIRE $\phi: S \times A \to \R^m; \; \gamma; \; \epsilon$
\STATE Generate $\{x_i, a_i\}_{i=1}^n \subset S \times A$ uniformly distributed
\STATE Measure all $x'_i$ and $r_i$ in experiments based on $x_i$ and $a_i$
\STATE Generate all $a'_i$ uniformly distributed for initial random policy
\STATE // Policy iteration
\STATE $\ve w = \ve \infty$; $\ve w' = \ve 0$
\WHILE{$||\ve w - \ve w'|| \geq \epsilon$}
  \STATE$\ve w'$ = $\ve w$
  \STATE $\ve w$ =  LSQ($\phi; \; \gamma; \{x_i,a_i,r_i,x'_i,a'_i\}_{i=1}^n$)
  \FOR {$i \in \{1, \ldots, n\}$}
    \STATE $a'_i$ = $\argmax\limits_{a}$($\ve w^\top \phi(x'_i,a)$)
  \ENDFOR
\ENDWHILE
\STATE \textbf{return} $\ve w$
\end{algorithmic}
\end{algorithm}

\paragraph{LSPI}
It is therefore important to draw $a_i$ as well as 
$s_i$ randomly, uniform distributed at best \cite{Lagoudakis03}.
One way to achieve this is to record an
initial \textit{dictionary} with a random
policy $\hat \pi$ and change only $a'$ between
the iterations.
This approach is called \textit{least squares policy iteration}
and ensures equally well approximated Q-values.

On the one hand,
this approach eliminates 
the need for resampling every iteration,
which reduces the experimental overhead.
On the other hand,
there is no way to compensate unbalanced sampling
in the dictionary.
A random walk constrained by a complex
environment can be very erratic,
especially with few but long trajectories.
Optimally, one would record trajectories
of length $1$ with $\hat s$ and $\hat \pi$ 
being uniform distributions over $S$ and $A$, respectively.

Algorithm \ref{alg_lspi} shows the complete LSPI method,
for more information see \cite{Lagoudakis01} and
\cite{Lagoudakis03}.


\begin{algorithm}[t]
\caption{Control}
\label{alg_control}
\begin{algorithmic}
\REQUIRE $\phi: S \times A \to \R^m; \; \ve w^*$
\WHILE{true}
  \STATE $x$ = observe()
  \STATE $a$ = $\argmax\limits_{a}$($\ve w^{*\top} \phi(x,a)$)
  \STATE execute($a$)
\ENDWHILE
\end{algorithmic}
\end{algorithm}

\section{Conclusion}
\label{sec:rl_conclusion}
In this chapter we presented an outline how to
solve control problems with reinforcement learning.

\begin{itemize}
\item
Algorithm \ref{alg_control} shows how the control
works after an optimal policy, represented by
the parameter vector $\ve w^*$, is obtained
by LSPI.
\item
For LSPI (alg.~\ref{alg_lspi}), it is sufficient to record one
random walk to train the complete control procedure.
To fulfill LSPIs requirement of uniformly distributed
start states and actions, one can rely on the
original random walk properties (given enough examples)
or sample a balanced set out of the video to create
a dictionary.
\item
Using LSQ (alg.~\ref{alg_lsq})
and LSTD (alg.~\ref{alg_lstd}) 
we can estimate the Q-value without
knowledge of the underlying MDP.
Only some samples in the dictionary have to
be labeled as \textit{reward} or \textit{punishment}.
\item
The states can be given in any representation, if
we use definition \ref{def:sa_pairs} to generate
state action representations.
However, a representation suited for linear
functions is recommendable, 
e.g. a trigonometric representation (def.~\ref{def:trig_rep}).
\end{itemize}

\chapter{Slow Feature Analysis}
\label{chapter:sfa}
In this chapter we want to introduce
an unsupervised learning criteria
called \textit{slow feature analysis} (SFA,
sec. \ref{sec:sfa_problem}).
SFA is used to find continuous states in
temporal data.
As a big advantage,
theoretical tools are available that 
predict the behaviour in a variety of situations 
(sec. \ref{sec:sfa_opt_responses}).
We give an overview of recent applications in section
\ref{sec:sfa_applications} and conclude with
the commonly used \textit{linear SFA} algorithm
(sec. \ref{sec:SFA_linear}).
Since linear SFA is not suited for the
task in this thesis, we also develop a kernelized version
of the linear SFA algorithm (sec. \ref{sec:SFA_kernel}).

\section{Introduction}
\label{sec:sfa_introduction}

Biological systems start with little hard-coded 
instructions (e.g. genes).
They grow, however, far more complex over time.
It is estimated that the human genome consists of
approximately 25.000 genes;
the human brain alone is composed of $10^{11}$ neurons, 
each with its own individual properties.
Therefore, biological systems must have a way of
self-organizing and diversification, depending
on the environment.

To understand and reproduce this
learning-without-a-teacher,
the field of \textit{unsupervised learning}
was developed.
In contrast to \textit{supervised learning}
(e.g. regression, section \ref{sec:regression})
no \textit{target value} is available,
thus the design of the cost function
alone determines the optimal function.
Since there is no explicit target,
every optimization problem has to rely on
a \textit{self-organizing principle}.

\paragraph{Temporal coherence}
The underlying principle of
\textit{slow feature analysis} are
temporal coherent signals.

Imagine a sensor of undetermined type.
All we can see are the multidimensional
sensor readings, which probably are distorted
by some kind of uncorrelated noise.
Whether the sensor itself is moved
or the environment changes around it,
the readings will vary over time.
We are searching for a \textit{filter}
which applied on the sensor readings
yields a meaningful representation
of the current \textit{state of environment}.

As shortly addressed in the last chapter
(sec. \ref{sec:rl_introduction}),
there is no mutual consent \textit{what}
exactly a state of the environment \textit{is}.
The approach taken here is to concentrate
on the things \textit{changing} over time.
For example, 
if an object is moved in front of a static
background, than the background
(as complex as it might be) is completely
irrelevant for the state.
The object, however,
is equally unimportant.
Only its \textit{position} varies
over time, and is therefore the state
of the environment.

Keeping in mind that noise is considered
uncorrelated to the state 
and therefore not \textit{markovian}, 
a filter specializing on noise will change quickly.
On the other hand, most important things 
in the real world change \textit{slowly} 
and \textit{continuously} over time.
Therefore we should look for filters that
produce a slowly varying output,
while making sure \textit{that}
they vary at all and not just represent background.
Because the state of something as
complex as a real environment must be
multidimensional, we should
expect a set of filters,
which have to be independent of
each other.
This means at least they have to be decorrelated.

State spaces extracted with such a
filter will be \textit{temporal coherent},
but ignore fast changing processes,
e.g. the ball in table tennis.

\vspace{.5cm}
\subsection{Optimization problem}
\label{sec:sfa_problem}
There are several approaches to learn a filter
as described above.
They share some basic claims on the filters output
which were formulated best by Wiskott et al.
\cite{Wiskott02, Wiskott03}.
\begin{itemize}
\item
The mean \textit{temporal derivation} of the output
should be as small as possible.
Normally the derivation is not known,
so the \textit{discrete temporal derivation},
i.e. the difference between successive outputs,
is used instead.
To punish large changes more than small
ones, all approaches square the
derivation pointwise.
This value is called \textit{slowness}.

\item
To ensure that the output varies \textit{at all},
the \textit{variance} of the output is normalized, either
by constraint or by multiplying the cost function 
with the inverse variance.
This claim is equivalent to
a \textit{zero mean} and \textit{unit variance}
constraint.

\item
The different filters should be independent
of each other.
Since statistical independence is hard to achieve,
all sighted works restrict themself to
claim \textit{decorrelation} between the filters.
\end{itemize}

\newpage
\begin{definition}[SFA Problem \cite{Wiskott02}]$ $\\
\label{def:sfa_problem}
Given a temporally ordered sequence of observations 
$ \ve{x}^{(1)}, \ldots, \ve{x}^{(n)} \in \Set X$,
we want to find a set of $k$ mappings 
$\phi_i(\ve{x}): \Set X \rightarrow \R$ 
that satisfy:
\begin{eqnarray*}
\text{min} \quad s(\phi_i)
  &:= & \E[\dot{\phi}_i(\ve x)^2] \quad \text{(slowness)}\nonumber\\
\text{s.t. }  \E[\phi_i(\ve x)]
  & = & 0 	\text{\quad\quad\quad\quad\; (zero mean)} \nonumber \\
\E[\phi_i(\ve x)^2]
  & = & 1 	\text{\quad\quad\quad\quad\; (unit variance)} \\
\forall j \neq i: \E[\phi_i(\ve x) \phi_j(\ve x)]
  & = & 0 	\text{\quad\quad\quad\quad\; (decorrelation)} \nonumber 
\end{eqnarray*}
\end{definition}
where $\dot\phi$ denotes the temporal derivative and 
$s(\phi_i)$ the \textit{slowness} of mapping $\phi_i$.
The indices $i$ are sorted in ascending order according to their slowness,
so the first mapping is the slowest.

Grouping the outputs and its discrete derivatives
into matrices $\mat \Phi_{it} = \phi_i(\ve x^{(t)})$
and $\mat{\dot \Phi}_{it} = 
\phi_i(\ve x^{(t+1)}) - \phi_i(\ve x^{(t)})$
we can express problem \ref{def:sfa_problem} more compact:
\begin{eqnarray}
\text{min}\quad  s(\mat \Phi) &=& \trace(\mat{\dot \Phi}^\top \mat{\dot \Phi}) \nonumber \\
\text{s.t.}\quad\; \mat \Phi \ve 1 &=& \ve 0 \\
{\scriptstyle \frac{1}{n}} \mat \Phi \mat \Phi^\top &=& \mat I \nonumber
\end{eqnarray}

\paragraph{Other approaches}
Beside Wiskott et al. \cite{Wiskott02}, we
found two other optimization problems
that follow the principle of temporal coherence:
\begin{itemize}
\item
Bray and Martinez \cite{Bray02} formulated 
as a related cost function the ratio of long-
and short-term variance $V$ and $S$:
\begin{equation}
\text{max} \; F(\mat\Phi) = 
  \frac{V}{S} = 
  \frac{\E[\bar \phi(\ve x)^2]}
  {\E[\tilde \phi(\ve x)^2]}
\end{equation}
where $\bar \phi(\ve x)$ and $\tilde \phi(\ve x)$
are the centered long- and short-term means, e.g.
$\bar \phi_i(\ve x^{(t)}) = {\scriptstyle \frac{1}{2m}}
\sum_{j=t-m}^{t+m-1} \phi_i^{(j)}$
and an equivalently defined $\tilde\phi$
with smaller $m$.

Minimizing the short term change (derivation) while
maximizing the long term change (variance) 
follows the same principle as SFA.
Decorrelation, however, is not required by this
optimization problem and only appears in the
results of \cite{Bray02}, because the
problem is solved with an eigenvalue decomposition,
for which the solution is decorrelated by default.

\item
Einh\"auser et al. \cite{Einhauser05} used
a cost function which is based on the same constraints
but was formulated differently\footnote{
The presented formalism is no 
direct citation but a readable version.}:
\begin{eqnarray}
\label{eq:sfa_problem_alt}
\text{max} \; F(\mat \Phi) = 
  - \sum\limits_{i=1}^k \frac{\E[ \dot \phi_i(\ve x)^2]}
            {\V[\phi_i(\ve x)]}
  - \frac{1}{(k-1)^2} \sum\limits_{i=1}^k
      \sum\limits_{{j=1 \atop j \neq i}}^k
      \frac{\C[\phi_i(\ve x), \phi_j(\ve x)]^2}
      {\V[\phi_i(\ve x)] \, \V[\phi_j(\ve x)]}
\end{eqnarray}
The first term on the right side incorporates the
minimization of the slowness with the unit variance
constraint, the second term represents
the decorrelation constraint.

Wyss et al. \cite{Wyss06} used a weighted version
of this cost function with an additional zero
mean constraint embedded the same way.
\end{itemize}

\subsection{Optimal responses}
\label{sec:sfa_opt_responses}
Consider we could solve problem
\ref{def:sfa_problem} in the class
of continuous functions.
Since the form of the optimal function 
depends on the training sequence at hand,
we are interested in the form of the 
\textit{output}, which should always be the same.

\begin{definition}[Optimal responses]
Let 
$\Set O$ be an unsupervised optimization problem
based on samples from set $\Set X$.
Let further $f$ be the optimal function of $\Set O$ 
based on any infinite training set $\Set T \subset \Set X$.
Then $\{f(x) | x \in \Set T\}$ is called the
optimal responses of $\Set O$.
\end{definition}
In our case, we look for the function
that describes the slowest output
that still fulfills the constraints
of problem \ref{def:sfa_problem}.
One can use \textit{variational calculus}
to find a solution independent of
a specific input.
\cite{Wiskott03}.

Formally we assume some stochastic
process that generates the training set.
The process consists of a state in a \textit{bounded}
and \textit{continuous} state space $\Set S$ and
some generating parameters, e.g. a position $s \in [0,1] \subset \R$
plus a Gaussian random variable: 
$s^{(t+1)} = s^{(t)} + \normdist(0,\sigma)$.
$s$ would be the state and
$\sigma$ the \textit{velocity} of change.
A function $\nu: \Set S \to \Set X$ generates
observations $x \in \Set X$ out of the true
states.

There have to be some rules
that constraint the stochastic process to $\Set S$.
The optimal responses will depend on what happens
if a boundary is reached, i.e. if the state
is stopped (reflected) or continues at the opposite boundary.
These rules are incorporated into the
variational calculus approach by
\textit{boundary conditions}.

\begin{proposition}
\label{pro:orth_der}
A set of functions which applied on a training sequence
fulfills zero mean, unit variance, decorrelation
and has orthogonal time derivatives,
minimizes problem \ref{def:sfa_problem} 
within the space spanned
by these functions.
\end{proposition}
\proof{See \cite{Wiskott03}.}\gap

This proposition essentially gives us
the means to validate a proposition
about optimal responses.
We start with the simple case of a compact subset of
the one dimensional Euclidean space.
Without loss of generality, assume $\Set S = [0,1]$.

\newpage
\begin{proposition}[Cyclic boundary condition]
Assuming a cyclic boundary condition, i.e. $\nu(0) = \nu(1)$,
the optimal responses are $\forall x \in [0,1],\forall i \in \N^+:$
\begin{equation}
 \phi_i^*(\nu(x)) = \Bigg\{
    {\sqrt{2} \; cos\big(j \pi x\big) \quad\quad\quad\; i \text{ even}
    \atop \sqrt{2} \; sin\big((j+1) \pi x\big) \quad          i \text{ odd}}
\end{equation}
\end{proposition}\gap
\proof{Proposition \ref{pro:orth_der} holds. 
See \cite{Wiskott03} for details.}\gap

Any cyclic state, e.g. orientation or electrical phase,
is subject to this condition.
Note the similarity to trigonometric basis functions
(eq. \ref{eq:trig_basis_fun}),
i.e. with the exception of the constant $\psi_0$, all
basis functions are present.
The domain, however, is only $[0,1]$ instead of
$[-1,1]$, reflecting the cyclic boundary constraint.

\begin{proposition}[Free boundary condition]
Assuming no boundary condition, the optimal responses are
$\forall x \in [0,1], \forall i \in \N^+:$
\antigap
\begin{equation}
 \phi_i^*(\nu(\ve x)) = 
   \sqrt{2} \; cos\big(j \pi x \big)
\end{equation}
\end{proposition}\gap
\proof{Proposition \ref{pro:orth_der} holds. 
See \cite{Wiskott03} for details.}\gap

Any state \textit{between} two limits, e.g. coordinates
restricted by walls, is subject to the free boundary condition.
In fact, this is another expression of the trigonometric
polynomial on the interval $[0,1]$ instead of $[-1,1]$. 
Since the even trig. basis functions
can approximate any \textit{even} continuous function
in the intervale $[-1,1]$, they can approximate
any continuous function in $[0,1]$ \cite{Wiskott03}.

However, many realistic scenarios require
multidimensional state spaces.
For example, a robot who can move freely
on a plane and look in any direction,
requires a 3-dimensional space.
Franzius et al. \cite{Franzius07} derived
a solution for this case.
They considered a stochastic generation
process where an agent (e.g. a robot) can move in the
direction it faces and rotate to change
this direction.
This implies a 
state space $\Set S = [0,1]^3$,
where the first two directions represent two
spatial degrees of freedom and the last
the orientation, all scaled to the
interval $[0,1]$.

\begin{proposition}[3d state space]
\label{pro:opt_res_3d}
Let $\Set S = [0,1]^3$ be a state space with
free boundary conditions in the first two,
and a cyclic boundary condition in the last dimension.
Under the condition of decorrelated movements in the training set,
the optimal responses are
$\forall x,y,\theta \in [0,1]: \forall (i,j,l) \in \N^3 \setminus \{(0,0,0)\}:$
\begin{equation}
\phi_{ijl}^*(\nu(x,y,\theta)) = \Bigg\{
        {\sqrt{2}^3\; cos(i \pi x)\, cos(j \pi y) \, cos(l \pi \theta)
            \quad\quad\quad\; i \text{ even}
  \atop \sqrt{2}^3\; cos(i \pi x)\, cos(j \pi y) \, sin((l+1) \pi \theta)
             \quad i \text{ odd}}
\end{equation}
\end{proposition}\gap
\proof{Proposition \ref{pro:orth_der} holds. 
See \cite{Franzius07} for details.}\gap

The solution resembles the
trigonometric representation (def. \ref{def:trig_rep}),
but uses the basis functions of the one dimensional
boundary cases instead of $\psi$.

However, these analytical solutions are exceptions.
Experiments show that non-rectangular state spaces
of the same dimensionality
develop different optimal responses (sec. \ref{sec:exp_sfa_two}).

\begin{proposition}[Mixture responses]
\label{pro:sfa_mix}
If $\phi_i$ and $\phi_j$ are two optimal functions
of problem \ref{def:sfa_problem} with equal
slowness, then any linear combination 
$a \phi_i + b \phi_j$ has the same slowness,
as long as $a^2+b^2 = 1$.
\end{proposition}
\antigap\antigap
\begin{eqnarray*}
\text{\bf Proof:}\quad\; s(a\phi_i + b\phi_j) 
&=& \E[(a \dot\phi_i(\ve x) + b \dot\phi_j(\ve x))^2]\\
&=& a^2 \E[\dot\phi_i(\ve x)^2] 
- 2ab \E[\dot\phi_i(\ve x) \dot\phi_j(\ve x)]
+ b^2 \E[\dot\phi_j(\ve x)^2]\\
&\stackrel{(\text{Pr.\ref{pro:orth_der}})}{=}& 
  a^2 s(\phi_i) + b^2 s(\phi_j) \stackrel{s(\phi_i) 
= s(\phi_j)}{=}(a^2+b^2) s(\phi_i) \quad\Box
\end{eqnarray*}
Intuitively, if we have two sets of optimal responses 
with  equal slowness,
any rotation between them is also a set of optimal responses,
since it neither violates the constraints nor
changes the slowness.
In the case of one dimensional state spaces,
only numerical errors can result in mixture responses.
Solutions of proposition \ref{pro:opt_res_3d},
on the other hand, are prone to exhibit mixtures.

Let us consider a mean driving velocity $v$ 
(influencing the slowness of spatial dependent filters)
and a mean rotational velocity $\omega$
(influencing the slowness of orientation dependent filters).
Due to the same mean velocity in both spatial
directions, we have for any $a^2+b^2=1$:
\begin{equation}
\forall j,l \in \N, \forall i \in \N^+:
s(\phi^*_{ijl}) = s(\phi^*_{jil}) 
= s(a \phi^*_{ijl} + b \phi^*_{jil})
\end{equation}
Therefore, we have to expect mixture responses
between all filters of equal slowness.
Note that this does not change the number 
of decorrelated filters, 
only their output is no longer
determined by proposition \ref{pro:opt_res_3d}.

\section{Applications}
\label{sec:sfa_applications}
To the authors knowledge, there are no
practical applications based on slow feature analysis.
There are, however, a couple of scientific works
that investigate its potential in different contexts.


\subsection{Receptive fields}
The first brain area processing visual information
from the eye is called \textit{visual cortex}.
It contains cells that specialize on specific
patterns (e.g. bars of specific orientation)
in a small part of the perceived image, a
so called \textit{receptive field}.
The question arises how these cells differentiate
and adopt patterns that allow an
efficient coding of the perceived scene.

\begin{figure}[t]
\begin{center}
\includegraphics[scale=.6]{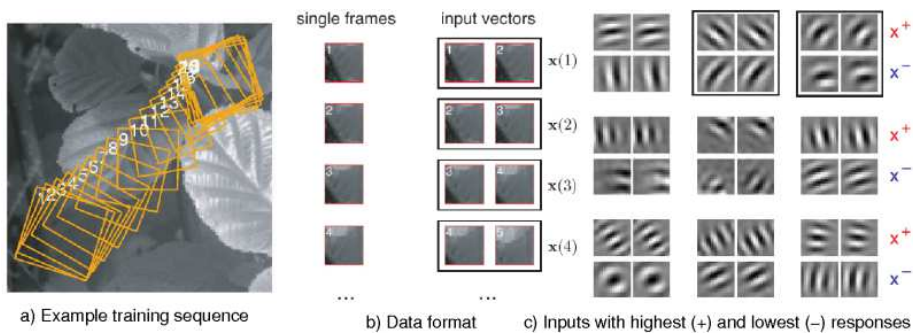}
\vspace{-.7cm}
\caption{Experimental setup and some results of Berkes and Wiskott \cite{Berkes05a}.}
\label{fig:Berkes05a}
\end{center}
\end{figure}

Berkes and Wiskott \cite{Berkes05a} used SFA
to learn filters that react on patterns
similar to those of real cells.
They constructed a random process that moved
a frame over natural pictures by
translation, rotation and/or zooming 
(fig. \ref{fig:Berkes05a}a)
Input samples consisted of two successive frames
to represent a special neuronal structure
called \textit{complex cells} (fig. \ref{fig:Berkes05a}b).
The results (fig. \ref{fig:Berkes05a}c)
resemble phase shifted \textit{gabor filters},
as they are found in these cells.

The results build a convincing argument,
but for the sake of completeness
one has to add that there are other
unsupervised methods that
produce similar filters.
For example, Lindgren and Hyv\"arinen
used quadratic ICA and also produced
comparable results, though in a smaller resolution
\cite{Lindgren07}.

\subsection{Pattern recognition}
Naturally SFA predicts continuous states.
While discrete states (e.g. classes of patterns)
present no problem in principle,
the absence of a temporal structure \textit{is}.

Berkes constructed a classificator
for handwritten digits from the
MNIST database based on SFA \cite{Berkes05b}.
Instead of a temporal derivative,
two training digits of the
same class were subtracted, 
thus minimizing the distance
between alike patterns.
For the final classification,
Gaussian distributions
were fitted to the digit classes,
each representing the probability 
of membership.

With 1.5\% error rate
on a test set of 10,000 samples, 
the specified classifier performed
comparable to established
algorithms in this field,
e.g. the LeNet-5 algorithm 
misclassified 0.95\% of the test set
in comparison to 12\% by a linear
classifier.


\begin{figure}[b]
\begin{center}
\includegraphics[scale=.42]{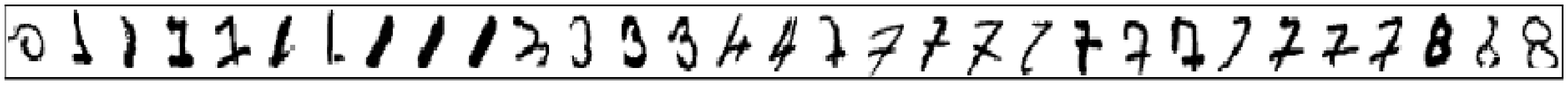}
\vspace{-.3cm}
\caption{Some handwritten digits from the MNIST database used in \cite{Berkes05b}.}
\label{fig:Berkes05b}
\end{center}
\end{figure}

\subsection{Place cells}
\label{sec:sfa_place_cells}
\begin{figure}[t]
\begin{center}
\includegraphics[scale=.8]{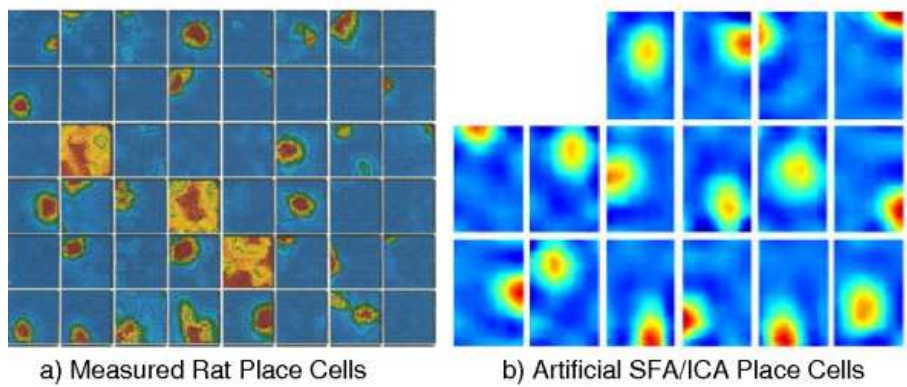}
\caption{Simultaneous recorded place cells (a) in comparison with
results from Franzius et al. \cite{Franzius07} (b).
All sub-images represent cell activity levels at different places
with a ceiling camera view on a
square/rectangular room at which the rat familiarized itself by
a random walk.
Image source: \cite{Franzius08}}
\label{fig:FranziusPC}
\end{center}
\end{figure}

Rodents are known to develop specialized \textit{place cells}
in hippocampal areas after familiarizing them self
with their surrounding.
These cells are only active in one part of the room,
independent to the heads orientation (fig. \ref{fig:FranziusPC}a).
Together with \textit{head-direction cells},
for which the output depends only on the orientation,
independent of the rodents position,
they are believed to play a mayor role in the
rodents sense of navigation \cite{Franzius07}.

\begin{figure}[t]
\hspace{-.8cm}
\includegraphics[scale=.47]{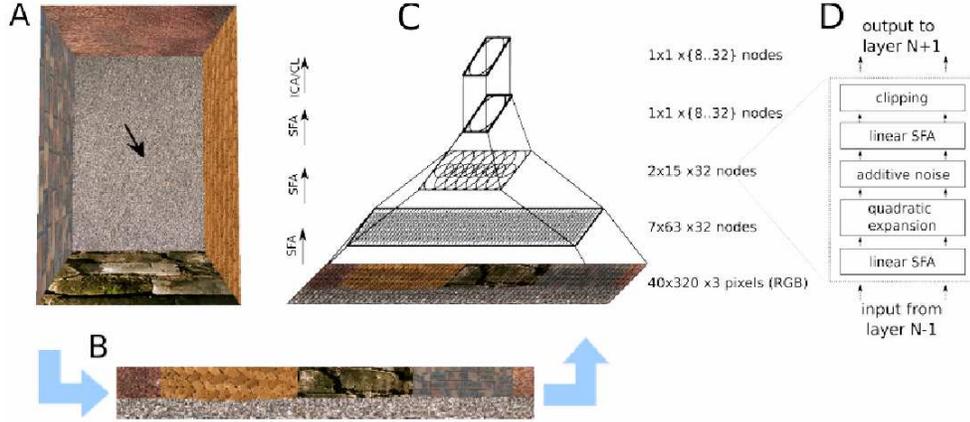}
\begin{center}
\vspace{-.3cm}
\caption{Experimental setup of Franzius et al. \cite{Franzius07}. See text for details.}
\label{fig:FranziusHierarchy}
\end{center}
\end{figure}

Franzius et al. \cite{Franzius07, Franzius08} developed
a system that exhibits place and head-direction cell like characteristics
by applying SFA to the video of a virtual rats random walk.
First, the random walk of a virtual rat in a rectangular room
(fig. \ref{fig:FranziusHierarchy}a) was recorded with a
virtual wide angle camera mounted on the head
(fig. \ref{fig:FranziusHierarchy}b).
The single frames were segmented in overlapping
patches (\textit{nodes} in fig. \ref{fig:FranziusHierarchy}c) 
in line with the concept of receptive fields.
SFA was performed on all patches, then the output
was rearranged to overlapping receptive fields 
of the next layer, and so forth.
This way the multi-layer network was able to
cope with the high dimensional video input,
despite the fact that \textit{linear SFA}
(sec. \ref{sec:SFA_linear}) with a 
\textit{quadratic expansion} 
(sec. \ref{sec:SFA_expanded}) 
of the patches was used for layer wise training
(fig. \ref{fig:FranziusHierarchy}d).

\begin{figure}[t]
\begin{center}
\includegraphics[scale=.9]{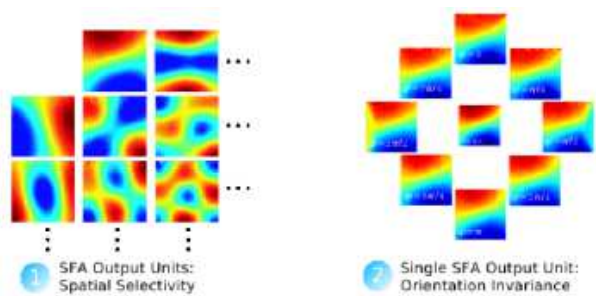}
\caption{SFA responses from Franzius et al. \cite{Franzius07}.
Every pixel in a sub-image represent one position in
a square room. The colors indicate SFA responses.
Image source: Poster presentation by M. Franzius
at the 2007 meeting of the German neuro-science society.}
\label{fig:Franzius07}
\vspace{-0.5cm}
\end{center}
\end{figure}

By using a suitable (i.e. powerful enough) function class
in the SFA training, the output of the last layer
has to resemble the theoretical predictions of proposition
\ref{pro:opt_res_3d}. Since the mean \textit{rotational}
and \textit{translational velocity} of the rat
influences the slowness
of the optimal responses $s(\phi_{ijl}^*)$ and therefore
their ordering, it is possible to extract
\textit{orientation} or \textit{location invariant} 
features by controlling the random walk.
Slow movements with fast rotations lead to orientation
invariance in the first SFA filters (fig. \ref{fig:Franzius07}).
Slow rotations and fast movements, on the other hand,
will produce position invariance.

If we want to derive place cells, higher SFA filters are useless.
The first orientation dependent filter $\phi_{001}^*$ will mix with
all orientation invariant filters of equal slowness,
so the mixed output will not be invariant to orientation anymore.
Therefore, the virtual rat moved slow and rotated fast to
obtain the results in figure \ref{fig:Franzius07}.

The last step of Franzius et al. (fig. \ref{fig:FranziusHierarchy}c) was
the application of \textit{independent component analysis}
(ICA, \cite{Hyvarinen99}) on the last layers output.
The results (fig. \ref{fig:FranziusPC}b) resemble the measured
place cells in rats (fig. \ref{fig:FranziusPC}a).

\newpage

\begin{figure}[b]
\begin{center}
\includegraphics[scale=.33]{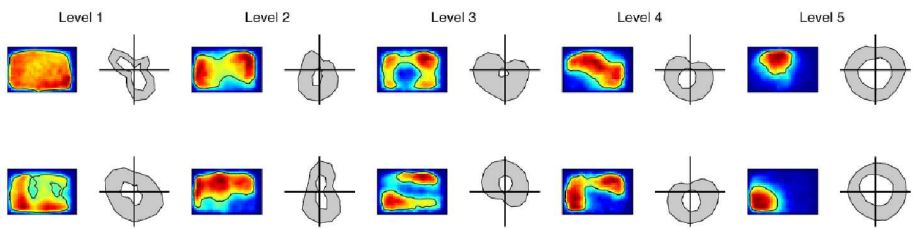}
\vspace{-.5cm}
\caption{Two examples for every layer
of Wyss et al. \cite{Wyss06}.
The left of each example is the mean response at every position
and the right represents the orientation stability in all directions
within one standard deviation (gray).}
\label{fig:Wyss06}
\end{center}
\end{figure}

Another approach to a similar problem, that
aimed to understand the diversification
in the visual system, has been made by 
Wyss et al. \cite{Wyss06}.
They designed an online algorithm
based on equation \ref{eq:sfa_problem_alt} with
a nonlinear function class.

As for Franzius et al., the key element was
a hierarchical processing of overlapping
receptive fields with the same learning
principle in every layer.
The video images were of smaller
resolution and they used 5 instead of 3 layer.
The main differences resides in the
use of a real robot and another
magnitude of training examples:
66h at 25 Hz $\approx 6 \cdot 10^6$ frames \cite{Wyss06}
in comparison to ''a few laps within 5,000 samples'' \cite{Franzius07}.

The results of the last layer are orientation 
invariant and look like (big) 
place cells (fig. \ref{fig:Wyss06}), but since no theoretical solutions are
available, it is hard to comprehend why.
In the end the results of Wyss et al.
set up more questions than they answer,
so this thesis sticks with the methodology of
Franzius et al. 

\section{Algorithms}
In this thesis we want to use optimization
problem \ref{def:sfa_problem}, i.e. the formalism
developed by Wiskott et al.
First we will review linear SFA and its common
extension, to finally define a kernelized SFA algorithm.
The latter is novel in this form, but has similarities to
existing works.

\subsection{Linear SFA}
\label{sec:SFA_linear}
We aim to find a set of $k$ functions that 
solve minimization problem \ref{def:sfa_problem}
with respect to $n$ observations
$\{\ve{x}^{(i)}\}_{i=1}^n \subset \R^d$.
Despite the name, we consider affine functions 
$\phi_i(\ve x) = \ve w_i^\top \ve x - c_i$
as model class.
To keep the notation simple, 
we use a matrix notation of 
filter responses ($\mat \Phi$ and $\mat{\dot{\Phi}}$,
as defined before) and observations 
$\mat{X}_{it} = x_i^{(t)}$  and
$ \mat{\dot{X}}_{it} = x_i^{(t+1)} - x_i^{(t)}$

The $k$ solutions are calculated simultaneously 
in 3 steps \cite{Wiskott02}:
\begin{enumerate}
\item \textit{Centering of the data}:  
$\ve{x}_c = \ve{x}-\ve{\bar x}$, 
where $\ve{\bar x} = \E[\ve{x}]$.
\item
\textit{Sphering of the centered data}: 
$\ve{x}_s = \mat{S}\ve{x}_c$, 
where $\E[\ve{x}_s \ve{x}_s^\top] = \mat{I}$. 
Sphering establishes \textit{unit variance} 
and \textit{decorrelation}.
An eigenvalue decomposition of the covariance matrix 
$\E[\ve{x}_c \ve{x}_c^\top] = {1 \over n} \, 
\mat{X}\mat{X}^\top  - \ve{\bar x} \ve{\bar x}^\top
= \mat{U} \mat{\Lambda}\mat{U^\top}$,
can be used to determine the \textit{sphering matrix}
$\mat{S} = \mat{\Lambda}^{-1/2} \mat{U}^\top$.
\item
\textit{Minimizing the slowness}: 
This can only be achieved by a rotation of the sphered data, 
since any other operation would violate 
the already fulfilled constraints.
Again, an eigenvalue decomposition of a covariance matrix 
$ \E[\ve{\dot{x}}_s \ve{\dot{x}}_s^\top] = 
{1 \over n-1} \mat{S} \mat{\dot{X}} 
\mat{\dot{X}}^\top \mat{S}^\top 
= \tilde{\mat{U}} \tilde{\mat{\Lambda}} \tilde{\mat{U}}^\top$
can be used. The solutions are given by the 
eigenvectors $\tilde{\mat{U}}_k$ corresponding
to the $k$ smallest eigenvalues.
\end{enumerate}

\begin{algorithm}[t]
\caption{Linear SFA}
\label{alg_linearSFA}
\begin{algorithmic}
\REQUIRE $k \in \N^+; \; \{\ve x^{(i)}\}_{i=1}^n \subset \R^d$
\STATE $ $
\STATE // Collect covariance matrices
\STATE $\ve {\bar x}_1$ = $\ve x^{(1)}$
\STATE $\mat C_1$ = $\ve x^{(1)} \ve x^{(1)\top}$
\STATE $\mat{\dot C}_1$ = zeros$(d,d)$
\FOR {$i = \{ 2, \ldots, n \}$}
  \STATE $\ve{\dot x}^{(i)} = \ve{x}^{(i)} - \ve{x}^{(i-1)}$
  \STATE $\ve {\bar x}_i = \ve {\bar x}_{i-1} + \ve x^{(i)}$
  \STATE $\mat C_i = \mat C_{i-1} + \ve x^{(i)} \ve x^{(i)\top}$
  \STATE $\mat{\dot C}_i = \mat{\dot C}_{i-1} + \ve{\dot x}^{(i)} \ve{\dot x}^{(i)\top}$
\ENDFOR
\STATE $ $
\STATE // Calculate sphering matrix $\mat S$
\STATE $\mat U \mat \Lambda \mat U^\top$ 
  = eig$\left({1 \over n} \, \mat C_n  - \frac{1}{n^2}\ve{\bar x}_n \ve{\bar x}_n^\top \right)$
\STATE $(\mat{\hat U}, \mat{\hat \Lambda})$ = remove\_zero\_eigenvalues$(\mat U, \mat \Lambda)$
\STATE $\mat S = \mat{\hat \Lambda}^{-1/2} \mat{\hat U}^\top$
\STATE $ $
\STATE // Find $k$ slowest direction in sphered data
\STATE $\mat{\tilde U} \mat{\tilde \Lambda} \mat{\tilde U}^\top$
        = eig$(\frac{1}{n-1} \mat S \mat{\dot C}_n \mat S^\top)$
\STATE $(\mat{\tilde U}_k, \mat{\tilde \Lambda}_k)$ 
        = remove\_all\_but\_lowest\_k\_eigenvalues$(\mat{\tilde U}, \mat{\tilde \Lambda}, k)$
\STATE $ $
\STATE // Return affine function $\phi(\ve x) = \mat W^\top \ve x - \ve c$
\STATE $\mat W$ = $\mat S^\top \mat{\tilde U}_k$
\STATE $\ve{c} = \frac{1}{n} \mat W^\top \ve{\bar x}_n$
\STATE \textbf{return} ($\mat W, \ve c$)
\end{algorithmic}
\end{algorithm}

Combining the steps, the affine function $\ve{\phi}$ is
given by $\ve{\phi}(\ve{x}) = \mat{W}^\top \ve{x} - \ve c$, where
$\mat{W} = \mat{U} \mat{\Lambda}^{-1/2} \tilde{\mat{U}}_k^\top$
and $\ve c = \mat W^\top \ve{\bar x}$.
Since the eigenvalue decompositions have complexity $O(d^3)$,
algorithm \ref{alg_linearSFA} 
has an overall complexity of max($O(d^2n)$,$O(d^3)$).

\subsection{Expanded SFA}
\label{sec:SFA_expanded}
When linear SFA does not yield sufficient slowness,
i.e. linear functions are too weak, we might wish
to use a nonlinear function class.
A common way is to expand the data into a nonlinear
feature space and perform linear SFA on the
expanded data \cite{Wiskott02}.

The most popular expansion is the set of
all monomials up to degree 2 
(\textit{quadratic expansion})
\cite{Berkes05a, Franzius07} or degree 3 \cite{Berkes05b}.
Linear SFA with these expansions is sometimes
referred to as \textit{quadratic SFA} and \textit{cubic SFA},
respectively.
Since the dimensionality of expanded inputs
grows rapidly, most authors reduced
it first with a \textit{principal
component analysis} (PCA \cite{Schoelkopf97}) 
\cite{Berkes05a, Berkes05b, Franzius07}.
Because monomials of extreme positions
can also take extreme values,
Franzius et al.~included a clipping of
high outputs in their procedure \cite{Franzius07}.

\vspace{1cm}
\subsection{Kernel SFA}
\label{sec:SFA_kernel}
The idea of projection into a high dimensional
feature space reminds of kernel techniques,
so we might avoid explicit expansions
and develop a \textit{kernelized SFA} algorithm, instead.
For a related cost function
a kernel approach has already been made by Bray and Matrinez \cite{Bray02}.
Besides the difference in the cost function their approach 
differs from this one because they do not 
use a kernel matrix but work directly with 
support vectors. Using the projected process 
kernel matrix approximation method,
their approach turns out to be a special case of this algorithm.
In the following let $k(\cdot,\cdot)$ be a given kernel 
and $\ve \psi(\cdot)$ the corresponding feature mapping.
For a definition of these terms, read section \ref{sec:kernel}.
Again, we group the data projected in feature space in a
matrix $\mat\Psi_{it} = \psi_i(x^{(t)})$.


In line with linear SFA, the kernel SFA algorithm consists of three steps:
\begin{enumerate}
\item 
\textit{Centering} can be performed directly on the kernel matrix 
(sec. \ref{sec:kernel_cov}):
$\mat{K}_c = (\mat{I} - \frac{1}{n} \ve{1}\ve{1}^\top) \mat{K} 
(\mat{I} - \frac{1}{n} \ve{1}\ve{1}^\top)$.
Note that we implicitly centered the data in
feature space this way, since $\mat K_c = \mat \Psi_c^\top \mat \Psi_c$.

\item[2.]
\textit{Sphering}: $\quad$ As discussed in section \ref{sec:kernel_cov}, 
we substitute the nonsingular eigenvectors 
$\mat{U}$ in the sphering matrix (in feature space)
with those of the kernel matrix:
$\mat{S} = \frac{1}{\sqrt{n}} \mat{{\Lambda}}_r^{-1} \mat{{V}}_r^\top \mat{\Psi}^\top$.

\item[3.]
\textit{Minimizing the slowness}: Let $\mat{\dot{\Psi}} := \mat{\Psi} \mat{D}$, with
$\mat{D} \in \R^{n \times n-1}$ being a matrix which is zero everywhere 
except for $\mat{D}_{i,i} = -1$ and $\mat{D}_{i+1,i} = 1$.
This way we can express the covariance matrix 
${1 \over n-1}\mat{S}\mat{\dot{\Psi}}\mat{\dot{\Psi}}^\top \mat{S}^\top$ 
in feature space as
${1 \over n(n-1)} \mat{\Lambda}^{-1} \mat{V}^\top \mat{\dot{K}} 
  \mat{\dot{K}}^\top \mat{V} \mat{\Lambda}^{-1} $
with $\mat{\dot{K}} = \mat{\Psi}^\top \mat{\Psi} \mat{D}$.
The eigenvectors corresponding to the smallest $k$ eigenvalues 
of this covariance matrix are the rotations 
$\tilde{\mat{U}}_k$ which optimize the problem.
\end{enumerate}
The kernel solution to optimization problem \ref{def:sfa_problem}
is given by $\ve{\phi}(\ve{x}) = \mat{A}^\top \ve{k(x)} - \ve{c}$,
with the weight matrix $\mat{A} = (\mat{I} - \frac{1}{n} \ve{1} \ve{1}^\top)\,
\mat{V}_r \mat{\Lambda}_r^{-1} \tilde{\mat{U}}_k^\top$,
the kernel expansion 
$\ve{k}(\ve{x}) := [k(\ve{x}^{(1)},\ve{x}), \ldots, k(\ve{x}^{(n)},\ve{x})]^\top$ 
and the bias $\ve{c} = \frac{1}{n} \mat{A}^\top \mat{K} \ve{1}$.

The algorithm collects the matrix $\mat{K}$ and derives $\mat{\dot{K}} \mat{\dot{K}}^\top$
from it. The two eigenvalue decompositions in steps 2 and 3 are responsible
for an overall complexity of $O(n^3)$.

\newpage
\paragraph{Kernel matrix approximation}
Kernel SFA performs much better than linear SFA,
but raises the complexity to $O(n^3)$.
One would like to apply the \textit{projected process}
method described in section \ref{sec:kernel_approximation}, 
but this method only provides approximations of $\mat{K}^2$.
As already discussed, 
$\mat{K}^2 = \mat{V} \mat{\Lambda}^2 \mat{V}^\top$
has the same eigenvectors and squared eigenvalues of $\mat{K}$.
If we perform the eigenvalue decomposition on $\hat{\mat{K}} \hat{\mat{K}}^\top$
and take the square-root of the eigenvalues, the solution
is the projected process approximation of $\mat{K}$.

With this approximation, algorithm \ref{alg_kernelSFA} has complexity $O(m^2n)$ 
because it requires the collection of the matrices 
{\small $\mat{\hat{K}} \mat{\hat{K}}^\top$}
and {\small $\mat{\dot{\hat{K}}} \mat{\dot{\hat{K}}^\top}$}.
Additionally, one has to pick a set of support vectors, 
which usually is costly, too. 
For more information on the selection of support vectors, 
read section \ref{sec:kernel_sv_selection}.

\begin{algorithm}[t]
\caption{Kernel SFA with kernel matrix approximation}
\label{alg_kernelSFA}
\begin{algorithmic}
\REQUIRE $\ve {\hat k}: \R^d \times \PM{\R^d} \to \R^m; \; k \in \N^+; \; 
          \{\ve{\hat x}^{(i)}\}_{i=1}^m \subset \{\ve x^{(i)}\}_{i=1}^n \subset \R^d$
\STATE $ $
\STATE // Collect kernel matrices
\STATE // In the following: $\ve{\hat k}^{(i)} := \ve{\hat k}
          \big(\ve x^{(i)}, \{\ve{\hat x}^{(i)}\}_{i=1}^m\big)$
\STATE $\ve {\bar k}_1$ = $\ve{\hat k}^{(1)}$
\STATE $\mat{\hat K}\mat{\hat K}^\top_1$ =
         $\ve{\hat k}^{(1)} \ve{\hat k}^{(1)\top}$
\STATE $\mat{\dot{\hat K}}\mat{\dot{\hat K}^\top_{\rm 1}}$ = zeros$(m,m)$
\FOR {$i = \{ 2, \ldots, n \}$}
  \STATE $\ve{\dot k}^{(i)} = \ve{\hat k}^{(i)}
            - \ve{\hat k}^{(i-1)}$
  \STATE $\ve {\bar k}_i = \ve {\bar k}_{i-1} 
            + \ve{\hat k}^{(i)}$
  \STATE $\mat{\hat K}\mat{\hat K}^\top_i = \mat{\hat K}\mat{\hat K}^\top_{i-1} 
          + \ve{\hat k}^{(i)} \ve{\hat k}^{(i)\top}$
  \STATE $\mat{\dot{\hat K}}\mat{\dot{\hat K}^\top_{\it i}} 
          = \mat{\dot{\hat K}}\mat{\dot{\hat K}^\top_{\it i-{\rm 1}}} 
          + \ve{\dot k}^{(i)} \ve{\dot k}^{(i)\top}$
\ENDFOR
\STATE $ $
\STATE // Calculate sphering matrix $\mat S$
\STATE $\mat V \mat \Lambda \mat V^\top$ 
  = eig$\big( (\mat I - \frac{1}{n} \ve 1 \ve 1^\top) \mat{\hat K}\mat{\hat K}^\top_n
               (\mat I - \frac{1}{n} \ve 1 \ve 1^\top) \big)$
\STATE $(\mat{V}_r, \mat{\Lambda}_r)$ = remove\_zero\_eigenvalues$(\mat V, \mat \Lambda)$
\STATE $\mat S = \mat{\Lambda}_r^{-1/2} \mat{U}_r^\top$
\STATE $ $
\STATE // Find $k$ slowest direction in sphered data
\STATE $\mat{\tilde U} \mat{\tilde \Lambda} \mat{\tilde U}^\top$
        = eig$\big(\frac{1}{n(n-1)} \mat S 
          \mat{\dot{\hat K}}\mat{\dot{\hat K}^\top_{\it n}} \mat S^\top \big)$
\STATE $(\mat{\tilde U}_k, \mat{\tilde \Lambda}_k)$ 
        = remove\_all\_but\_lowest\_k\_eigenvalues$(
          \mat{\tilde U}, \mat{\tilde \Lambda}, k)$
\STATE $ $
\STATE // Return kernelized function $\phi(\ve x) 
         = \mat A^\top \ve{\hat k}\big(\ve x,\{\ve{\hat x}^{(i)}\}_{i=1}^m\big) - \ve c$
\STATE $\mat A$ = $\mat S^\top \mat{\tilde U}_k$
\STATE $\ve{c} = \frac{1}{n} \mat A^\top \ve{\bar k}_n$
\STATE \textbf{return} ($\mat A, \ve c$)
\end{algorithmic}
\end{algorithm}

\newpage
\section{Conclusion}
\textit{Slow feature analysis} is a unsupervised
method to extract the current state out of
\textit{any} data with a temporal structure.
It learns an array of filters that aim to
extract all \textit{non-static} parameters
that distinguish the input from
other samples.
Given a powerful function class
and enough training samples, 
the output will converge to a
\textit{trigonometric representation} of the current state,
as far as it can be extracted out of the current input.
As such it is well suited for the approximation
of continuous functions on this state,
as the discussion of trigonometric representations
in section \ref{sec:trig_poly} shows.

However, instead of converging to the theoretical
\textit{optimal responses}, filters with equal slowness
can mix linearly.
This is not a problem for function approximation,
since the mixed solutions span the same space
of trigonometric basis functions.
If, on the other hand, one is only interested in
a subset of the state components,
e.g. only the spatial component of the
state space described in proposition \ref{pro:opt_res_3d},
the mixture prevents simple exclusion of 
unwanted filters.
Because the number of filters grows exponentially in the
number of state components, every unwanted
component is a nuisance.
If possible, one can obtain at least
the first few filters undisturbed by
controlling the relative velocity between state
components in the training set.

Under controlled circumstances (e.g. in a simulator)
we are interested in the whole state space
and can therefore ignore these effects.
Real life experiments, however, have shown
many additional components,
e.g. altitude of the sun, artificial light, etc.
Every independent source of change in the light conditions
is considered a state component as well as
any change in the actual scene.

At last one has to point out that this method
can only be applied to \textit{static} scenes.
Something as complex as a group of humans
induces a state space far too complex for any
function class feasible in the near future.

\chapter{Empirical validation}
\label{chapter:empiric}


\section{Experiment description}
The main experiment of this thesis is a 
\textit{proof of concept} of the following proposition:

\begin{proposition}
Policy iteration can learn a control based only
on a current video image, which is preprocessed
with filters obtained by slow feature analysis.
\end{proposition}

We showed in chapter \ref{chapter:sfa} that
SFA, applied on a random walk in bounded state spaces,
converges to filters that form a trigonometric
representation of this state space.
This claim holds, independent of the representation
in which the state is originally presented to SFA.
Therefore, the state can be presented in form
of video images, recorded from the perspective of
a robot which is moving through a static scene.
SFA will converge to a trigonometric representation
of the robots position.

As we have shown in chapter \ref{chap:rl}, the policy
iteration method LSPI is able to learn a control
out of a random walk presented in \textit{any} representation.
Of course, the quality of the control depends on representation
and considered function class. 
However, we have also shown
that trigonometric representations are especially
suited for the class of linear functions,
as employed by LSPI.

To validate this approach, we chose a navigational experiment
in which a robot is supposed to drive into a \textit{goal area}.
The only available sensor is a \textit{head mounted camera}.
Obviously, its video images are too high dimensional\footnote{
LSPI has a complexity of $O(a^2d^2)$ in space and $O(a^3d^3)$ in
time, where $a$ is the number of actions and $d$ the dimensionality
of the state representation.} and too complex\footnote{
For example, a small rotation of the robot leads 
to drastic changes in nearly every pixel.}
to promise LSPI solutions with reasonable quality.

However, as results in section \ref{sec:nav_sfa} show,
after preprocessing the learned control
was able to find the goal in $\approx 80\%$
of the test trials.

\newpage
\subsection{Overview}

\begin{figure}[t]
\begin{center}
\includegraphics[scale=.5]{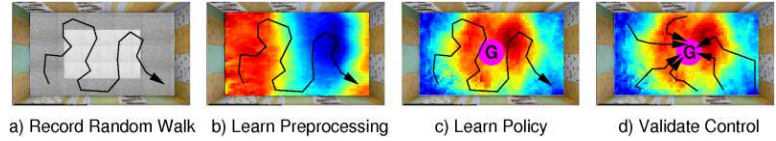}
\caption{The experiment: a) Record a random walk video with the robot.
         b) Learn the SFA filters based on this video.
         c) Estimate the Q-value based on the preprocessed video.
         d) Validate the control at random start positions.}
\label{fig:experiment}
\end{center}
\end{figure}

The main experiment is conducted in 4 phases,
as depicted in figure \ref{fig:experiment}:
\begin{enumerate}
\item[a)]
Record of a \textit{random walk video} with a camera mounted
on the head of a robot.
\item[b)]
Learn a \textit{preprocessing} 
based on the random walk video with kernel SFA.
\item[c)]
\textit{Estimate the Q-values} based on the preprocessed
random walk video with LSPI, to derive a near optimal policy.
\item[d)]
\textit{Validate} the learned control
at random positions by choosing the action associated
with the highest Q-value until
the robot reaches the goal.

\end{enumerate}

\subsection{Formal description}

\paragraph{Task}
The task is to navigate as quick as possible into a
\textit{goal area}. This area is not marked or
discriminable to other parts of the environment,
besides by the reward given when the robot is inside.
The only other reward is a punishment for getting
too close to walls, i.e. most of the environment
is without reinforcement signal.

\paragraph{Environment}
The environment is \textit{static}, \textit{bounded} and 
the robots position \textit{recognizable}
almost anywhere based on the camera images.
That means we have a closed room which is
asymmetric at least in the textures and therefore
has few positions that resemble each other,
which is the case in many indoor scenarios.
It is the claim of staticity that introduces the most problems,
since it restricts the application to uninhabited
areas. It is, however, a restriction of the slow
feature analysis and therefore necessary for this thesis.

\paragraph{Behaviour}
We assume decisions at \textit{discrete time steps}
between a limited number of \textit{discrete actions}.
As a side effect, the discretization 
liberates us of any time constraint.
The learning algorithm does not know the actions
in use, but they have to
allow the robot to navigate anywhere.
For our experiments we chose 3 actions:
\textit{move forward} ca. 30cm and
\textit{turn left/right} ca. 45 degrees.

\subsection{Robot}

\begin{figure}[t]
\begin{center}
\includegraphics[scale=1]{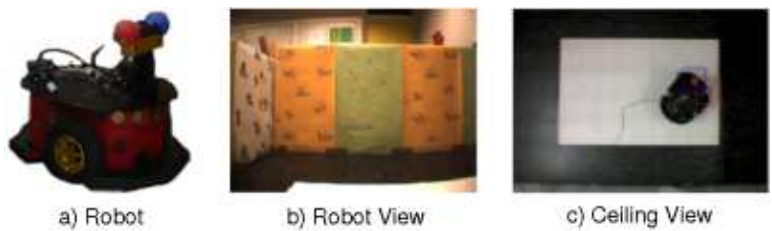}
\caption{The wheeled \textsc{Pioneer 3DX} 
        robot used in the experiments,
        a picture from the head mounted camera and
        from the observing camera at the ceiling.}
\label{fig:problem_description}
\end{center}
\end{figure}

For simplicity, the robot is assumed to be 
\textit{wheeled} and is able to
\textit{execute commands}
like ''turn 45 degrees'' or ''move 30cm''.
However, the execution can be flawed, e.g.
the robot can turn 48 degrees instead of 45
or move 32cm instead of 30.
The \textsc{NeuRoBot} project used the 
\textsc{Pioneer 3DX} robot build by
\textsc{Mobilerobots Inc} 
(fig. \ref{fig:problem_description}a).

\paragraph{Camera}
The head mounted camera has a \textit{wide field of view}
to impede similar images at different positions.
Beside this, the only restrictions are a reasonable
resolution and image quality.
The \textsc{Bumblebee}\TReg camera 
used by \textsc{NeuRoBot} has a field of view
of 66\Deg, which seems to be sufficient anywhere
but directly in front of walls.
It also records a pair of stereo images, 
of which we only considered the left one
(fig \ref{fig:problem_description}b).

\paragraph{Other sensors}
For navigation, no other sensors are necessary.
The considered policy iteration method,
however, requires a \textit{random walk}
before training.
This implies at least one dependable sensor
system to avoid/react to walls during this phase.
The \textsc{Pioneer 3DX} is equipped with ultrasonic
sensors and sensitive bumpers.
To treat the test environment with care,
we employed the ultrasonic sensors to avoid walls
during random walk.

\paragraph{Validation}
To validate the learned control, one needs to
know the robots position in training
and testing.
The internal position estimator of the
\textsc{Pioneer 3DX} turned out to be useless,
since local estimation errors are integrated
over time. After a few minutes, the error was unacceptable.
Instead, we installed a web cam at the ceiling
and extracted the position of the illuminated blue
and red ball on the robots head from it
(blue and red squares in fig. \ref{fig:problem_description}c).

\newpage
\subsection{Environment}
\label{sec:exp_env}
In more detail, the requirements on the environment are:
\begin{itemize}
\item \textit{Bounded space}: Training SFA requires to travel 
      the whole state space, multiple times if possible.
      An open field (without boundaries) would make
      this impossible, therefore we need a closed room.
\item \textit{Asymmetry}: If two positions produce the same image, 
      SFA will assign both positions the same response.
      This violates the assumption that the learned representation of
      states is injective (def. \ref{def:representation}).
\item \textit{Static scene}:  
      The robots position is the only variable in the environment.
      If other factors change the image reliably, e.g.
      whether an operator is in the room or not,
      they will introduce new SFA filters.
      This would increase the number of extracted filters
      without gaining additional information useful for
      the navigation problem.
\end{itemize}

\begin{figure}[t]
\begin{center}
\includegraphics[scale=.275]{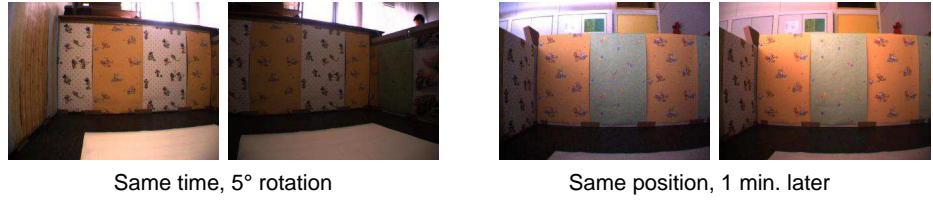}
\caption{Illumination changes drastic under daytime conditions.}
\label{fig:illumination}
\end{center}
\end{figure}

\begin{figure}[b]
\begin{center}
\includegraphics[scale=.7]{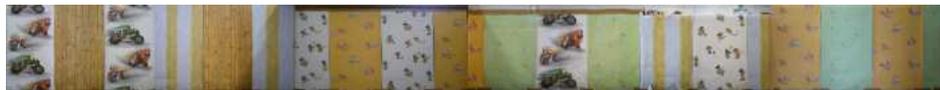}
\caption{Randomly assembled wallpaper of the experimental environment.}
\label{fig:textures}
\end{center}
\vspace{-.4cm}
\end{figure}

As it turns out, natural light conditions are not static at all.
The suns position and every cloud shifting in front of it
changes illumination and reflection in the scene.
Due to the build-in brightness correction of the camera,
the images suffered shifts in color, contrast
and brightness almost independent of the robots position
(fig. \ref{fig:illumination}).
To circumvent this, we restricted our experiments to
artificial illumination at night.
Reflection and brightness still change, 
but remain almost static with respect to the robots
position.

The experimental environment used in this thesis
consists of a rectangular $3m \times 1.8m$ area, which
is bounded by tilted tables and a wall.
All four sides were covered by the randomly assembled
wallpaper in figure \ref{fig:textures}.
The walls were ca. 90cm in height, so
the robot could see parts of the laboratory and
therefore the human operators.
To avoid human interference, the upper part
of the image was clipped out (as depicted in
fig. \ref{fig:preprocessing}).

\subsection{Simulator}
\label{sec:exp_sim}

\begin{figure}[t]
\begin{center}
\includegraphics[scale=.6]{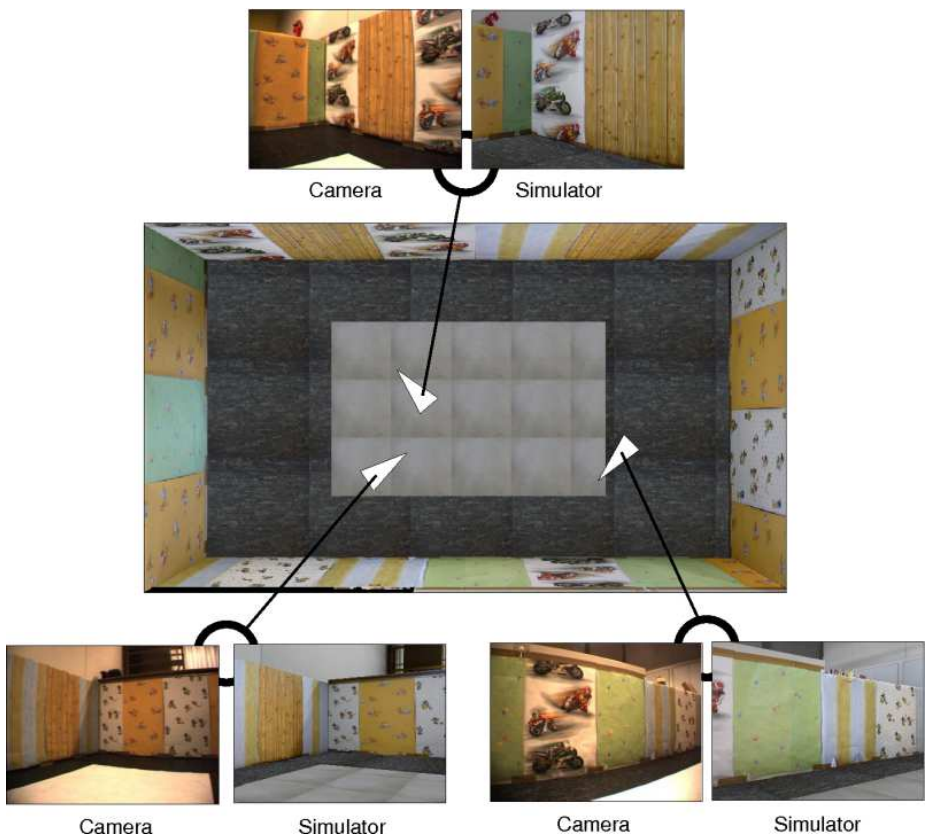}
\caption{Simulated environment with rendered images of 
         3 example positions and their real-world counterparts
         recorded by the \textsc{Pioneer} robot.}
\label{fig:setup_overview}
\end{center}
\end{figure}

Working with the \textsc{Pioneer} robot, however,
proved to be time consuming.
In addition, producing high resolution test maps
(sec. \ref{sec:exp_video}) is nearly impossible.

The author implemented a simulated version of the
experiment to verify the results in large scale tests.
The simulator is written in \textsc{Java3D} and
communicates with the \textsc{Matlab} implementation
of algorithm \ref{alg_control} by providing
a network service.
The render engine does not provide shadows or directed
illumination (only \textit{ambient light}).
Instead it relies on photographed textures of 
walls and laboratory.

Figure \ref{fig:setup_overview} shows an
overview of the rendered environment from a
ceiling camera perspective.
It also gives three example images and their
real-life counterparts, recorded by the robot.
The first difference one will notice is the
difference in color. 
The textures were photographed at daytime, the
camera images recorded at night.
Artificial light has a shift to yellow 
compared to sun light reflected on white walls.
The second inaccuracy is the unrealistic floor
texture. The whole floor as texture simply
did not fit into the graphic cards memory.
Additionally, the real floor provides
reflections that could not be modelled properly.
At last, the \textsc{Bumblebee} camera
lens distorts the outer areas of the image,
which also could not be simulated.

The simulator is not meant to copy reality
perfectly, but to allow a thorough verification
under \textit{similar} conditions.

\paragraph{Other environments}
Until now we remained inside the boundaries of theory.
We know that slow feature analysis converges to a
trigonometric representation in rectangular rooms.
The only remaining question is \textit{how well} 
(which will be answered in section \ref{sec:exp_sfa}).
Predictions for other room geometries, however,
are hard (and maybe impossible) to derive,
so the question arises how SFA will
handle such a case.
With the simulator at hand, we will 
examine this question for an example room.

\begin{figure}[t]
\begin{center}
\includegraphics[scale=.60]{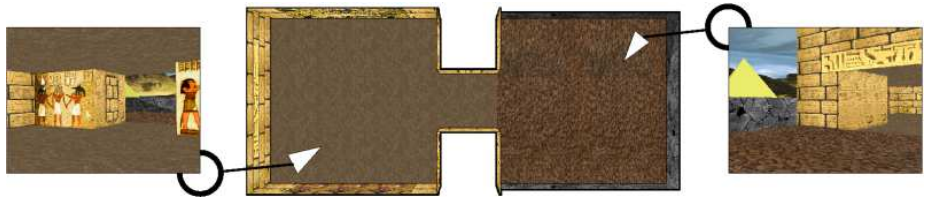}
\caption{Simulated two room environment with 
         2 example images.}
\label{fig:egypt_overview}
\end{center}
\end{figure}

Many room shapes are possible and interesting.
We chose two quadratic rooms, connected by a 
small corridor, which we will call 
\textit{two-room environment}
(fig. \ref{fig:egypt_overview}).
Since we do not have to rebuild any existing room,
we are free to use publicly accessible libraries of textures,
from which we chose an Egyptian theme.
To make the scene more appealing the author decided to
place one room outdoors, restricted by a wall
of medium height. 
To prevent similar images at different positions, 
all ''inside'' walls differ in portrayed reliefs and
''outside'' a couple of different sized 
pyramids were placed arbitrary in the far distance.

\subsection{Video generation}
\label{sec:exp_video}
Both \textit{policy iteration} and \textit{slow feature analysis}
need random walks as training sets. Both methods require
balanced sampling over the whole state space.

As simple as a random walk is to implement, 
the robot can get caught between boundaries
as trivial as rectangular corners.
Special movement statistics (as required for
\textit{place cells}, sec. \ref{sec:sfa_place_cells})
and coarse actions (as the proposed 45\Deg  rotation)
introduce additional jamming opportunities.
Instead of facing these difficulties
we decided to record one video (in three sessions) 
with movement statistics
that seem to visit all parts of the environment equally often.
Out of this video we sampled one training set for
SFA and one for policy iteration.
Since we recorded the random walk with the
ceiling camera, we were able to duplicate
it in the simulator.

The more complex the environments geometry gets,
the more likely is a random walk to get stuck.
For example, in the two room environment
the connecting corridor is a bottleneck.
It is much more likely to stay in the room
than to hit the corridor and change it.
In the long run this effect will cancel out 
statistically, but with limited training samples
it is unlikely to obtain a balanced 
set by random walk.

Since we only performed this experiment with
the simulator, we sampled the training set
as pairs of successive frames at random
start positions and movements.
This may sound unrealistic but is an easy
way to achieve guaranteed uniform distributed
training samples.

\paragraph{Test sets}
Sampling a test set (unseen in training)
out of the random walk is not complicated.
However, sampling a \textit{meaningful} test set is.

To compare the orientation/place independence
of single filters one optimally expects
a high resolution training set of images at
equidistant coordinates with the same orientation.
A set of these \textit{test maps} with different 
orientations, can give an overview of the
filter responses.

The simulator can produce exact test images, so we
created a set of 8 test maps (in 45\Deg steps) with
$64 \times 32$ equidistant coordinates, each
(e.g. in figure \ref{fig:sfa_compare}b).
Creating test maps for the real experiment, however,
is not that trivial.

To find a test set which comes near
to the described test map,
we filtered all frames within
5\Deg of the desired target map orientation.
Out of this set, we used frames as equidistant
distributed as possible.
As a result, we can only give a scatter plot
of the test map in which the test samples
can differ in angle up to 10\Deg
(e.g. in figure \ref{fig:sfa_compare}a).

\begin{figure}[b]
\begin{center}
\includegraphics[scale=.4]{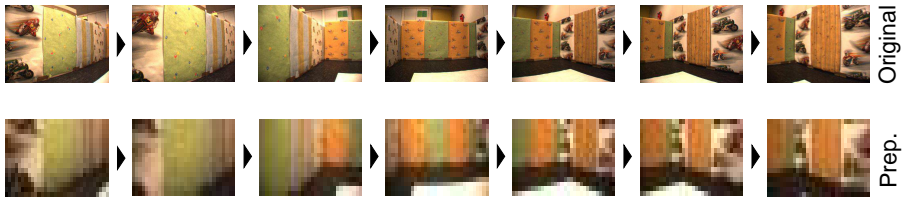}
\caption{Example video sequence before and after preparation.}
\label{fig:preprocessed_images}
\end{center}
\vspace{-.5cm}
\end{figure}

\paragraph{Image resolution}
Kernel SFA with projected process matrix
approximation (algorithm \ref{alg_kernelSFA})
has a complexity of $O(m^2)$ in space and 
$O(m^2 n)$ in time. Here $m$ is the
number of support vectors and $n$ the
number of training samples. 
This complexity does not seem to be affected by
the input images dimensionality $d$.

However, We have to store the support vectors and
perform a kernel expansion at every frame, too.
This induces an additional complexity of
$O(md)$ in space and $O(mdn)$ in time
(assuming a kernel with complexity $O(d)$ in time,
e.g. the RBF kernel).
As long as $d < m$, this cost does not
exceed the cost of performing kernel SFA
and we are relatively free to choose
the image resolution.

If we exceed this threshold significantly, however, 
the new costs depend linear on the number of pixels,
which is quadratic in image resolution.
For example a $320 \times 240$ RGB image contains
$d = 230400$ independent values which exceeds
the maximum number of support vectors used in this
thesis ($m < 8000$) by a factor of $28$.
Therfore, we scaled the image down to 
a relatively conservative resolution of $32 \times 16$
RGB pixels ($d = 1536$).
To investigate the influence of image resolution,
we also duplicated one experiment with a
resolution twice as high ($64 \times 32$ RGB pixels 
$\Rightarrow d=6144$, see section \ref{sec:exp_sfa}).

\begin{figure}[t]
\begin{center}
\includegraphics[scale=.4]{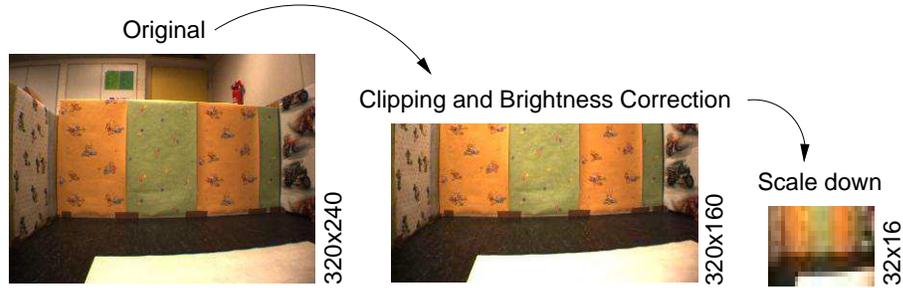}
\caption{Video image preparation steps.}
\label{fig:preprocessing}
\end{center}
\end{figure}

Keeping considerations from section \ref{sec:exp_env} in mind,
we performed a three step transformation on the
original video images before presenting them to SFA
(see fig.~\ref{fig:preprocessing} for the process and
fig.~\ref{fig:preprocessed_images} for the results):
\begin{enumerate}
\item Clip out the upper 80 lines to avoid human 
      interference from the laboratory.
\item Correct mean brightness to counter the cameras
      automatic adjustment.
\item Shrink the image to size $32 \times 16$.
\end{enumerate}

\subsection{Alternative approaches}
\label{sec:slam}
Most reinforcement learning approaches have
customized representations (see for
example Stone and Sutton \cite{Stone01}).
These are not comparable to an adaptive
technique like SFA.
To the authors knowledge, no other approach
can estimate the current position
based on a single frame.

The technique coming closest to position estimation
based on video images is called 
\textit{simultaneous localization and mapping} 
(SLAM or Visual SLAM in the context of video input). 
Originally, Smith et al. \cite{Smith90} investigated
the problem for vehicles with sonar sensors.
Roughly a decade ago, Davison \cite{Davison98}
adapted the method to \textit{Visual SLAM},
which is currently under development by several groups 
\cite{Davison03, Sim05, Eade06}.

The classical solution,
as well as many modern approaches \cite{Davison98, Eade06}
employ \textit{extended kalman filters} (EKF)
to estimate \textit{landmark} positions.
These suffer $O(n^2)$ complexity in the 
number of landmarks $n$, due to statistic dependencies
between their estimation.
Landmarks are small 2d patches 
with their estimated 3d position,
which should be as relocatable as possible.
For this, a variety of 2d transforms are in use,
e.g. affine transformation \cite{Eade06}
or SIFT \cite{Sim05}.
Comparing all possible patches with all known
landmarks is computationally expensive \cite{Sim05}.
Instead, it is possible to incorporate information
about the expected location 
in so called \textit{active measurement} strategies.
This way, Visual SLAM can be performed in real time
\cite{Davison03, Clemente07, Eade06}.

To cover larger environments, the 
\textit{Rao-Blackwellized particle filter}
was first introduced by Murphy \cite{Murphy99}
followed by a practical framework 
by Montemerlo \cite{Montemerlo02}.
Out of the perspective of every \textit{particle},
the camera position  is fixed
and the landmark estimation thus independent,
yielding $O(mn)$ complexity for $m$ particles.
In the field of Visual SLAM, Sim et al. \cite{Sim05}
and Eade et al. \cite{Eade06} successful applied 
these particle filters.

Alternatively, Estrada et al. \cite{Estrada05}
extended the EKF approach by producing
independent local maps of limited size.
These are stitched together in a global layer
thus the algorithm is called \textit{hierarchical maps}.
Recently, Clemente et al. \cite{Clemente07} 
applied this approach to Visual SLAM.

Whereas Visual SLAM is a well established 
field with remarkable success,
our approach differs 
in some mentionable ways:
\begin{itemize}
  \item SLAM depends on a specific type of sensor input,
        whereas SFA automatically adapts to any sensor.
  \item SLAM holds a current state, which is crucial
        for localization and prone to initialization
        errors. In contrast, SFA works instantaneous
        and can localize the robot using one frame only.
  \item SFA extracts a trigonometric representation
        of the position. SLAM, on the other hand,
        estimates position in 3d coordinates.
\end{itemize}
To compare SFA to SLAM, one would have to expand the
position estimates to a trigonometric representation.
These could be used instead of the SFA output to
learn a control.
Because SLAM algorithms are quite complex and not
easy to use, the author could not provide such 
a comparison in the available time.

Anyway, even if we could, SLAM is bound to have
an internal state. Comparison would only be fair
when we grant such a state to SFA, too.
This would violate the \textit{proof of concept}
approach of this thesis.

\newpage
\section{Learn preprocessing}
\label{sec:exp_sfa}
Before we learn the navigational control
in the next section, we want to
validate the theoretical predictions
of SFA as a preprocessing (chapter \ref{chapter:sfa}).
The main concern is, how well kernel SFA
performs in our context, i.e. how close
to the optimal responses one can come
with current computers.

We evaluate different SFA algorithms, which
in fact reflect different function classes
fitted to optimize the SFA cost function 
(sec. \ref{sec:sfa_problem}).
The point of reference are always the
optimal responses that can be expect in
the limit (sec. \ref{sec:sfa_opt_responses}).

\paragraph{The videos}
We recorded 3 videos in different environments
to test SFA predictions
(sec. \ref{sec:exp_env}, \ref{sec:exp_sim} \& \ref{sec:exp_video}):
\begin{itemize}
\item \textit{Robot experiment}:
      The original random walk video recorded by 
      the \textsc{Pioneer} robot.
      We reduced the frame rate to $\sim 1$Hz,
      resulting in 35128 frames under
      relative stable conditions at night.
\item \textit{Simulated experiment}:
      Due to the ceiling camera, we were able to
      record the same trajectory as in the
      robot experiment in the simulator.
\item \textit{Two room experiment}:
      To test the SFA responses in different
      geometries, we recorded a simulated
      training set in the two room environment.
      The video contains 35000 random \textit{transitions}
      at random start positions.
\end{itemize}

\paragraph{The algorithms}
We applied \textit{linear SFA} 
(algorithm \ref{alg_linearSFA}) directly on
the raw video images. 
This algorithm restricts the image resolution
stronger than kernel SFA.
Using $32 \times 16$ RGB pixel images turned out to be 
feasible without problem, but $64 \times 32$ RGB pixel
images rise the computational demand by a factor of $64$.
We therefore had to omit the latter experiment with linear SFA.

Our main focus centered on \textit{kernel SFA}
(algorithm \ref{alg_kernelSFA}).
Because all videos contain $\sim 35000$ frames,
we employed the projected process kernel matrix
approximation method with a greedy support vector
selection (algorithm \ref{alg_select_sv}).
Through the number of support vectors, one
indirectly controls the model
complexity of this approach.
Therefore it is especially suited to investigate
the SFA behaviour for different function classes.
While the complexity of kernel SFA 
is susceptible to large image resolutions
too, the $64 \times 32$ pixel resolution did not
cause overwhelming overhead, as for linear SFA.

We only considered the RBF kernel (sec.~\ref{sec:kernels}).
Other kernels seemed promising too, but RBF kernels
proved to be very reliable as long as the support vectors
are drawn out of the complete state space.
Because the greedy selection algorithm allows
only a hyper parameter $\nu$ to adjust the
sparsity of the support vector set
(sec.~\ref{sec:kernel_sv_selection}),
finding a suitable sized set demands multiple
trials with either changed $\nu$ or kernel
parameter $\sigma$.
Because we have no other means to determine 
a good $\sigma$, we decided to vary it
and keep the sparsity parameter fixed at $\nu = 0.1$.

\subsection{Slowness}
\label{sec:exp_sfa_class}

\begin{figure}[t]
\hspace{-1.0cm}
\includegraphics[scale=.53]{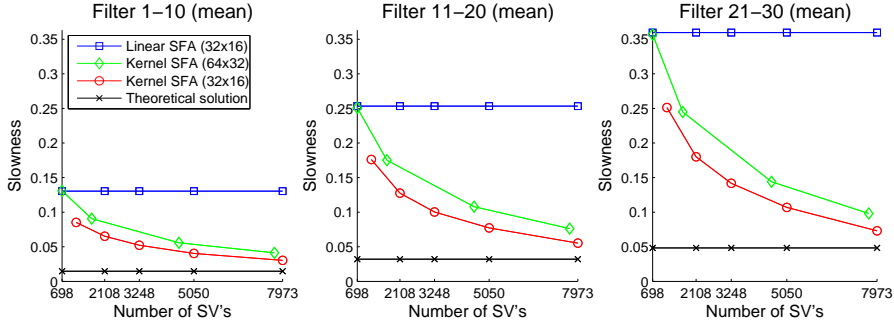}
\begin{center}
\vspace{-.3cm}
\caption{Mean slowness on the training set
         of the \textit{robot experiment} 
         for different SFA classes and
         kernel matrix approximations as 
         number of support vectors (SV).
         $32 \times 16$ and $64 \times 32$ denote the
         image resolution.
         Note that \textit{linear SFA} does not use support vectors
         and the \textit{theoretical solutions} are the
         slowest functions possible.}
\label{fig:SvC}
\end{center}
\vspace{-.5cm}
\end{figure}

\begin{figure}[b]
\hspace{-1.0cm}
\includegraphics[scale=.6]{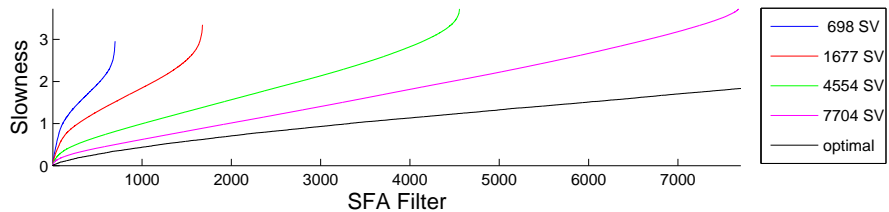}
\begin{center}
\vspace{-.7cm}
\caption{Slowness of all SFA filters in the \textit{robot experiment} for
         different number of support vectors used by kernel SFA.
         Note how close the slowest filters of the 7704 SV line 
          are to the optimum.}
\label{fig:slowness}
\end{center}
\vspace{-.5cm}
\end{figure}

The foremost question is whether the considered SFA 
algorithms are able to produce a 
trigonometric representation of the robots position.
Since SFA filters are likely to be mixtures of
optimal responses, a direct comparison is only 
possible in the slowest, unmixed filters.

Instead, one can compare the \textit{slowness} 
of SFA filters and optimal responses on the training set.
We know that no filter can be slower than the optimal
responses, so the difference in slowness can serve
as a measure how well SFA has converged to a
trigonometric representation.
Because this does not involve any test sets,
we compared the slowness in the
\textit{robot experiment}, as it is
the most realistic case.

In figure \ref{fig:slowness}, the slowness of all filters
obtained by kernel SFA with
different sets of support vectors are plotted.
One can observe that the first filters are much closer
to the optimum than latter ones, which are more likely
to be influenced by (fast) noise.

To see \textit{how} close the results are to a
trigonometric representation, figure \ref{fig:SvC} plots the
mean slowness every 10 filters
against the number of support vectors used.
Note that linear SFA 
is comparable to kernel SFA 
with $\sim 700$ SV and that
the slowness seems to converge against the
optimum with more SV.

Another noteworthy result depicted in figure
\ref{fig:SvC} is that a higher image resolution
($64 \times 32$) performs \textit{worse}
at equal model complexity.
Given that a higher resolution represents a
more complicated \textit{problem}, it makes
sense that easier problems perform better.
It can be expected that this relationship will
switch at a high model complexity, since 
the high resolution images contain more information.
Due to current computational limitations,
we were not able to verify this prediction.

\subsection{Responses}
\label{sec:exp_sfa_resp}

\begin{figure}[t]
\hspace{-1.3cm}
\includegraphics[scale=.38]{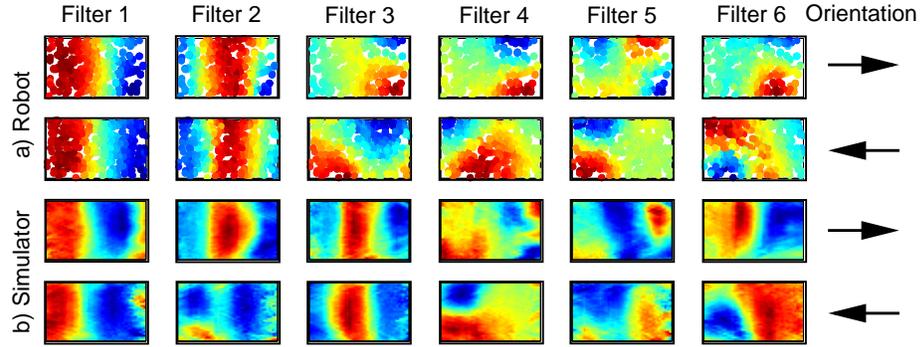}
\begin{center}
\vspace{-.8cm}
\caption{Comparison of the first 6 filters of (a) the
         \textit{robot experiment} with 7973 SV 
         and (b) the \textit{simulated experiment} 
         with 8064 SV.}
\label{fig:sfa_compare}
\end{center}
\vspace{-.5cm}
\end{figure}

\begin{figure}[b]
\hspace{-1.8cm}
\includegraphics[scale=.31]{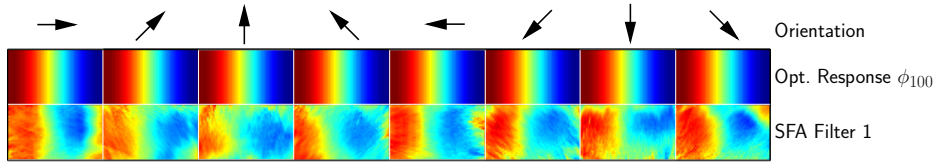}
\begin{center}
\vspace{-.5cm}
\caption{SFA filter $F_1$ of the \textit{simulated experiment} 
         resembles optimal response $\phi_{100}$.}
\label{fig:nomix}
\end{center}
\vspace{-0.5cm}
\end{figure}

Having established the performance of kernel SFA 
in terms of slowness on the training set, 
we now want to verify the predictions on test sets.
Because of the inherent inaccuracy of the robot
experiment test set, we will focus mainly on the
simulators results.

Figure \ref{fig:sfa_compare} shows test maps of the first
6 filters of both experiments in two opposite directions.
While the first filter looks more or less the same,
the later ones differ significantly.
Since the trajectory (and therefore the movement statistics)
are the same, the only explanation is the difference in images,
which led to another solution.

\begin{figure}[t]
\hspace{-1.8cm}
\includegraphics[scale=.31]{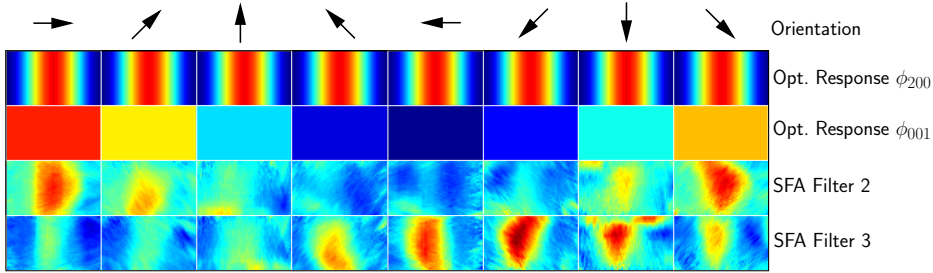}
\begin{center}
\vspace{-.5cm}
\caption{SFA filters 2 and 3 of the \textit{simulated experiment}
         with 8064 SV resemble mixtures of optimal responses $\phi_{200}$
         and $\phi_{001}$, as predicted in proposition \ref{pro:sfa_mix}.}
\label{fig:mix}
\end{center}
\vspace{-.5cm}
\end{figure}

We can also observe that the first filter in both experiments
look like the optimal response $\phi_{100}$, which is
depicted in more detail in figure \ref{fig:nomix}.
In the robot experiment, the second filter also resembles
$\phi_{200}$. In contrast, the second simulator filter
is not even orientation invariant.
The latter situation is shown in figure \ref{fig:mix}.
As theoretical predicted in proposition \ref{pro:sfa_mix},
SFA filters with the same slowness can mix,
e.g. $F_2 \approx a \phi_{200} + b \phi_{001}$
and $F_3 \approx b \phi_{200} - a \phi_{001}$
as long as $a^2+b^2=1$.
We see here a mixture of the second orientation invariant
basis function in horizontal direction $\phi_{200}$ (filter 2 in the
robot experiment) with the first place invariant basis function
$\phi_{001}$.

Filters above 3 are hardly similar to any
optimal response, but as filter 2 and 3 showed, they
might as well be arbitrary rotations in the subspace
of basis function with equal slowness.

\subsection{Other environments}
\label{sec:exp_sfa_two}

As close as the previous results came 
to the predicted responses,
it is of great practical interest how far
they will differ when we leave the
domain of theoretically predictable environments.
The responses of the \textit{two-room experiment}
should give us some insight at a simple yet common example.

\begin{figure}[t]
\hspace{-.25cm}
\includegraphics[scale=.35]{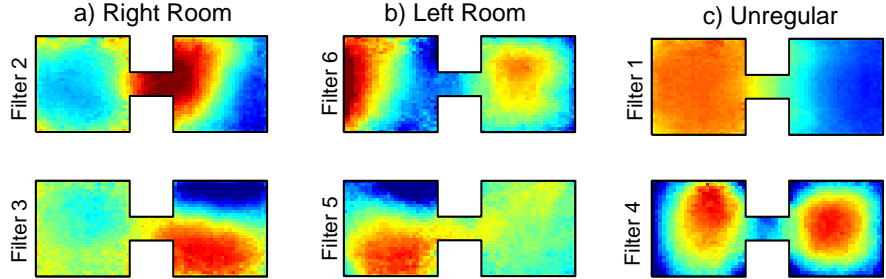}
\begin{center}
\vspace{-.5cm}
\caption{Mean response of the first 6 filters 
         of the \textit{two-room experiment} with 6800 SV.
         The specific room shape favors filters 
         that are exclusive for one room (a, b)
         but some do not fit this scheme (c).}
\label{fig:egyptfeats}
\end{center}
\end{figure}

Figure \ref{fig:egyptfeats} shows the response
of first 6 filters from a birds perspective.
In difference to the rectangular room,
the filters seem to \textit{specialize}
in single rooms.
Interestingly, in the chosen room
the response look like a
trigonometric basis function.
For example, the filters 2 and 6 resemble
the optimal response $\phi_{100}$ in one room,
but are near zero (green color) in the other.
In this sense, filters 3 and 5 resemble
optimal response $\phi_{010}$.
Higher filters are less intuitive, since as
before they mix strongly with orientation
sensitive responses of equal slowness.
However, in the limit, one would expect
\textit{two} full sets of trigonometric basis functions,
that each span one room \textit{exclusively}.

As promising as these results look in
terms of function approximation,
unexpected filters also emerge.
For example, filter 1 resembles a room ID
function which is positive (red) in the
left and negative (blue) in the right room.
While this somehow still fits in the room specific framework
described above, filter 4 seems to be active in
the center of both rooms.

\subsection{Discussion}
We were able to verify that, with growing sets of support vectors,
kernel SFA filters converge
to trigonometric basis functions of the robots position.
At some examples, we could also find evidence for
other theoretical predictions,
like mixture of basis functions with equal slowness.
Leaving the area of predicted room shapes,
the results also look promising, albeit
not all filters resemble trigonometric basis functions.

In conclusion, SFA indicates to be a well suited
preprocessing for linear value approximation.
The inherent property of mixtures in basis functions,
on the other hand, forces us to extract the
complete state space.
These mixtures seem to follow a broad interpretation
of ''equal slowness'', i.e. all features
but the first few mix with at least one other.
Without any means to un-mix the basis functions,
we can not exclude unwanted states (like humans,
sun position or blinking lights),
and are therfore restricted to static scenes.

The adaption to a two room environment
apparently led to a functional basis for each room,
and some unique additional filters.
A partition in separate rooms can actually 
be seen as an advantage.
For example, it reduces the number of involved
filters when the behaviour is learned
in one room only.
This could be exploited by a clever 
reinforcement algorithm to reduce the computation time.

Some filters do not fit in this scheme
and seem to be useless in terms of function approximation.
However, without analytical solution it is hard
to validate such statements.

\section{Learn navigation}
First, we want to validate the performance of policy
iteration with a trigonometric representation
(sec.~\ref{sec:nav_trig}).
To compare policies, we also define a quality measure
named \textit{convergence quality} and show
how well the learned control behaves compared to
a near optimal hand-crafted \textit{reference policy}.

In section \ref{sec:nav_sfa}, we will learn
navigational control based on video images only.
Here we validate the advantage an SFA preprocessing
yields and how alternative room geometries influence
the performance.

\paragraph{The algorithms}
Given a training set in any 
linear representation (def.~\ref{def:representation}),
we use LSPI (algorithm \ref{alg_lspi})
to learn the optimal policy.
LSPI employs LSQ (algorithm \ref{alg_lsq}),
which in turn calls LSTD (algorithm \ref{alg_lstd}),
to estimate Q-values.
We chose a discount factor of $\gamma=0.9$
in the laboratory and
$\gamma = 0.95$ in the two room environment
because the latter was of much bigger size.

After determining the best policy,
the control (algorithm \ref{alg_control})
navigates by letting the policy choose
one of the 3 actions: 
move forward ca. 30cm, rotate left 45\Deg
or rotate right 45\Deg.

The representation, the reward distribution,
and the training set influence the resulting policy.

\paragraph{Robot training set}
We were able to extract a set of 3091 rotations
and 5474 movements out of the random walk video
recorded by the robot.
After mirroring the rotations 
(if $x_2$ is 45\Deg to the \textit{right} from $x_1$,
then $x_1$ is 45\Deg to the \textit{left} of $x_2$),
we reached a training set of 11656 transitions.

\begin{figure}[b]
\begin{center}
\includegraphics[scale=.4]{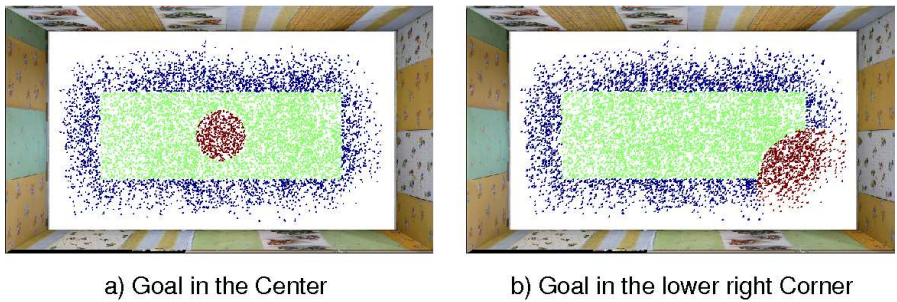}
\caption{The test set (dots) for the \textit{robot-} 
         and \textit{simulation-experiment},
         colored in received reward (blue: -1, green: 0, red: +1),
         for two different goals.}
\label{fig:pi_overview}
\end{center}
\end{figure}

A reward of +1 was given for entering the goal
area, and a punishment of -1 for coming 
too close to the walls ($<50$cm).
We tested two goals, the center of the room
with a radius of 20cm (fig.~\ref{fig:pi_overview}a) 
and the lower right corner
with a radius of 45cm (fig.~\ref{fig:pi_overview}b).

\paragraph{Two room training set}

\begin{figure}[t]
\begin{center}
\includegraphics[scale=.6]{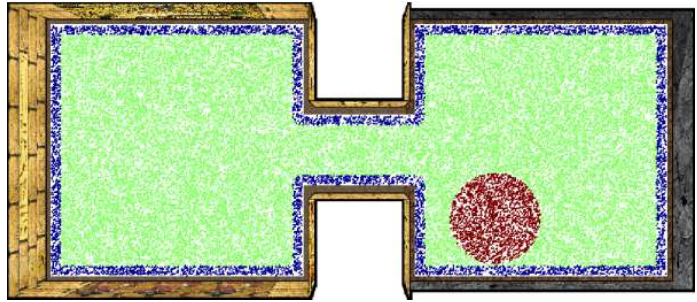}
\caption{The test set (dots) for the \textit{two-room experiment},
         colored in received reward (blue: -1, green: 0, red: +1).}
\label{fig:egyptreward}
\end{center}
\end{figure}

As discussed in section \ref{sec:exp_video},
we sampled the random transitions uniformly.
Because the environment is larger, we
decided to use a training set of 50,000 transitions.
Figure \ref{fig:egyptreward} shows the distribution
and received reward of the used training set.

It took the author some time to find a working
reward distribution.
There are many hyper parameters to set
and some were surprisingly influential.
Two observations stroke us as particular noteworthy:

(1) The discount factor $\gamma$ has an
undeniable impact on the learned control.
Large environments seem to require large $\gamma$.
One explanation would be the exponential
decay of value the more one departs from the goal area.
The quadratic cost function of LSTD distributes
the approximation error equally on all samples.
As a result, areas of close-by Q-values
will be more likely to pick the wrong action.
This includes all areas far away from the goal
where the over-all value is small.
Therefore, large environments need to ''spread''
the positive reward further, i.e. need a larger $\gamma$,
to counter this effect.

(2) When the punished distance to walls is
too large, the positive reward is not propagated
in the other room.
Most likely, the close proximity to negative reward
prevented the distant positive reward to
be propagated through the corridor.
However, since this is no theoretical handicap,
it must be a result of weak
approximation at this critical point.

\subsection{Policy evaluation}
Evaluating a learned policy is more complicated than
in the case of, for example, regression.
Applying the policy on a representative test set
(e.g. a \textit{test map} as described in
section \ref{sec:exp_video})
shows us the distribution of actions, but
does not allow any conclusions about
the controls \textit{behaviour}.
For example, there could exist ''loops''
from which the robot can not escape
and never reach the goal.
Therefore we have to evaluate the trajectories,
produced by the policy.

The initial task of this experiment was
to reach the goal area as quickly as possible.
Counting the steps  to the goal $\tau(\pi, \ve x_0)$
would be the most intuitive measure
how well a policy $\pi$ fulfills this task
at some start position $\ve x_0$.
However, since we are interested in the 
quality of \textit{all} possible
trajectories, an area of ''loops''
(i.e. that can not be left)
would drive this measure to infinity.

The approach we want to take here uses the value
$V^\pi(\ve x_0)$ to evaluate the policies fitness.
Given the simplified case that we ignore punishment
at walls and treat the goal area as absorbing states,
this measure is a modification of the former:
$V^\pi(\ve x_0) = \gamma^{\tau(\pi, \ve x_0)-1}$.
On the one hand, this measure can handle infinitely
long trajectories.
On the other hand, compared to the optimal policy $\pi^*$,
unnecessary detours near the goal have more influence on
the measure that those far away.
When this is not desired, one can normalize the result with the
optimal value $V^{\pi^*}(\ve x_0)$.
We will call the mean of this ratio the \textit{convergence quality}:

\begin{definition}[Convergence quality]
\label{def:conv_qual}
Let $p(\ve x)$ be a uniform distribution over the
state space of $\ve x$, except the goal area, and let
$V^\pi(\ve x)$ denote the value of state $\ve x$,
with a reward of 1 at the absorbing states of the goal area.
Given the optimal policy $\pi^*$,
the following will be called convergence quality:
\begin{equation}
C(\pi) = \int \frac{V^\pi(\ve x_0)}{V^{\pi^*}(\ve x_0)} \, p(\ve x_0) \; d\ve x_0
\end{equation}
\end{definition}
This measure of policy convergence has some nice properties:
\begin{itemize}
\item $C(\pi) \in [0,1]$ where 1 indicates that $\pi=\pi^*$ and
      0 that no trajectories hit the goal.
\item $\log_\gamma(C(\pi)) \approx \int \big(\tau(\pi,\ve x_0)
      -\tau(\pi^*,\ve x_0)\big) p(\ve x_0) d\ve x_0$
      (mean difference in $\tau(\cdot,\cdot)$ between $\pi$ and $\pi^*$),
      but does not go against infinity in the presence of loops.
      The approximation holds only for high $C(\pi)$.
\end{itemize}
Evaluating the integral in definition \ref{def:conv_qual}
is not possible but of course can be approximated given
example trajectories.

\paragraph{Reference policy}
The definition of convergence quality requires
the optimal policy $\pi^*$
that will depend largely on the reward distribution.
Building a control that maximizes the sum of
expected reward by hand is hardly possible.
Instead one can define a policy that aims to
reach the goal as quickly as possible.
Though this is not $\pi^*$, 
there is probably little difference in
in terms of $\tau(\pi^*,\cdot)$.

Given the coordinates of robot and goal, one can define
a simple greedy policy (algorithm \ref{alg_reference_policy}).
The algorithm does not describe the best possible policy, 
but is probably close to it.

\begin{algorithm}[t]
\caption{Reference policy}
\label{alg_reference_policy}
\begin{algorithmic}
\REQUIRE $\ve p \in \R^2; \; \theta \in [0,2\pi]$ // Position $\ve p$ and orientation $\theta$ of the robot
\REQUIRE $\ve g \in \R^2$ // Center of the goal area $\ve g$
 \STATE $\Delta\theta$ = $\theta$ $-$ direction\_of\_vector($\ve g - \ve p$)
 \IF {$|\Delta \theta| \leq 45^\circ$} 
  \STATE \textbf{return}(MOVE\_FORWARD)
 \ELSE
  \IF {$\Delta\theta < 0$}
    \STATE \textbf{return}(TURN\_LEFT)
  \ELSE
    \STATE \textbf{return}(TURN\_RIGHT)
  \ENDIF
 \ENDIF
\end{algorithmic}
\end{algorithm}

\paragraph{Two room environment}
To use the convergence quality measure in the two-room
experiment, one can enhance algorithm \ref{alg_reference_policy}
by subgoals at each end of the corridor.
The new policy would first determine
which goal is most reasonable
and then call algorithm \ref{alg_reference_policy}
with the selected (sub)goal.

\subsection{Artificial input}
\label{sec:nav_trig}

Before we apply SFA as a preprocessing, 
we want to evaluate LSPI on the predicted 
optimal responses, i.e. trigonometric polynomials.
We used the same training set, but instead of the
video images we presented the real position in a
trigonometric representation.
Besides the question how well LSPI will
perform, we are also interested in 
potential overfitting and other factors
that would complicate the main experiment.

SFA sorts its filters with respect to slowness 
on the training set, so we did the same
with the optimal responses predicted in
proposition \ref{pro:opt_res_3d}.
We omitted mixtures of filters
(which appear in SFA, as verified in the last
section), because LSTDs result is invariant to rotations
of the input space.

\begin{figure}[t]
\begin{center}
\includegraphics[scale=.35]{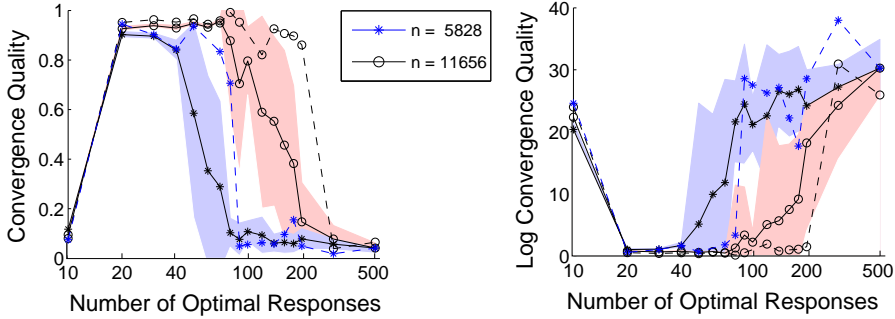}
\caption{Left: Convergence quality $C(\pi)$ on artificial
        state representations for different training
        set sizes $n$.
        Every data point on the solid line
        represents the mean over 10 trials
        with different initial random policies,
        evaluated on 1000 trajectories each.
        The colored area is the respective standard deviation.
        The dashed line represents 
        a modified initial policy.
        Right: $\log_\gamma(C(\pi))$ is an approximation
        of the the mean difference in steps-to-goal
        between the learned and the reference policy.}
\label{fig:pi_overfitting}
\end{center}
\end{figure}

Figure \ref{fig:pi_overfitting} shows on the left
the convergence quality $C(\pi)$ for the training set
(size $n=11656$) respectively the first half of it 
($n=5838$). 
The goal area resides in the center of the room.
The evaluated policies were trained with
state representations of a 
dimensionality $d$ between 10 and 500,
i.e. using the $d-1$ slowest optimal responses and a constant.
However, using huge state representations, 
policies often did not converge at all, 
so we evaluated every experiment 10 times 
with different initial policies.
The colored areas show one standard deviation
and demonstrate how \textit{reliable}
LSPI works.

Though one can observe a regime in which
the convergence quality depends largely on
the initial policy, it is unclear \textit{why}.
Apparently, some initial actions get in the way
of convergence while others lead to good results.
This might be due to slight overfitting
in the initial value estimation which
gets amplified by policy iteration,
but there is no analytical explanation
that would back up the experimental data.

A simple way to circumvent this problem
is to use all possible actions at once,
simulating a policy that truly chooses them at random.
Superficially, this would triple the
training set, but due to the state-action
representation in use (def.~\ref{def:sa_pairs})
it can be implemented without any computation
overhead.
The results of this approach are plotted
as dashed lines in figure \ref{fig:pi_overfitting}. 
Since no randomness
is involved anymore, one experiment
per representation size is sufficient.

However, to provide a more intuitive comparison, 
the right side of figure \ref{fig:pi_overfitting} 
shows $\log_\gamma(C(\pi))$,
which can be considered an approximation 
for the mean number of
additional steps the learned policy needs,
compared to the reference policy.
Note that the approximation becomes less reliable
for low convergence quality values.

In conclusion, one can observe:
\begin{enumerate}
\item Using a suitable number of optimal responses
      (small set/blue area: 20-40, full set/red area: 20-70),
      LSPI converges reliably to a policy comparable
      to the reference policy.
      The mean difference in terms of $\tau(\pi,\cdot)$ 
      in this regime is below one step.
      We will call this the \textit{working regime}.
\item More optimal responses (i.e. SFA filters)
      do not automatically produce better policies.
      While \textit{few} input dimensions yield a
      similar (even though not always good) convergence quality, 
      results of \textit{higher} dimensionality become unstable.
      Note that this effect is very similar in both curves,
      but appears earlier in the smaller training set.
      This indicates that the effect is the result of \textit{overfitting}.
\item Sampling all possible actions at once
      improves the initial policy,
      doubling the working regime in size.
      However, since the cause of the unstable regime
      is unclear, no theoretical statement
      guarantees an improvement.
\end{enumerate}

\subsection{Video input}
\label{sec:nav_sfa}

\begin{figure}[t]
\begin{center}
\includegraphics[scale=.30]{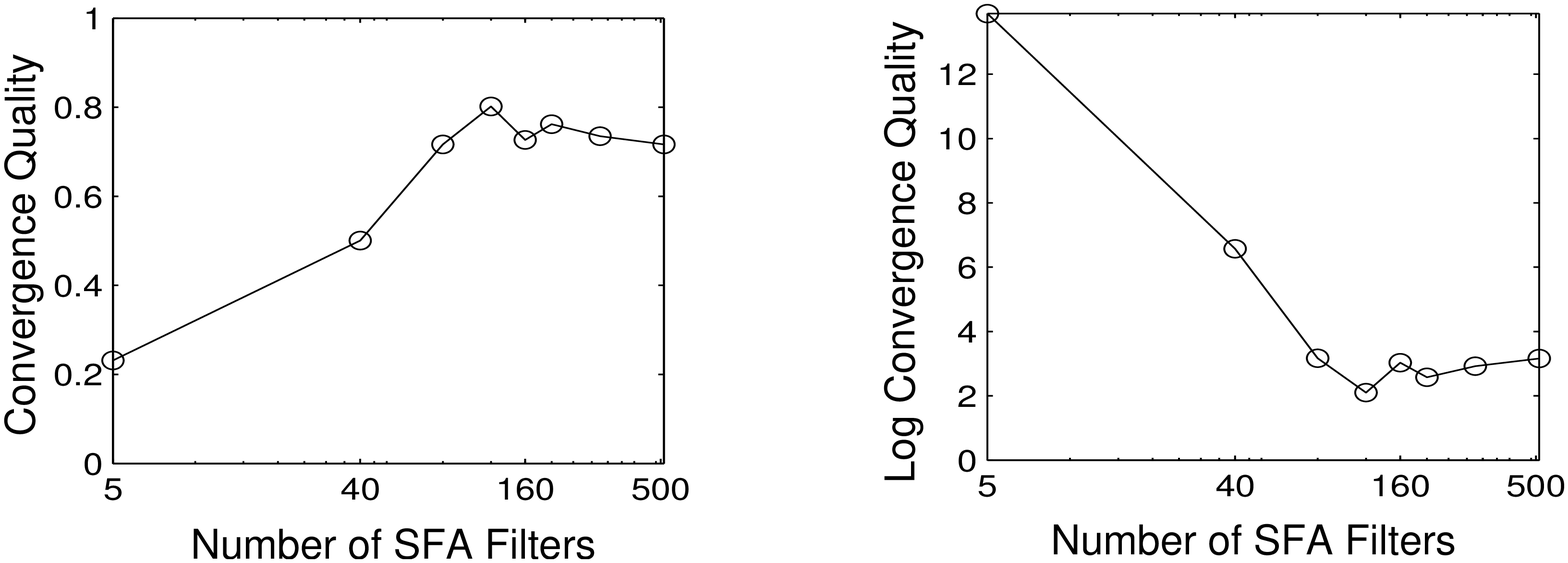}
\caption{The same experiments as in
        figure \ref{fig:pi_overfitting}
        with $n=11656$ samples, but
        on simulator video images preprocessed 
        by SFA. One trial per data point, which were
        evaluated at 200 trajectories, each.}
\label{fig:pi_nf_sim}
\end{center}
\end{figure}

\begin{figure}[b]
\begin{center}
\includegraphics[scale=.23]{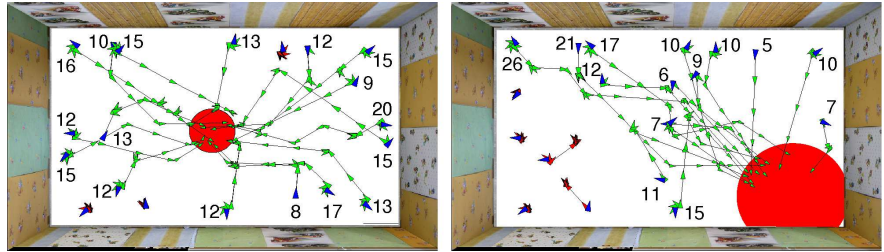}
\caption{Test trajectories of the \textit{robot experiment}.
         The blue triangles mark the start positions and the
         numbers the length of each trajectory which reached
         the red goal area.}
\label{fig:robotest_sfa}
\end{center}
\vspace{-.5cm}
\end{figure}

Backed up by the knowledge that LSPI works
on the theoretically predicted output of SFA,
we proceed with the analysis of LSPI on preprocessed
video images.

\paragraph{Simulated experiment}
First we wanted to know where the working regime
with respect the number of used SFA filters is.
We therefore repeated the former experiments
with the \textit{simulated experiment}
preprocessing described in section \ref{sec:exp_sfa}.
Utilizing the initial policy modification discussed
above, we eliminated the need of multiple trials.

As one can observe in figure \ref{fig:pi_nf_sim},
one needs much more SFA filters to reach a
comparable quality.
As a side effect, it seems as if we
did not reach the overfitting regime, at all.
The policies trained with 80 to 500 SFA
filters seem to be in the working regime.
Though the value is low compared
to the predictions, the right side of
figure \ref{fig:pi_nf_sim} shows that
this translates only to 2 to 3 steps
more than the reference policy.

\paragraph{Robot experiment}
Based on this experiment, we chose $d=128$
SFA filters in the \textit{robot experiment}.
A test trajectory driven by a real robot 
took the author between 5 and 10 minutes
and thus generating enough trajectories to
estimate the convergence quality was not possible.
Figure \ref{fig:robotest_sfa} shows all
20 test trajectories and the time each took.
With the exception of the lower left area in
the right plot the control seems to work
reasonably well.
However, the steps needed from near by
positions vary and most
trajectories took surprisingly long.

\subsection{Other environments}
\label{sec:exp_nav_two}

\begin{figure}[t]
\begin{center}
\vspace{-.2cm}
\includegraphics[scale=.31]{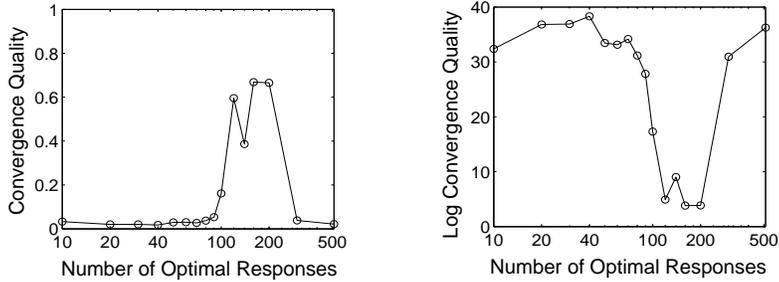}
\caption{The same experiments as in
        figure \ref{fig:pi_overfitting},
        but with $n=50000$ training samples
        in the \textit{two-room experiment} with
        artificial state representation.}
\label{fig:pi_nf_egypt}
\end{center}
\vspace{-.5cm}
\end{figure}

Now that we have established our method
to be working in simple cases, we want to examine
the more complex case of the two-room environment.
For once, the length scale is much larger.
For example, crossing the environment from left to right,
previously done in 9 move-actions, requires 45 actions now.
Moreover, the non-rectangular shape will lead to non-concave
value functions with sharp edges at the walls of the corridor.
One has to expect that at least the latter will put 
a strain on the value approximation quality.

Reviewing the results of section \ref{sec:exp_sfa_two},
the diversification of the filters in room dependent
trigonometric basis functions appear in a new light.
Since the covered areas are convex, abrupt
changes in the value function appear only \textit{at the
contact point} between rooms and therefore filters.

However, since we do no know the optimal
SFA responses of the two room environment,
we decided to stick to the known artifical
responses as described in section \ref{sec:nav_trig}.
The convergence quality with respect to the number
of optimal responses is shown in figure \ref{fig:pi_nf_egypt}.
Compared to the original task depicted in figure
\ref{fig:pi_overfitting}, one needs much more
responses (at least 100) to reach the working regime.
Also the convergence quality is much lower,
translating in the best cases to ca. 4 unnecessary steps.

To sum it up, the task has proven to be \textit{much} 
more complicated.
Besides, the SFA task is probably harder too,
leading to a more distorted state representation.
However, we performed a series of test with 
preprocessed video images from the simulator, 
using the first 10 to 500 filters.
In terms of convergence quality,
no learned control differed much from the base line in
figure \ref{fig:pi_nf_egypt}.

\subsection{Discussion}
We successfully conducted the main experiment.
Assuming optimal responses, as few as 20 filters
are sufficient to learn a near optimal control.
Using too much filters, however,
produces catastrophic overfitting.
Switching to SFA preprocessed video images
yields, at least in principle, the same result.
However, we need much more filters (at least 80)
and the resulting control needs considerably
more steps to reach the goal area.
Albeit not as well validated as the simulator experiment,
our method has proven to work on a real robot as well.

However, both successful experiments based on video images 
exhibited erratic behaviour at some points. 
The robot first gets caught or is ''undecided''
between left and right rotations but then escapes
without obvious reason.
Close observation reveals the effect to be
present only (but not always) when actions \textit{change}
(in opposition to continuous driving or rotation).
At these positions two or more actions
must have similar Q-values and
noise in the representation can induce
erratic behaviour.
This kind of behaviour has not been observed using
artificial state representations, which supports
the conclusion.


\section{Conclusion}
We wanted to give a proof of concept that LSPI
can learn a efficient control based only on
the video images preprocessed by filters learned by SFA.
Given the results of the last section, we can say we have.
However, we also demonstrated that with the current computational
capacity we can only solve simple problems, yet.

From our analysis we can identify the most likely 
sources of error:
\begin{itemize}
\item \textbf{Sensor errors}:
      Real video images are not constant over
      time, so the SFA output will always 
      be afflicted by noise.
      While noise afflicted \textit{transitions} were always
      part of the policy iteration framework,
      up to now noise afflicted \textit{state representations}
      received little attention.
      The effects of this noise appear in form of
      the discussed ''uncertain'' behaviour.
\item \textbf{Preprocessing errors:}
      SFA output does not resemble trigonometric
      polynoms \textit{perfectly}. 
      This can induce local anomalies and therefore 
      local value estimation errors.
      For example, the simulated experiment
      requires much more SFA filters
      even though sensory noise does not play a role.
\item \textbf{Approximation errors:}
      Abrupt changes in the value function
      (e.g. at walls) can lead to local approximation
      errors, because the function class at hand
      can not represent the slope.
      For example, in the artificial two-room experiment
      one needs much more optimal responses,
      due to the tight corridor.
\end{itemize}
However, the most dangerous source of error is over- 
and underfitting.
While underfitting is simply a question of
computational power, overfitting
could be lessened by regularization and more
training samples.

\chapter{Conclusion}
\label{chapter:outlook}

\section{Summary}

Applying reinforcement learning methods
onto real-world scenarios bears one
overwhelming obstacle: The state of the world.
The most common mathematical formalism of
Markov decision processes (MDP) requires at all times
the \textit{complete} state in a \textit{suitable} representation.
Even if one assumes real-world sensor
readings to hold the complete state,
its complex and noise afflicted structure
is hardly suitable for any learning algorithm.

The most common approach to this problem are
hand-crafted heuristics to filter state information
out of raw sensor data.
However, since unforeseen situations and interactions \textit{will} occur,
this approach is bound to fail 
outside the controlled conditions of a laboratory.
Biological systems, on the other hand,
demonstrate the ability to \textit{learn}
efficient representations of their environment.
These representations do not only
represent the environment well,
but can also adapt when
conditions change.
Thus, faced with real-world scenarios, 
one should rather \textit{learn} a 
representation of sensor data
than simply \textit{define} a heuristic one.

In this thesis we have explored the
case of robot navigation.
As a handicap, the only accessible state
information were the images of a head mounted video camera.
We chose a well known reinforcement learning
algorithm, \textit{least squares policy iteration} (LSPI),
to learn a navigational control.
The only reinforcement signal were a reward in the
goal area and a punishment near walls.

LSPI is based on the 
\textit{least squares Q-value} algorithm (LSQ),
which fits a linear function to approximate
the Q-value of all state-action pairs.
After learning, the control will choose
the action with the highest Q-value in the current state.
Given the state in form of the current video image,
a linear function is not powerful enough
to approximate the Q-value function well.

From approximation theory we know of
several sets of basis functions, that are particularly
suited to approximate continuous functions
by linear weighting.
To allow LSPI to learn the task, we therefore
must aim for a set of filters that
extract the state in the form of suitable basis functions.
This thesis is focused on
the unsupervised method of \textit{slow feature analysis} (SFA),
which produces filters that are
in the limit equivalent to
trigonometric polynomials.

\subsection*{Approach}
Assume a static environment in which a
wheeled robot can navigate
by moving forward and rotating.
The only non-static variable is the
robots position and orientation.
Since all possible images depend
exclusively on these variables
they form the complete \textit{state}.
In this environment, we record a random walk in
which orientation and position changes \textit{slowly}.

The main principle of SFA, temporal coherence,
aims to construct filters that
minimize the change in the filtered training set.
To avoid trivial solutions, the filters output
is bound to change by
constraining it to unit variance.
Another constraint, decorrelation,
ensures a set of \textit{orthonormal} filters.

Since these constraints force the filtered output
to change, the \textit{slowest} change will follow
the state of the random walk.
The specific optimization problem employed by SFA
leads to trigonometric polynomials
as optimal solution, i.e. assuming infinite
number of training samples and unrestricted model class.

The main experiment in this thesis consists of an
initial random walk video on which we learn a set of filters
with SFA.
After applying the learned filters onto the initial video,
the output is used as input of LSPI.
We obtain a linear weighting of 
filter outputs for every action, which 
approximates the Q-value.
Given a current video image, the control
will select the action with the highest Q-value,
i.e. that promises the
most future reward.
In our case that means to navigate as fast as
possible into the goal area, since it is the only
available reward, while avoiding walls.

\subsection*{Results}
We examined two environments.
The first was a simple rectangular laboratory area,
which was evaluated with a real robot and
as a simulation.
The second consisted of two quadratic rooms connected by
a small corridor. Here we performed only simulated experiments.

In the simulation of the rectangular environment, our method
produced controls that differed only around
2 to 3 steps per trajectory from the optimal choices.
The real robot also succeeded, but 
reached the goal only in $80\%$ of the test trajectories
and took significantly longer than expected.

The more complicated two-room environment
did not yield a working control.
However, tests under optimal conditions show that
the task is significantly more complex than the
former one.
It is the authors believe that with more 
computational resources in both phases,
SFA and LSPI, a successful control can be learned.

\newpage
\section{Discussion}
After presenting the methodology and results
in the last section, we want to discuss some
problems the author noticed during his work
on this thesis.
The discussion will first review general flaws,
followed by a close look into detailed
problems and some suggestions how to overcome them.

We will pick up some of the topics in the next section
to make concrete suggestions on possible
future advancements.

\subsection*{Slow feature analysis}
Despite the extension to kernel SFA,
slow feature analysis still suffers performance issues,
as the failure of the two-room experiment shows.
However, even if all practical problems can be solved,
one has to ask if a task independent preprocessing
can handle real scenarios at all.

Without any insight in the task at hand,
the extracted basis functions
have to cover all 
state space dimensions
\textit{and} all combinations of them.
Given a certain approximation
quality in every dimension 
(e.g. polynomials up to some \textit{degree}),
the number of
required filters grows exponentially in the
dimensionality of the extracted state space.
For example, a trigonometric representation
of a $p$ dimensional state space
with degrees $(d_1, \ldots, d_p)$ and 
$\forall i,j: d_i \approx d_j$ 
grows exponentially in $p$.
Due to mixtures in all but the first few filters,
we can not exclude unwanted dimensions or combinations,
even if we would know them.

Even more, we can not be certain that 
the world \textit{can} be modelled 
by a finite dimensional state space.
A complex system, e.g. a laughing human
or a broad-leafed tree in the breeze,
would require an inconceivable high dimensionality.
Even with a model class that can represent
the whole space, the number of required training samples
would go against infinity.

Thus, extracting the complete state space
is only feasible in very controlled, unique cases.
In the end we would like to extract only
the subspace that is \textit{useful}
for the general ''class'' of problems at hand.
For example, if every considered task in
some room is independent of the horizontal
position, a trigonometric representation
of vertical position and orientation
would be sufficient for any navigational control.
In this context the state space must not represent
some underlying ''real'' causes
(e.g. of muscle movement) but more
problem dependent ''meta'' causes (e.g. smiling).
Section \ref{sec:out_out} will present
a more concrete proposal how to achieve this.

\begin{itemize}
\item
With the exception of the simple case of rectangular rooms,
there are no theoretical predictions for
optimal responses.
Given the variability of environmental states,
this can be seen as a major disadvantage.
For example, imagine a robot arm with 5 joints.
Ideally, one would expect a 5-dimensional 
combination of free- or cyclic-boundary conditions.
but as in the two room case, the arm movement
will be constraint by the environment.
Therefore the optimal responses will be unpredictable
in all but the most simple cases.
However, it is not clear if these unexpected responses
will facilitate or limit the filters approximation capability.

\item
This thesis always assumed the current
state of the environment to be encoded 
in the current video image.
Of course, this is not always the case.
In line with the simultaneous localization
and mapping (SLAM) method, one could include
short term memory to track the current position
in such cases.
However, this would require either a complete
reformulation of the optimization problem or
a sophisticated post-processing of the SFA output.
A much simpler heuristic solution is presented in
section \ref{sec:out_out}.
\end{itemize}

\subsection*{Policy iteration}
In the tabular representation,
i.e. with small set of discrete states,
policy iteration is guaranteed to converge
monotonically to an optimal policy.
This guarantee, however, does not extend
to the realistic case of approximated
Q-value functions.

The successive application of a policy
estimation and greedy policy improvement step
reminds of the \textit{expectation maximization algorithm}
(EM, e.g.~\cite{Bishop06}).
In a discrete version 
(e.g. \textit{K-means}, e.g.~\cite{MacKay03}), 
this popular algorithm is prone to 
become stuck at local minima.
This effect resembles the results
of policy iteration in the 
\textit{unstable regime} (see sec.~\ref{sec:nav_trig}).
In difference to policy iteration, 
for EM a number of
probabilistic extensions exist
(e.g. \textit{soft K-means}, e.g.~\cite{MacKay03}),
that greatly reduce (albeit not eliminate)
this problem.

However, as the modified initial policy in
section \ref{sec:nav_trig} demonstrated, it
is possible to use probabilistic
choices instead of a pure greedy policy.
In section \ref{sec:out_out} we will discuss
a consideration of \textit{error sources}
(discussed below) into the policy improvement step.
When these problems are discussed in literature,
the only considered errors
are
due to a weak model class.
In practice, however, two
other sources of error have come to light:
\begin{itemize}
\item
Every real preprocessing will exhibit
some anomalies. The resulting local approximation
errors can be amplified by the policy iterations
and thus ruin an otherwise good policy.

\item
Since real sensors are afflicted by noise,
the control will exhibit erratic behaviour in
areas in which Q-values of the best
actions are close by.
In the worst case, oscillating behaviour can occur,
e.g. turning back and forth.
\end{itemize}

\subsection*{Application}
At last we have to review the general applicability
of the presented algorithms:
\begin{itemize}
\item
Restricting oneself to \textit{discrete actions}
greatly reduces the area of application.
The only way to circumvent this, discretizing continuous actions,
increases the number of actions immensely.
Because the Q-value estimator LSQ scales cubic in
the number of actions, this quickly becomes infeasible.

\item
Constructing a suitable reinforcement
learning scenario is not as intuitive as is might look like.
In particular the choice of hyper parameters ($\gamma$,
number of filters and amount of reward/punishment)
influences the resulting policy considerably.

\item
The presented method looks promising,
but sticks out by its high computational demand.
Due to its biological motivation,
its computational elegance does not
transfer to von Neumann architectures
of most modern computers.
From this background it looks natural to
use parallel processing architectures, e.g.
graphic cards, programmable logic arrays or
vector processors.
Indeed, with these
one could consider far more complex models
in reasonable time.

However, both SFA and LSPI algorithms are
in the presented form not parallelizable.
Besides many matrix multiplications,
which can be parallelized well in a
shared memory architecture,
the key elements are matrix inversions
or eigenvalue decompositions.
There exist 
parallel versions of both problems,
but they are not as efficient
and may introduce additional side effects.

In the end, however, the human brain performs these tasks
in a massive parallel processed fashion.
This alone makes it a promising approach.
\end{itemize}

\section{Outlook}
\label{sec:out_out}
As discussed above, application of reinforcement learning
to real-world scenarios still bears many problems.
The author wants to give here only a small selection
of potential improvements, that followed directly
from his work on this thesis.

\subsection*{Exclude unwanted states from SFA}
\label{sec:out_exclude_states}
To apply the proposed method to non-static environments,
one has to exclude unnecessary state subspaces
from the SFA solution.
Due to mixtures of the optimal responses,
one can not directly exclude single filters.
Instead, the optimization problem has to include
a term that penalizes dependence on unwanted dimensions.
It is important that the resulting filters
do not specialize too much in the task at hand.
For example, in a navigation task a
set of filters that support only policies
that drive to the center of the room 
would not work with another goal.

In line with \textit{canonical correlation analysis}
and \textit{partial least squares} \cite{Shawe04}, 
one possible approach would maximize the
correlation of the filters to a target value.
As a complication, 
we do not know a target value directly, but only
the experienced reward.

\newpage
\subsection*{Short term memory in preprocessing}
\label{sec:out_memory}
A thorough consideration of holding a current state
would require a complete reformulation of the
SFA optimization problem.
However, an easy consideration of the predecessor state
(e.g.~\textit{temporal smoothing} 
$x_t = \eta\, x_{t-1} + (1-\eta) \phi_t$)
could stabilize the current state estimation.
Since the state $x_{t-1}$ depends on
all previous observations $\phi_0, \ldots, \phi_{t-1}$,
the same \textit{position} can be represented by
different \textit{states} $x_t$, depending on the past.
This will induce additional ''noise'' in the
representation, so the parameter $\eta$ should be
chosen very small.

\subsection*{Consider local errors in policy iteration}
In section \ref{sec:nav_trig} we introduced a simple
modification to the initial random policy of LSPI.
Instead of sampling the second action randomly,
we considered all actions:
$\phi'(x) = \frac{1}{|\Set A|} \, \sum_{a \in \Set A}\phi(x,a)$.
In the following iterations, however, we chose this
action greedy.
If we review the sources of error from the previous section,
we see that these errors are not distributed equally
over the state space:
\begin{itemize}
\item
Weak function classes 
(e.g. not enough filters) have problems to
approximate the slope of abrupt changes in the reward.
However, most areas of the value function are smooth enough.
\item
In areas with similar Q-values
noise affected controls will show erratic behaviour.
Where Q-values differ enough, 
the control will be stable.
\item
SFA filters tend to differ from the optimal responses
locally, when the considered function class 
can not map an image properly.
\end{itemize}
The first step would be to determine these local errors.
With errors and Q-values at hand, one can
calculate a realistic  probability $p(\cdot)$ for an action
to be chosen by a greedy policy.
Instead of just choosing one action greedy,
one could learn \textit{all} possible actions at once:
$\phi'(x) = \sum_{a \in \Set A} p(a)\, \phi(x,a)$.

After the experience with the new initial policy,
one can hope that the resulting policy
iteration will be more stable and therefore
converge better.

\subsection*{Policy dependent norm}
LSPI is based on the squared $L_2$ norm,
which means it aims to distribute the approximation
error equally on all training samples.
However, as we have seen in our previous discussion,
areas with nearby Q-values need much better
approximation than those far-off.
It would improve the reliability of the
control if one would use a policy dependent
norm, i.e. put more weight on samples with
similar Q-values.

\appendix 

\newpage

\end{document}